\documentclass{article} 
\usepackage{iclr2026_conference,times}

\usepackage{hyperref}
\usepackage{url}
\usepackage{booktabs}
\UseRawInputEncoding

\title{ReJump: A Tree-Jump Representation for\\Analyzing and Improving LLM Reasoning}

\usepackage{amsmath,amsfonts,bm}

\def\eqref#1{(\ref{#1})}

\def\set#1{\{#1\}}
\def\1{\bm{1}}

\newcommand{\valid}{\mathcal{D_{\mathrm{valid}}}}

\def\vi{{\bm{i}}}

\DeclareMathAlphabet{\mathsfit}{\encodingdefault}{\sfdefault}{m}{sl}
\SetMathAlphabet{\mathsfit}{bold}{\encodingdefault}{\sfdefault}{bx}{n}

\def\sI{{\mathbb{I}}}

\definecolor{yz}{HTML}{f15bb5}
\definecolor{lz}{HTML}{9b5de5}
\definecolor{sz}{HTML}{00bbf9}
\definecolor{sw}{HTML}{00f5d4}
\definecolor{wk}{HTML}{fee440}
\definecolor{jk}{HTML}{f1ad5b}

\def\trees{{ReJump}}
\def\treesextractor{{\trees{}-Extractor}}

\usepackage{enumitem}
\usepackage{cleveref}
\usepackage{multirow}
\usepackage{wrapfig}

\usepackage{amsthm}
\usepackage{amsmath,amssymb}
\usepackage{graphicx}
\usepackage{subcaption}

\def\leaf{{S_{\text{leaf}}}}
\def\corrleaf{{S_\text{leaf}^\star}}
\def\newleaf{{\vi_\text{leaf}}}
\def\newcorrleaf{{\vi_\text{leaf}^\star}}

\def\nth{{^{(n)}}}

\def\consec{{\operatorname{consec}}}

\def\solcount{{\mathrel{\#}_{\text{solution}}}}
\def\jd{{d_{\text{jump}}}}
\def\success{r_{\text{success}}}
\def\forget{\mathbf{1}_{\text{forget}}}
\def\valid{r_{\text{verify}}}
\def\overthinking{r_{\text{overthinking}}}

\def\similaritytree{\operatorname{Sim}_{T}}
\def\similaritywalk{\operatorname{Sim}_{J}}

\newcommand{\oset}[1]{{\operatorname{set}(#1)}}

\def\asolcount{{$\mathrel{\#}_{\text{solution}}$}}
\def\ajd{{$d_{\text{jump}}$}}
\def\asuccess{{$r_{\text{success}}$}}
\def\aforget{{$r_{\text{forget}}$}}
\def\avalid{{$r_{\text{verify}}$}}
\def\aoverthinking{{$r_{\text{overthinking}}$}}
\def\asimilaritytree{$\operatorname{Sim}_{T}$}
\def\asimilaritywalk{$\operatorname{Sim}_{J}$}

\def\passone{{pass$\char64$1}}

\crefname{section}{Sec.}{Secs.}
\Crefname{section}{Sec.}{Secs.}
\crefname{figure}{Fig.}{Figs.}
\crefname{table}{Tab.}{Tabs.}

\usepackage{amssymb}

\definecolor{pinegreen}{rgb}{0.0, 0.47, 0.44}
\definecolor{cornellred}{rgb}{0.7, 0.11, 0.11}
\definecolor{cadmiumgreen}{rgb}{0.0, 0.42, 0.24}
\definecolor{royalblue}{rgb}{0.0, 0.14, 0.4}
\definecolor{spirodiscoball}{rgb}{0.06, 0.75, 0.99}
\definecolor{mylightblue}{rgb}{0.85, 0.90, 0.94}
\definecolor{kaistblue}{RGB}{20,135,200}
\definecolor{auburn}{RGB}{166,38,57}

\hypersetup{
  urlcolor = cadmiumgreen,
  linkcolor = cornellred,
  citecolor  = cadmiumgreen,
  colorlinks = true,
}

\def\gemini{{Gemini 2.5 Pro}}
\def\grok{{Grok 3 Mini Beta}}
\def\qwq{{QwQ-32B}}
\def\deepseekr{{DeepSeek-R1}}
\def\phif{{Phi-4-reasoning-plus}}
\def\claude{{Claude 3.7 Sonnet}}

\usepackage{tcolorbox}
\newtcolorbox{highlight}[1][]{
    colback=yellow!10,
    colframe=gray!30,
    boxrule=0.5pt,
    arc=2pt,
    leftrule=3pt,
    rightrule=3pt,
    toprule=0pt,
    bottomrule=0pt,
    before skip=4pt,
    top=3pt,
    bottom=3pt,
    after skip=4pt,
    boxsep=.5pt,      
    #1
}

\def\vphi{{\boldsymbol{\phi}}}

\usepackage{tikz}

\usepackage{tikz}
\definecolor{mycolor1}{RGB}{241, 241, 241}
\definecolor{mycolor2}{RGB}{59, 66, 102}
\definecolor{mycolor3}{RGB}{246, 252, 255}
\definecolor{mycolor4}{RGB}{255, 251, 242}
\definecolor{mycolor5}{RGB}{254, 247, 250}

\newcommand{\boxtex}{}     
\newcommand{\boxtitle}{}   
\newcommand{\boxheight}{}   

\tikzset{
  pics/usert/.style args={#1,#2,#3}{
     code={
        \def \WA {#3}
        \def \WB {\WA/2}
        \draw [rounded corners, fill=mycolor1] (0,0) rectangle ++(10,-\WA);
        \node[text width=\linewidth] at (5,-\WB) {\textit{#2}};
        \path [rounded corners, fill=mycolor2] (10.2,0) rectangle ++(2,-0.8) node[pos=.5,text=white] {User};
     }
  }
}
\tikzset{
  pics/useri/.style args={#1,#2,#3,#4}{
     code={
        \def \WA {#3}
        \def \WB {\WA/2}
        \draw [rounded corners, fill=mycolor1] (0,0) rectangle ++(10,-\WA);
        \node[text width=0.5\linewidth] at (3,-\WB) {\textit{#2}};
        \path [rounded corners, fill=mycolor2] (10.2,0) rectangle ++(2,-0.8) node[pos=.5,text=white] {User};
        \node at (8,-\WB) {\includegraphics[width=.36\linewidth]{#4}}; 
     }
  }
}

\tikzset{
  pics/modelt/.style args={#1,#2,#3}{
     code={
        \def \WA {#3}
        \def \WB {\WA/2}
        \draw [rounded corners, fill=mycolor1] (0,0) rectangle ++(10,-\WA);
        \node[text width=\linewidth] at (5,-\WB) {\textit{#2}};
        \path [rounded corners, fill=mycolor2] (-2.2,0) rectangle ++(2,-0.8) node[pos=.5,text=white] {#1};
     }
  }
}
\tikzset{
  pics/modeli/.style args={#1,#2,#3,#4}{
     code={
        \def \WA {#3}
        \def \WB {\WA/2}
        
        \draw [rounded corners, fill=mycolor1] (0,0) rectangle ++(10,-\WA);
        \node[text width=.5\linewidth] at (3,-\WB) {\textit{#2}};
        \path [rounded corners, fill=mycolor2] (-2.2,0) rectangle ++(2,-0.8) node[pos=.5,text=white] {#1};

        \node at (8,-\WB) {\includegraphics[width=.36\linewidth]{#4}}; 
     }
  }
}

\tikzset{
  pics/simple/.style args={#1,#2,#3}{
     code={
        \def \WA {#3}
        \def \WB {\WA/2}
        \draw [rounded corners, fill=mycolor1] (0,0) rectangle ++(15,-\WA);
        \node[text width=\linewidth] at (7.5,-\WB) {\textit{#2}};
        \path [rounded corners, fill=mycolor2] (0,0.2) rectangle ++(15,0.8) node[pos=.5,text=white] {#1};
     }
  }
}

\usepackage{caption}
\usepackage{listings}
\usepackage{xcolor}
\definecolor{codegreen}{rgb}{0,0.6,0}
\definecolor{codegray}{rgb}{0.5,0.5,0.5}
\definecolor{codepurple}{rgb}{0.58,0,0.82}
\definecolor{backcolour}{rgb}{0.95,0.95,0.92}
\lstdefinestyle{mystyle}{
    backgroundcolor=\color{gray!10},
    basicstyle=\ttfamily\footnotesize,
    breakatwhitespace=false,         
    breaklines=true,                 
    captionpos=b,                    
    keepspaces=true,                 
    numbers=left,                    
    numberstyle=\tiny,
    numbersep=5pt,                  
    showspaces=false,                
    showstringspaces=false,
    showtabs=false,                  
    tabsize=2
}
\crefname{lstlisting}{Listing}{Listings}
\Crefname{lstlisting}{Listing}{Listings}

\usepackage{titletoc}

\author{
\parbox{\textwidth}{ 
Yuchen Zeng\thanks{Equal Contribution. Emails: \texttt{yuchen.zeng.1998@gmail.com}, \texttt{shuibai@cs.wisc.edu}, \texttt{kangwj1995@furiosa.ai}.}~~\textsuperscript{\mdseries 1,2},~
Shuibai Zhang$^{*}$\textsuperscript{\mdseries 1},
Wonjun Kang$^{*}$\textsuperscript{\mdseries 3,4}, 
Shutong Wu\textsuperscript{\mdseries 1}, 
Lynnix Zou\textsuperscript{\mdseries 1}, \\
Ying Fan\textsuperscript{\mdseries 1,2},
Heeju Kim\textsuperscript{\mdseries 3},
Ziqian Lin\textsuperscript{\mdseries 1}, 
Jungtaek Kim\textsuperscript{\mdseries 1},
Hyung Il Koo\textsuperscript{\mdseries 3}, \\
Dimitris Papailiopoulos\textsuperscript{\mdseries 1,2}, 
Kangwook Lee\textsuperscript{\mdseries 1,5} 
}
\\
\\
\parbox{\textwidth}{
\textsuperscript{\mdseries 1} UW-Madison \quad 
\textsuperscript{\mdseries 2} Microsoft Research \quad
\textsuperscript{\mdseries 3} FuriosaAI \quad \\
\textsuperscript{\mdseries 4} Seoul National University \quad 
\textsuperscript{\mdseries 5} KRAFTON
}
}
\newcommand{\std}{\mathrm{std}}
\newcommand{\PromptSensitivity}{\text{PromptSensitivity}}

\iclrfinalcopy 
\begin{document}

\maketitle

\vspace{-.2in}
\begin{abstract}
Large Reasoning Models (LRMs) are Large Language Models (LLMs) explicitly trained to generate long-form Chain-of-Thoughts (CoTs), achieving impressive success on challenging tasks like math and programming. 
However, their underlying reasoning ``algorithms'' remain poorly understood. 
To investigate this, we propose \textit{ReJump}, which represents a reasoning trace as a visitation order over nodes in a tree of intermediate problem-solving steps. 
Transitions between nodes, which we term \textit{jumps}, include adjacent moves that capture behaviors such as calculation, and non-adjacent moves that capture behaviors such as backtracking and verification.
ReJump enables analyzing LLM reasoning with diverse metrics that quantify exploration, exploitation, overthinking, forgetting, and verification.
Using our proposed LLM agent to extract reasoning traces into ReJump format, we evaluate state-of-the-art LRMs on two tasks and find that models with similar accuracy can exhibit distinct reasoning behaviors, while different tasks favor different reasoning styles (e.g., varying balance between exploration and exploitation). 
To further understand how learning strategies shape reasoning, we use ReJump to compare distilled LRMs with their teachers, CoT-prompted LLMs with LRMs, and to examine how the number of reasoning examples and reinforcement learning affect reasoning behavior.
Finally, we show that ReJump can improve reasoning quality at test time through strategies such as ReJump-guided Best-of-N selection and prompt selection.
Our code is publicly available at~\url{https://github.com/UW-Madison-Lee-Lab/ReJump}.
\end{abstract}

\vspace{-.2in}
\section{Introduction}
\vspace{-.1in}

Chain-of-Thought (CoT) prompting improves the performance of Large Language Models (LLMs) on complex tasks, such as mathematical problem solving.
This was achieved either by providing exemplars of step-by-step reasoning~\citep{cot} or by simply adding the instruction ``Let's think step by step'' to the prompt~\citep{zero-shot-cot}, which encourages the model to decompose problems into intermediate steps, yielding more accurate and interpretable outputs. 
Recent work goes further by internalizing multi-step reasoning through supervised fine-tuning or reinforcement learning, leading to the recent flourishing of Large Reasoning Models (LRMs), a class of LLMs explicitly trained to generate long-form CoT, such as \deepseekr{}~\citep{deepseek-r1}, o1~\citep{o1}, and QwQ-32B~\citep{qwq32b}.

\begin{figure}[t]
    \vspace{-.4in}
    \includegraphics[width=\linewidth]{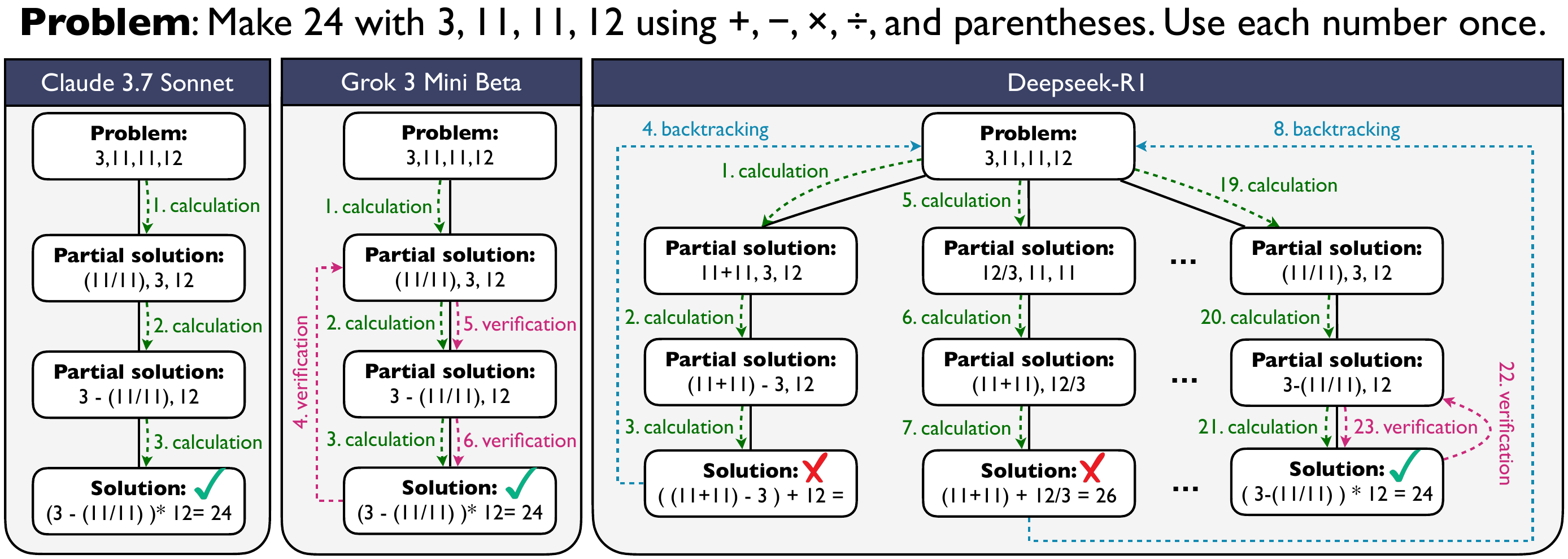}
    \vspace{-.25in}
    \caption{
   \textbf{\trees{} representations of reasoning traces generated by \claude{}, \grok{}, and \deepseekr{} on a Game of 24 problem.}
    All three models arrive at the same final answer, but their reasoning behaviors differ.
    Here, both \claude{} and \grok{} follow a single linear reasoning path; however, \claude{} adopts the answer without verification, while \grok{} verifies it before concluding.
    In contrast, \deepseekr{} explores multiple solution paths, exhibiting more deliberate behaviors such as backtracking and verification.
    }
    \label{fig:motivation}
    \vspace{-.3in}
\end{figure}

Comparisons among LRMs have so far focused primarily on final-answer accuracy. 
Yet, models arriving at the same answer may follow very different reasoning strategies, as illustrated in \cref{fig:motivation}.
Recent work has begun to explore other crucial dimensions of reasoning, such as overthinking~\citep{chen2025do} and underthinking~\citep{underthinking}, a comprehensive understanding of reasoning behavior is still lacking.
For instance, analyzing how a model balances exploration and exploitation or how much it forgets during reasoning could offer deeper insights into its core capabilities. 
This motivates the need for tools to systematically analyze and compare reasoning processes, raising the following question:

\begin{center}
\vspace{-.1in}
    \emph{How can we represent an LLM's reasoning trace to facilitate a comprehensive analysis and comparison of its internal behaviors?}
    \vspace{-.0in}
\end{center}

A natural way to address this question is through a tree-based representation, which captures the overall structure of reasoning, including planning and action transitions.
The value of such frameworks was noted by \citet{wu2025thought}, who introduced an abstract reasoning tree as a theoretical formalism capable of identifying when a model's thought process plateaus.
However, that work does not operate directly on a reasoning tree and instead relies on hidden states to predict whether a model's thought process plateaus. 
To bridge this gap, we propose \textit{\trees{}}, a tree-jump representation of LLM reasoning that enables comprehensive evaluation and comparison of reasoning traces and can further be leveraged to improve reasoning accuracy.
We summarize our main contributions as follows.
    \vspace{-.1in}
    \paragraph{We introduce \textit{\trees{}}, a tree-jump representation of LLM reasoning.}
    \trees{} encodes a reasoning trace as a sequence of visited nodes in a hierarchical tree of intermediate problem-solving steps, where the visitation order reflects execution flow and behaviors such as verification, backtracking, and calculation.
    Unlike a standard tree walk~\citep{west2001introduction}, in which each pair of consecutive nodes is connected by an edge, reasoning traces may include transitions between non-adjacent nodes due to behaviors like backtracking or verification.
    We therefore term this movement between nodes as a tree \textit{jump}.
    As shown in \cref{fig:motivation}, this representation makes it easy to distinguish between different reasoning behaviors.
    Building on this representation, we define six metrics to quantify reasoning behaviors, along with tree and jump similarity metrics for comparison.

    \vspace{-.1in}
    \paragraph{We design an LLM agent that extracts reasoning traces into the \trees{} format.}
    Given a reasoning trace, our agent, termed \textit{\treesextractor{}}, perform this via two steps: (i) it first parses it into the \textit{tree layer}, where each node represents an intermediate step, and each edge encodes the logical dependency between steps, (ii) based on the tree layer, it constructs the \textit{jump layer}, which captures transitions between nodes along with their corresponding action types (verification, calculation, or backtracking).
    In automatic evaluation, the \trees{} representations produced by \treesextractor{} reach over 0.9 tree and jump similarity with human annotations on Game of 24. 
    In human evaluation, where annotators assess whether each generated \trees{} is correct, \treesextractor{} achieves over 80\% accuracy on MATH-500.

    \vspace{-.1in}
    \paragraph{We utilize \trees{} to evaluate and comparing reasoning traces across models, tasks, and settings.}     
    We show that models with similar final accuracy can reason in completely different ways, and different tasks also favor different types of reasoning strategies (e.g., varying exploration-exploitation balances). 
    Our analysis further compares reasoning traces across 
    (i) CoT-prompted LLMs and LRMs, showing that LRMs exhibit more deliberate reasoning behaviors such as exploration and verification, and improve performance by generating more diverse solutions, though not necessarily with higher per-attempt accuracy;
    (ii) distilled models and their teacher LRMs, showing that distilled models inherit reasoning behaviors from their teachers; 
    (iii) varying number of in-context reasoning examples, showing that including more reasoning examples does not always enhance problem decomposition but can induce reasoning actions such as verification and backtracking; and 
    (iv) different checkpoints during reinforcement learning with verifiable reward (RLVR)~\citep{deepseek-r1}, showing that RL reinforces task-preferred reasoning behaviors (e.g., models trained on tasks requiring more exploration exhibit increased exploratory reasoning throughout RL training).

    \vspace{-.1in}
    \paragraph{We leverage \trees{} to improve the reasoning performance of LLMs.} 
    Beyond analyzing the reasoning processes of LLMs, we show that \trees{} can enhance performance.  
    \trees{} enables Best-of-N (BoN) selection and prompt selection based on desired reasoning properties (e.g., more exploration when helpful).
    When applied to the Game of 24 benchmark, both methods yield improvements to the \passone{} score, with performance gains ranging from +6.8\% to +9.1\%.
    

\vspace{-.1in}
\section{Related Work}
\label{sec:related_work}
\vspace{-.1in}
\paragraph{Approaches for Reasoning Analysis.}
With the growing popularity of LRMs, multiple recent works have introduced approaches to analyze reasoning traces rather than focusing solely final accuracy~\citep{lot,minegishi2025topology,xiong2025mapping,meta_tree,dag-math}.
\citet{lot} introduce Landscape of Thoughts (LoT), a visualization method tailored for multiple-choice tasks. 
It represents each intermediate reasoning step as a vector by computing its perplexity-based distance to all answer options, and then projects these vectors into two dimensions using $t$-SNE for visualization of reasoning trace. 
They also propose three evaluation metrics: consistency, uncertainty, and perplexity, to analyze model behavior.
However, LoT relies on perplexity-based vectors that lack semantic interpretability of the reasoning process.
In contrast, several other studies have explored graph-based representations of reasoning traces hence enable quantitative analysis of reasoning traces~\citep{minegishi2025topology,xiong2025mapping,meta_tree}. 
\citet{minegishi2025topology} perform quantitative analysis via structural properties of graph such as cycles and diameter, and \citet{xiong2025mapping} propose metrics to evaluate the reasoning's exploration and idea integration behavior. 
\citet{meta_tree} also propose a graph-based view of reasoning to identify the failed-step fraction and investigate its effect on reasoning accuracy.
\citet{dag-math} propose DAG-Math, modeling reasoning traces as directed acyclic graphs (DAGs) and evaluating each trajectory by whether it reaches the correct sink node and by how many nodes in the trajectory fall outside the ancestor set of that sink node.
In contrast to all existing work, our representation, \trees{}, a tree-jump representation that more naturally reflects the thinking process, aligning with prior work such as Tree-of-Thought~\citep{tot}. 
Moreover, \trees{} enables evaluation of reasoning along key dimensions that prior approaches do not capture, including exploration–exploitation balance, overthinking, verification, and forgetting.

\vspace{-.1in}
\paragraph{Empirical Findings on Reasonings.}
Prior empirical studies on reasoning typically fall into three categories: (i) limitations in reasoning behavior~\citep{chen2025do,fan2025missingpremiseexacerbatesoverthinking,wu2025thought,underthinking}, (ii) impact of training algorithms~\citep{limit-of-rlvr,dang2025assessing}, and (iii) factors for effective reasoning~\citep{li2025llms}. 

First, a well-known issue of LRMs are overthinking~\citep{chen2025do,fan2025missingpremiseexacerbatesoverthinking,wu2025thought}, where models continue unnecessary reasoning even after reaching a correct solution, and underthinking~\citep{underthinking}, where they abandon promising reasoning paths too early, often reflecting excessive exploration.
To address overthinking, \citet{wu2025thought} introduce thought calibration to dynamically terminate generation by using probes to detect when the model's reasoning tree stops growing.
Second, the choice of training algorithm significantly influences reasoning behavior. 
\citet{limit-of-rlvr,dang2025assessing} observe that although RL-trained models outperform base models at small pass@$k$, they merely bias outputs toward rewarded reasoning paths without acquiring new reasoning capabilities, ultimately narrowing reasoning capacity and being surpassed by base models at large $k$. 
Third, recent work identifies key structural factors that contribute to effective reasoning~\citep{gandhi2025cognitivebehaviorsenableselfimproving,li2025llms}.
\citet{gandhi2025cognitivebehaviorsenableselfimproving} highlight behaviors in the base model such as verification and backtracking play a key role in enabling RL training to further develop reasoning ability and improve performance. 
Similarly, \citet{li2025llms} argue that the logical form of reasoning, rather than the content of individual steps, is key to LRM reasoning quality.
The \trees{} representation facilitates these analyses by systematically capturing overthinking, exploration-exploitation dynamics, and behavioral differences across different experiment settings. 

\vspace{-.1in}
\section{\trees{}: A Tree-Jump Representation of LLM Reasoning}\label{sec:tree}
\vspace{-.1in}
In this section, we introduce the \trees{} representation and metrics for analyzing a single tree-jump and comparing pairs of tree-jumps, and will describe how reasoning traces are extracted into the \trees{} representation in \cref{sec:extractor}.

\vspace{-.1in}
\subsection{Decomposing Reasoning into Tree and Jump Layers }
\vspace{-.1in}
We extract each model-generated reasoning into a \trees{}, a two-layer representation that captures both structure and actions of reasoning traces. 
\begin{itemize}[
    leftmargin=*,
    topsep=1pt,      
    partopsep=0pt,   
    parsep=1pt,      
    itemsep=1pt      
]
    \item 
    \textbf{Tree layer (structure)}: We define a tree $T = (V,E)$, where $V = \set{v_i}_{v=0}^{|V|}$ is the set of nodes and $E$ is the set of edges.
    Following \citet{tot}, each node $v \in V$ represents a partial solution, with the root node corresponding to the initial state containing no solution.
    An edge $e \in E$ indicates that the parent's partial solution is a direct prerequisite for the child's.
    \item      
    \textbf{Jump layer (action)}: 
    Let $\vi = (i_0, i_1, \ldots, i_K)$ denote the sequence of reasoning steps, where $i_k$ refers to the index of $k$-th visited node in the tree.
    The jump starts at $v_{i_0}$ (the root) and ends at $v_{i_K}$ (the final solution).
    Each transition between consecutive steps $(i_k, i_{k+1})$ is labeled with an action type $\phi_k \in \set{\texttt{calc}, \texttt{verify}, \texttt{backtrack}}$, where $k=0,\ldots, K-1$.
    Here, \texttt{calc} refers to generating an intermediate step via calculation or derivation.
    Both \texttt{verify} and \texttt{backtrack} involve returning to a previously visited node: \texttt{verify} checks its correctness, while \texttt{backtrack} restarts from it to explore an alternative reasoning path. 
    The sequence of actions in the jump layer is denoted by $\vphi = (\phi_0, \ldots, \phi_{K-1})$.
    A jump layer is defined as the pair $W = (\vi, \vphi)$, specifying the sequence of visited nodes and the corresponding transitions taken during reasoning.
\end{itemize}

\begin{figure}[t]
\vspace{-.5in}
        \includegraphics[width=\linewidth]{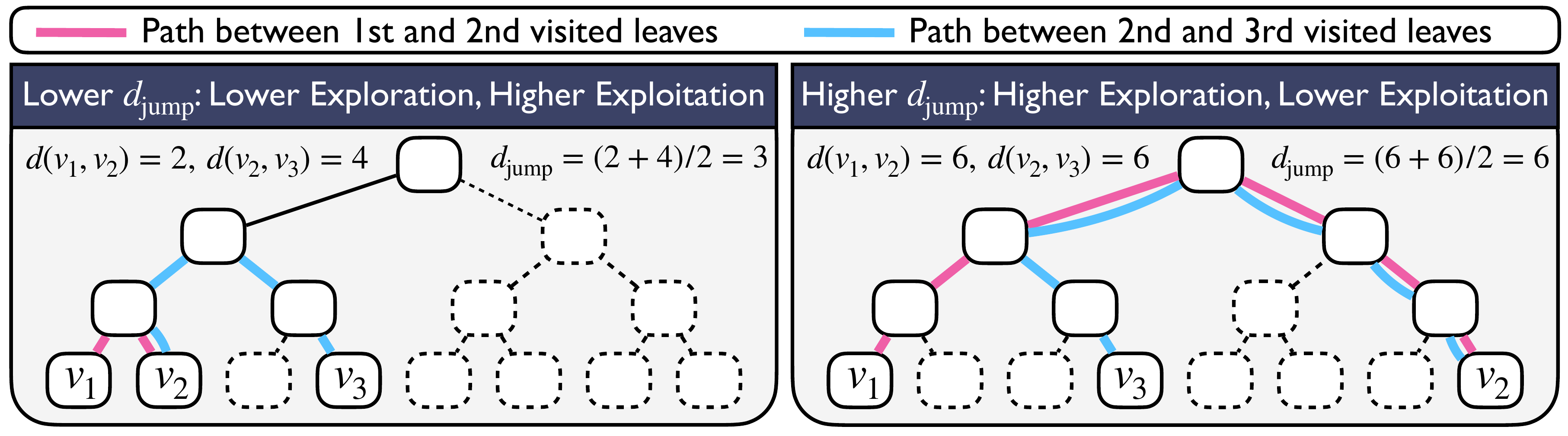}
        \vspace{-.2in}
        \caption{
        Illustration of how \ajd{} quantifies the exploration-exploitation trade-off in model reasoning.
        Given a sequence of visited leaf nodes $(v_1, v_2, v_3)$, the left panel depicts a trace exhibiting local exploration (shorter paths between nodes),
        while the right panel shows a trace with larger jumps to distant leaves, reflecting more global exploration.
        }
        \label{fig:jd}
        \vspace{-.2in}
\end{figure}

\vspace{-.1in}
\subsection{Quantifying Reasoning Behavior: Evaluation and Comparison Metrics}\label{sec:metrics}
\vspace{-.1in}
We define a \textit{derived solution step} as any step in the jump layer that reaches a leaf node via a \texttt{calc} transition, thereby contributing to the solution.
Steps that visit a leaf solely for verification are excluded, whereas revisiting an already-seen leaf via \texttt{calc} still qualifies as a derived solution step. 

\vspace{-.1in}
\paragraph{Evaluation Metrics.}
This tool enables analysis of LLM reasoning behaviors, including solution diversity, exploration--exploitation trade-off, effectiveness in identifying correct paths, frequency of overthinking, forgetting, and verification.
These aspects are quantified using the metrics below, computed across all reasonings and their corresponding \trees{}s within a task.
Each metric below is defined at the instance level and is directly averaged across instances to obtain the task-level score, except for \aforget{}, which is defined only at the task level and described accordingly.
\vspace{-.1in}
\begin{itemize}[
    leftmargin=*,
    topsep=3pt,      
    partopsep=0pt,   
    parsep=1pt,      
    itemsep=1pt      
]
    \item \textbf{Solution Count} (\asolcount{}): Number of distinct solutions, i.e., the total leaf nodes in the tree (including incomplete ones). 
    \item \textbf{Jump Distance} (\ajd{}): 
    Measures the exploration-exploitation balance by computing the tree distance (edge count) between the nodes of each consecutive pair of derived solution steps.
    For a reasoning instance, \ajd{} is the average distance across all consecutive pairs of derived solution steps within the jump layer.
    See \cref{fig:jd} for illustrative examples.
    \item \textbf{Success Rate} (\asuccess{}): Fraction of derived solution steps that yield a correct answer.
    Measures the efficiency and accuracy. 
    \item \textbf{Verification Rate} (\avalid{}): Fraction of all transitions labeled \texttt{verify}.
    Indicates how deliberate and self-critical the reasoning process is.
    \item \textbf{Overthinking Rate} (\aoverthinking{}):
     Fraction of derived solution steps that occur \emph{after} the first correct derived solution step is found, quantifying unnecessary exploration.
     Quantifies unnecessary exploration and inefficiency. 
    \item \textbf{Forgetting Rate} (\aforget{}): This metric is defined only at the task level. 
    A reasoning trace is flagged as \textit{forgetting} if it revisits the same leaf node via \texttt{calc}, indicating that the model has re-entered an already visited path; see \cref{fig:forget}. 
    \aforget{} is the proportion of such reasoning traces within the task, highlighting poor memory or state-tracking. 
\end{itemize}
See~\Cref{app:eval_metrics} for formal mathematical definition of these metrics. 
Furthermore, in \cref{app:redundancy}, we analyze redundancy among the proposed metrics and show that no metric in this set can be expressed as a deterministic function of the remaining metrics.

\vspace{-.1in}
\paragraph{Comparison Metrics.}
To assess the similarity between reasoning process produced by different models, we introduce similarity metrics to compare tree and jump representation, respectively. 
As with the evaluation metrics, all comparison metrics below are defined at the instance level and averaged across instances to obtain task-level scores.

\vspace{-.1in}
\begin{itemize}[
    leftmargin=*,
    topsep=3pt,      
    partopsep=0pt,   
    parsep=1pt,      
    itemsep=1pt      
]
    \item \textbf{Tree Similarity} (\asimilaritytree{}): 
    Measures the similarity in problem decomposition structure between two reasoning traces by comparing their corresponding trees.
    These metrics assess whether models adopt similar reasoning structures, without relying on the exact content of individual steps.
    This choice is justified by the work of \citet{li2025llms}, which demonstrates that overall logical structure, rather than the specific content at each node, is the primary factor influencing reasoning quality. 
    Given reasoning trees $T=(V,E)$ and $T' = (V', E')$, we compute their Zhang-Shasha Tree Edit Distance (TED) using Zhang–Shasha dynamic programming algorithm~\citep{paaßen2022revisitingtreeeditdistance}, which measures the minimum number of edit operations to transform one tree into another. 
    Note that we do not consider relabeling and only use insertions and deletions, since node semantics are not part of this setting.
    The tree similarity is then defined as $\similaritytree(T, T') = 1 - \operatorname{TED}(T, T') / \max(|V|, |V'|)$.
    \item  \textbf{Jump Similarity} (\asimilaritywalk{}): 
    Measures the similarity in action transition patterns between the jumps derived from two reasoning traces.
    For each reasoning jump $W = (\vi, \vphi)$, we construct a $3 \times 3$ transition probability matrix $P$, where $P_{a,b}$ is the empirical probability of transitioning from action $a$ to $b$, with $a,b \in \set{\texttt{calc}, \texttt{verify}, \texttt{backtrack}}$.
    Given two jump layers $W, W'$ with transition probability matrices $P$ and $P'$, we define their similarity as $\similaritywalk(W, W') = 1 - \operatorname{JS}(P \Vert P')$,
    where $\operatorname{JS}$ is the Jensen-Shannon divergence, a symmetric and bounded variant of KL divergence defined as
    $\operatorname{JS}(P \Vert P') = \frac{1}{2} \operatorname{KL}(P \Vert \frac{1}{2}(P + P')) + \frac{1}{2} \operatorname{KL}(P' \Vert \frac{1}{2}(P + P'))$.
    Higher \asimilaritywalk{} values indicate greater alignment in action transition behavior.
\end{itemize}

Note that our current tree and jump similarity metrics capture only the logical structure and action-transition distributions. 
This design keeps computation simple but can overlook important differences: two models may produce identically structured trees while differing in semantic content, and a perfect jump-similarity score may still mask distinct temporal patterns.
For example, one model may perform all derivations before verifying, while another alternates between computation and verification. 
Incorporating semantic information and temporal dynamics is therefore a promising direction of future work. 

\vspace{-.1in}

\vspace{-.1in}
\section{\treesextractor{}: Extracting CoTs into \trees{} Format}\label{sec:extractor}
\vspace{-.1in}
In this section, we introduce \treesextractor{}, an LLM agent for extracting reasoning traces into \trees{} format via two steps: 

\begin{figure}[t]
    \vspace{-.4in}
    \centering
        \includegraphics[width=\linewidth]{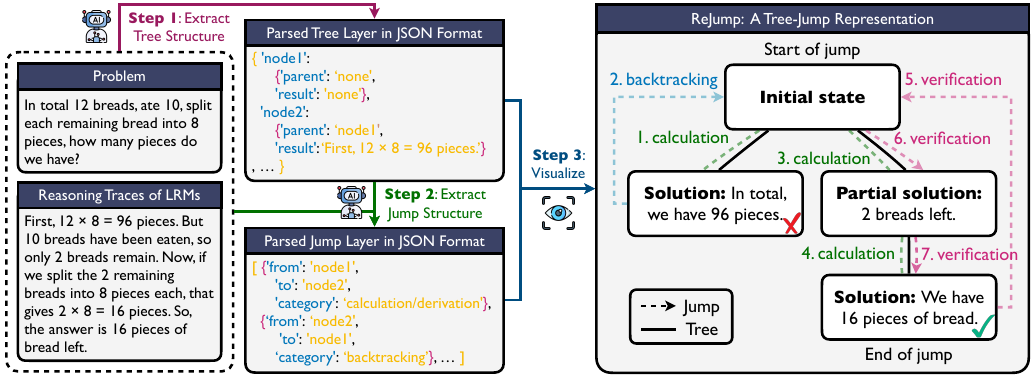}
            \vspace{-.2in}
    \caption{\textbf{Illustration of how \treesextractor{} converts a reasoning trace into the \trees{} representation for a math word problem.}
        This example is crafted for demonstration purposes.
        Nodes represent partial solutions, and tree edges indicate that the parent nodes serve as prerequisite for child nodes.
        Dashed arrows show how the reasoning moves between nodes, with each transitions corresponds to an action type (\texttt{calc}, \texttt{verify}, or \texttt{backtrack}), and color-coded accordingly.}
    \label{fig:tree_demo_simplified}
    \vspace{-.2in}
\end{figure}

\begin{itemize}[
    leftmargin=*,
    topsep=1pt,      
    partopsep=0pt,   
    parsep=1pt,      
    itemsep=1pt      
]
    \item \textbf{Tree Layer Extraction}:
    We use \gemini{}~\citep{gemini2.5} to extract both the tree and the jump representations from each reasoning trace.
    Given the original problem input and the model-generated reasoning, we prompt LLM to produce a JSON object that encodes the reasoning tree.
    This JSON is a dictionary where each key corresponds to a node index, and each value contains three fields: 
    ``problem'' (the subproblem addressed at that node), 
    ``parent'' (the index of the prerequisite node whose partial solution this node builds upon), and ``solution'' (the result corresponding to the subproblem).
    For the root node, all three fields are either left empty or labeled as ``initial state.''
    \item \textbf{Jump Layer Extraction}:
    We parse the JSON dictionary to construct the tree structure.
    Then, using the original input, the full reasoning, and the generated tree JSON as context, we prompt LLM again to extract the jump layer. 
    The jump layer is represented as a JSON list, where each entry describes a transition between nodes, with fields ``from,'' ``to,'' and ``category'' indicating the source node, target node, and transition type (e.g., calculation, verification, or backtracking). 
    We use this information to visualize the full reasoning trajectory overlaid on the constructed tree. 
\end{itemize}

\begin{figure}
    \centering
    \begin{minipage}[t]{.32\linewidth}
        \centering
        \vspace{0pt} 
        \begin{subfigure}[t]{\linewidth}
            \centering
            \includegraphics[width=.9\textheight, angle=90]{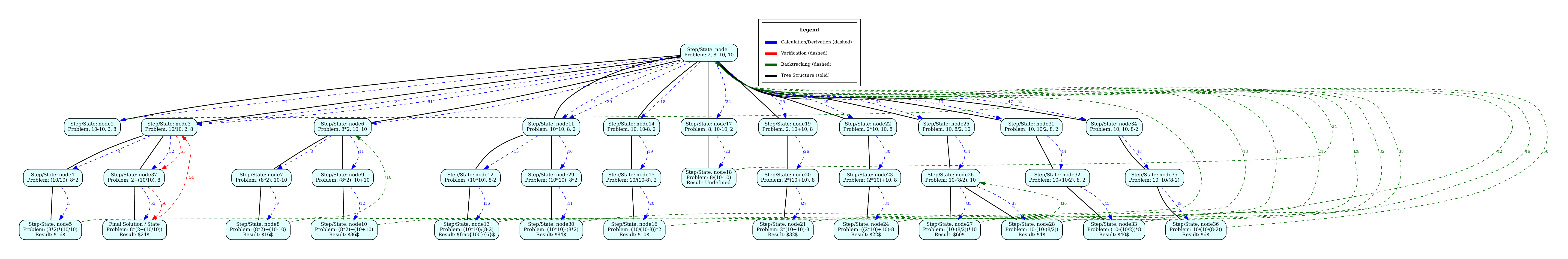}
            \caption{\deepseekr{}. Corresponding reasoning traces are shown in \cref{lst:deepseekr_game24}.}
            \label{fig:deepseek_36_game24}
        \end{subfigure}
    \end{minipage}%
    \hfill
    \begin{minipage}[t]{.66\linewidth}
        \centering
        \vspace{0pt} 
        \begin{subfigure}[t]{\linewidth}
            \centering
            \includegraphics[width=\linewidth]{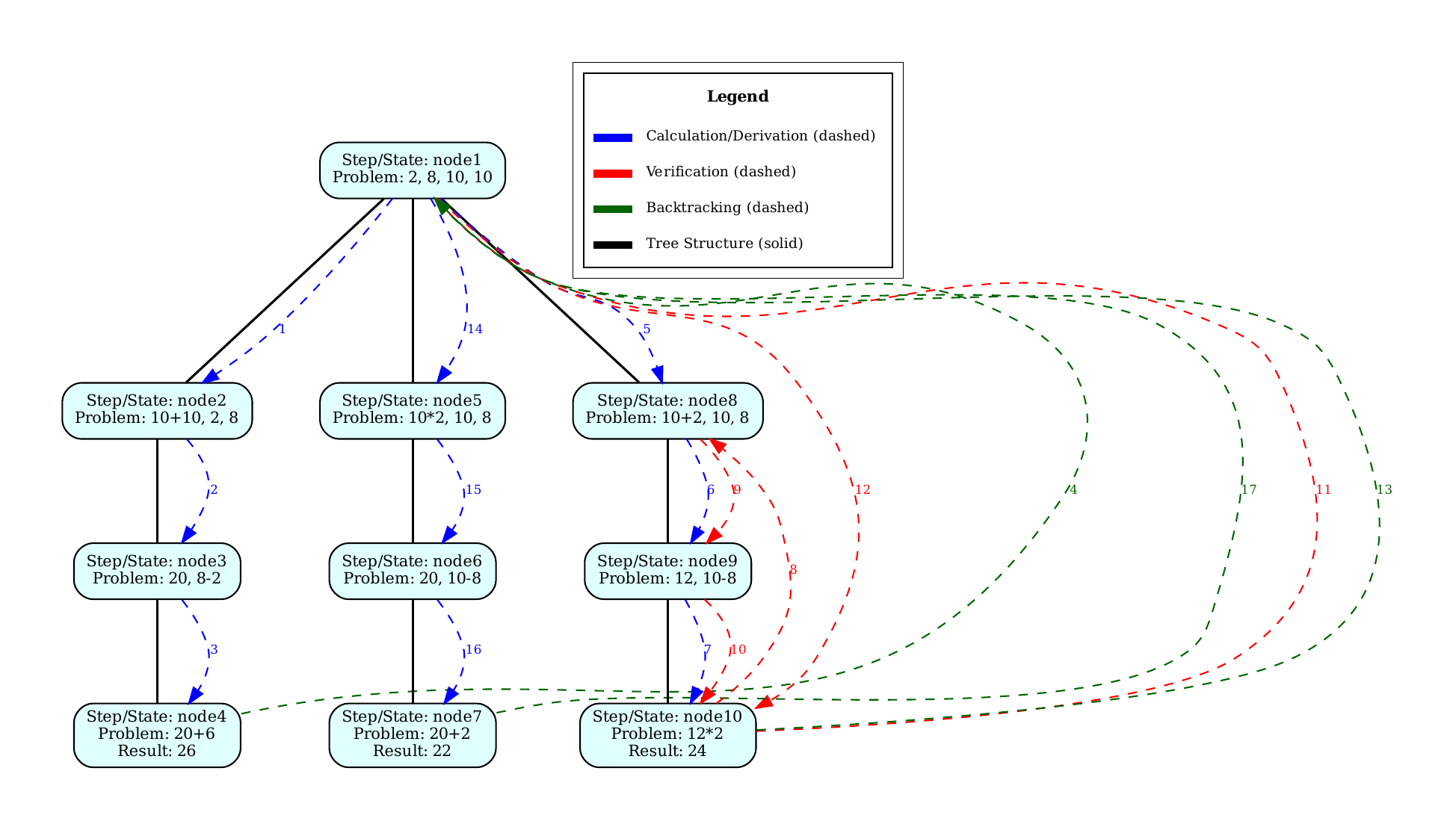}
            \caption{\phif{}. Corresponding reasoning traces are shown in \cref{lst:phif_game24}.}
            \label{fig:phi4_36_game24}
        \end{subfigure}
        
        \vspace{1em} 
        
        \begin{subfigure}[t]{\linewidth}
            \centering
            \includegraphics[width=\linewidth]{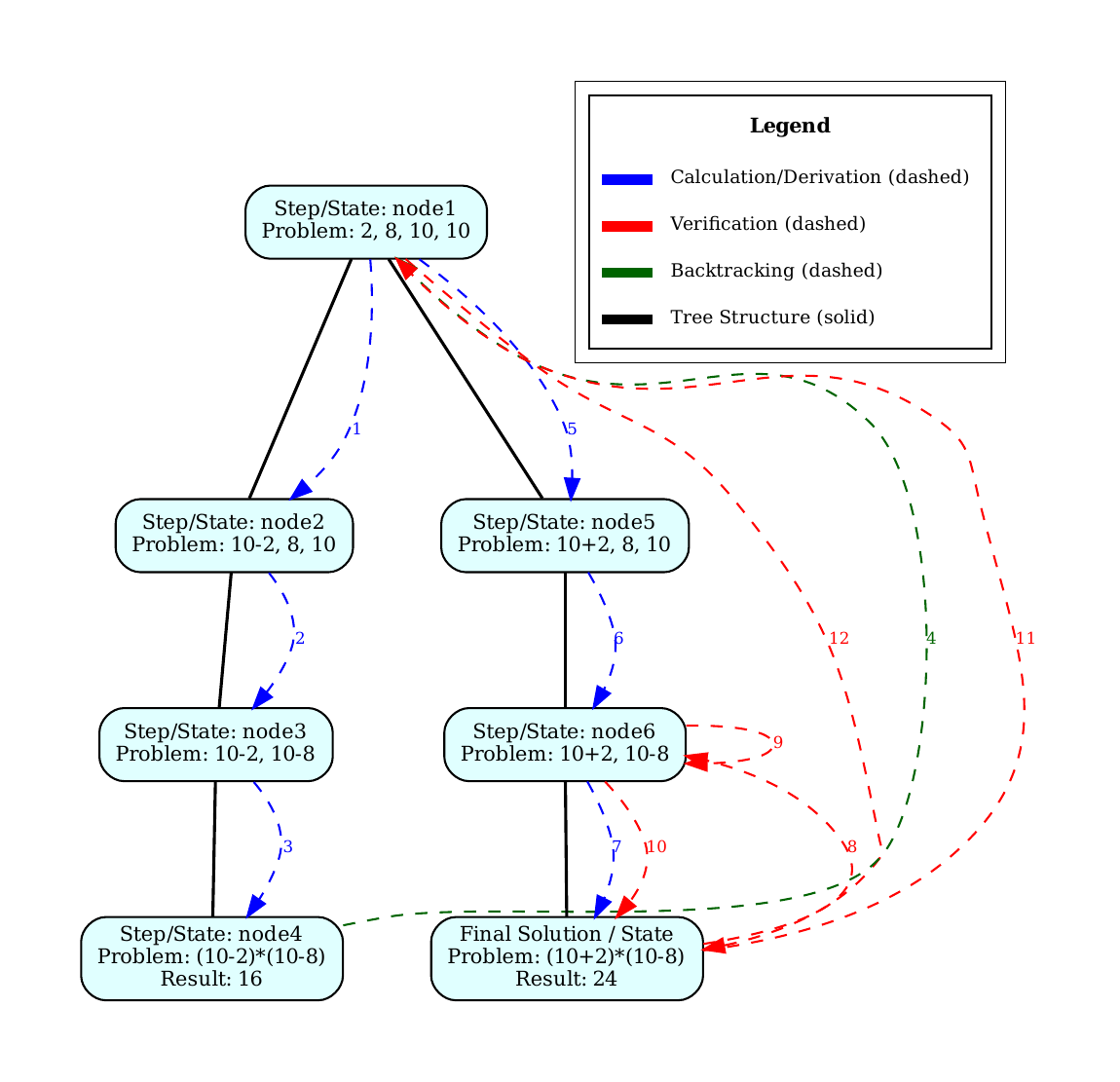}
            \caption{\claude{}. Corresponding reasoning traces are shown in \cref{lst:claude_game24}.}
            \label{fig:claude_36_game24}
        \end{subfigure}
    \end{minipage}
    
    \caption{\trees{} representations extracted by \treesextractor{} for reasoning traces generated by \deepseekr{}, \phif{}, and \claude{} for a Game of 24 problem.}
    \label{fig:real_3_game24}
\end{figure}

A visualization of how \treesextractor{} extracts CoTs into the \trees{} format is shown in \cref{fig:tree_demo_simplified}, and example \trees{} representations derived from real reasoning traces for Game of 24 are shown in \cref{fig:real_3_game24}. 
The extraction prompt is provided in \Cref{app:prompts}, and additional \trees{} examples from real MATH-500 traces are included in \Cref{app:real_trees}.


\vspace{-.1in}
\subsection{Assessing the Reliability of \treesextractor{}}\label{app:reliability}
\vspace{-.1in}
To evaluate the reliability of \treesextractor{}, we use two complementary methods: (i) automatic evaluation and (ii) human assessment. 
We apply automatic evaluation to Game of 24 for two reasons. 
First, it is straightforward to design reasoning traces for Game of 24 with controlled properties such as higher exploration or exploitation, allowing us to curate a dataset of traces paired with ground-truth trees for automated verification. 
Second, without this dataset, manual verification would be infeasible because Game of 24 traces often involve extensive exploration with many nodes. 
For MATH500, which is substantially easier for humans to verify and for which constructing a ground-truth dataset would require significantly more effort, we instead rely on manual inspection of the extracted \trees{} representations.

\vspace{-.1in}
\begin{table}[t]
\vspace{-.2in}
\begin{minipage}{.4\linewidth}
    \caption{Average alignment between \treesextractor{} outputs and GT \trees{} on Game of 24.}
    \label{tab:reliability_automatic}
    \vspace{-.1in}
    \centering
    \resizebox{\linewidth}{!}{
    \begin{tabular}{ccc}
    \toprule
       & \textbf{\asimilaritytree{}} & \textbf{\asimilaritywalk{}}  \\
        \midrule
       \treesextractor{} & .943 & .940 \\
    \bottomrule
    \end{tabular}}
\end{minipage}\hfill
\begin{minipage}{.52\linewidth}
    \caption{Human evaluation of decomposition of reasoning traces via \trees{} accuracy on MATH500. }
    \label{tab:reliability_human}
    \vspace{-.1in}
    \centering
    \resizebox{\linewidth}{!}{
    \begin{tabular}{cccc}
    \toprule
        \textbf{Reasoning Model} & \textbf{Pass@1} & \textbf{Pass@2} & \textbf{Pass@3} \\
    \midrule
        \deepseekr{} & 81\% & 87\% & 90\% \\
        \qwq{} & 80\% & 83\% & 94\% \\
    \bottomrule
    \end{tabular}}
\end{minipage}
\vspace{-.2in}
\end{table}

\paragraph{Automatic Evaluation.}
\textit{(Dataset)}
The dataset contains 82 reasoning traces, each paired with a manually constructed and verified ground-truth (GT) \trees{} representation. 
Constructing the dataset involves two steps. 
First, we create \trees{} examples with varying properties: low and high exploration, low and high verification, with and without forgetting, and with and without overthinking. 
The resulting trees range from 4 to 20 nodes. 
We then use \gemini{} to generate natural-language reasoning for each \trees{} representation, followed by human review and correction, which produces the final paired reasoning traces and GT \trees{} representations.
\textit{(Results)}
We use \treesextractor{} to process the reasoning traces and compare its outputs with the ground-truth data, reporting tree and jump similarities in \cref{tab:reliability_automatic}.
We find that \trees{} representations extracted by \treesextractor{} using \gemini{} align closely with the ground truth, confirming the reliability of our approach.
Comparison of \gemini{} with \claude{} as the base model for \treesextractor{} is provided in \cref{app:llm_compare}.
While \claude{} also performs well, \gemini{} remains the preferred choice due to its higher accuracy.


\vspace{-.1in}
\paragraph{Human Assessment.} 
We perform human evaluation of the \trees{} extracted by \treesextractor{} from reasoning traces generated by \deepseekr{} and \qwq{} on MATH500. 
For each prompt, we collect two reasoning traces, one from each model. 
For every trace, we run \treesextractor{} three times to obtain three \trees{} representations, then manually inspect each for correctness. 
This allows us to compute pass@1, pass@2, and pass@3 by considering the first representation only, the first two, and all three. 
Results are shown in \cref{tab:reliability_human}. 
A single extraction (pass@1) already yields high reliability, with accuracies of 81\% for \deepseekr{} and 80\% for \qwq{}, and multiple extractions further improve performance, surpassing 90\% accuracy by pass@3.

We also demonstrate that \treesextractor{} is robust to prompt variations in \cref{app:sensitivity}, and compare it against a simpler baseline that directly provides LLM reasoning traces and asking for metric values without converting them into \trees{} representations in \cref{app:direct_llm}.

\vspace{-.1in}
\section{Analyzing LLM Reasoning Behaviors with \trees{}}\label{sec:comparison}
\vspace{-.1in}

Building on the \trees{} representation introduced in \cref{sec:tree} and \treesextractor{} introduced in \cref{sec:extractor}, we analyze the reasoning structures of four state-of-the-art LRMs across two datasets.
The two datasets we consider are: (i) MATH-500~\citep{lightman2024lets}, a widely used benchmark for mathematical reasoning, and (ii) Game of 24, a task that requires strong planning and enables clear inspection of model behavior in terms of exploration, exploitation, and related strategies, as previously adopted by \citet{tot}.
Unless otherwise specified, all experiments use a decoding temperature of 0 to ensure deterministic and reproducible outputs. 
In addition, we examine the effect of alternative decoding strategies on reasoning behavior in \Cref{app:decoding}.

\vspace{-.1in}
\subsection{Comparing Reasoning Structure across State-of-the-Art LRMs and Tasks}\label{sec:benchmark}
\vspace{-.1in}

\begin{figure}[t]
\vspace{-.5in}
    \centering
    \includegraphics[width=\linewidth]{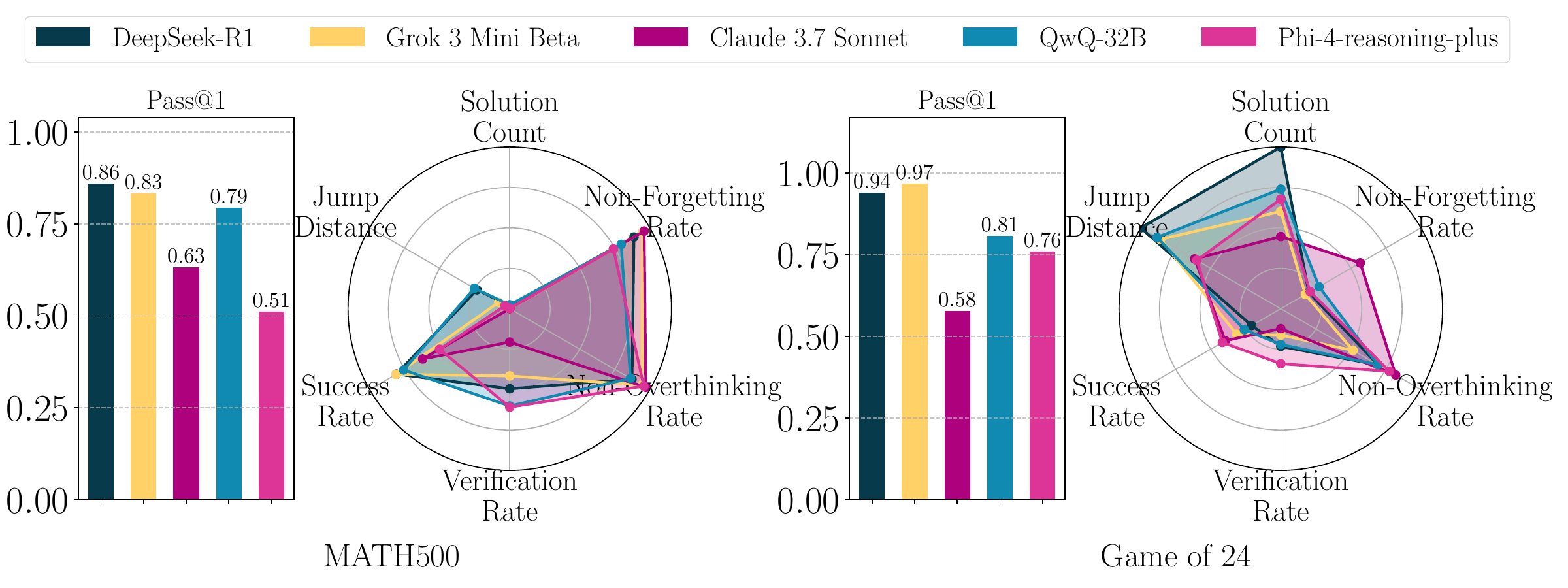}
    \vspace{-.3in}
    \caption{
        \textbf{Reasoning performance of \deepseekr{}, \grok{}, \qwq{}, \phif{}, and \claude{} on MATH-500 and Game of 24.}
        The bar plots present the final accuracy (\passone{}), while the radar plots detail six reasoning metrics. 
        For comparability, solution count and jump distance are normalized across all models and datasets. 
        To ensure that higher values consistently reflect preferred behavior, we report the non-forgetting rate and non-overthinking rate rather than forgetting rate and overthinking rate.
        The results show that models display distinct reasoning behaviors across datasets. Furthermore, even when models achieve similar final performance, their underlying reasoning processes can differ significantly.
        To better highlight metric differences among the strongest models \deepseekr{}, \grok{}, and \claude{}, \cref{fig:benchmark_top3} focuses exclusively on these three.
    }
    \label{fig:benchmark}
    \vspace{-.2in}
\end{figure}

This experiment focuses on the state-of-the-art LRMs for which we have access to intermediate reasoning traces: \deepseekr{}~\citep{deepseek-r1}, \qwq{}~\citep{qwq32b}, \grok{}~\citep{grok3}, \phif{}~\citep{phi4}, and \claude{}~\citep{sonnet3.7}.
Since \claude{} uses a fixed temperature of 1 in thinking mode, and \phif{} performs significantly better with temperature 1 than with 0, we set the decoding temperature to 1 for all models in this experiment. 
The more detailed setup and results for \deepseekr{}, \qwq{}, and \grok{} under temperature 0 are deferred to~\Cref{app:benchmark}.
We generate reasoning traces and final answers for both MATH-500 and Game of 24 using each model, and analyze their reasoning behaviors through \trees{}.
\cref{fig:benchmark} presents a unified view of final accuracy (\passone{}) alongside six reasoning evaluation metrics proposed in \cref{sec:metrics}.

\begin{wraptable}{r}{.6\linewidth}
\vspace{-.2in}
    \caption{
        \textbf{Contribution of reasoning metrics to solution correctness on MATH-500 and Game of 24 for \deepseekr{}, \grok{}, \qwq{}, \phif{}, and \claude{}.}
        ``Contribution'' is quantified by the feature-importance scores produced when an XGBoost model is trained on six reasoning-evaluation metrics to classify each solution as correct or incorrect; the model attains accuracies of 0.9197 (MATH-500) and 0.7980 (Game of 24) versus majority-class baselines of 0.6928 and 0.6768.
        Metrics with importance scores $>0.2$ are boldfaced.
        Game of 24 emphasizes exploration (i.e., \ajd{}), while MATH-500 emphasizes exploitation (i.e., \asuccess{}).
    }
    \label{tab:success_fail}
    \vspace{-.1in}
    \centering
    \resizebox{\linewidth}{!}{
    \begin{tabular}{ccccccc}
    \toprule
        Dataset & \asolcount{} & \ajd{} & \asuccess{} & \avalid{} & \aoverthinking{} & \aforget{}  \\ \midrule
        MATH-500 & .0177 & .0541 & \textbf{.8548} & .0173 & .0219 & .0341 \\
        Game of 24 & .1402 & \textbf{.2742} & \textbf{.2146} & .0980 & .1413 & .1317  \\ 
        \bottomrule
    \end{tabular}}
    \vspace{-.1in}
\end{wraptable}

\vspace{-.1in}
\paragraph{Comparison across Tasks.}
MATH-500 and Game of 24 differ in the structure and demands of reasoning. 
MATH-500 problems are typically deterministic, with only one or two valid solution paths, encouraging focused, step-by-step reasoning. 
In contrast, Game of 24 requires generating diverse arithmetic expression to reach the target number, promoting trial-and-error and exploratory strategies. 
This difference is evident in the results: as shown in \cref{fig:benchmark}, all five models yield a much lower \asolcount{} on MATH-500 compared to Game of 24.
The \ajd{} is also substantially higher for Game of 24.
Meanwhile, MATH-500 shows much higher success and verification rates, likely due to its proof-like structure that favors thorough verification and local exploitation over broad exploration. 
To systematically analyze which reasoning behaviors contribute most to final accuracy, we compute the feature importance of six reasoning metrics and summarize them in \cref{tab:success_fail}. 
We find that \asuccess{} has the strongest impact on \passone{} for MATH-500, while both \ajd{} and \asuccess{} are key predictors of \passone{} on Game of 24.

\vspace{-.1in}
\paragraph{Comparison across LRMs.}
Among the five models, \deepseekr{} and \grok{} achieve the highest \passone{}, followed by \qwq{}, with \phif{} and \claude{} lag behind.
Although \deepseekr{} and \grok{} reach similar final accuracy, their reasoning behaviors differ substantially.
Compared to \deepseekr{}, \grok{} adopts a narrower approach: it explores fewer paths and makes shorter jumps, yet reaches correct solutions more efficiently, as reflected in its higher success rate.
By contrast, \deepseekr{} engages in broader exploration, producing more candidate solutions and making longer jumps, though at the cost of a slightly lower success rate. 
Despite this, it ultimately achieves accuracy comparable to \grok{}.
\qwq{} exhibits less exploration than \deepseekr{} and a lower success rate than \grok{}, resulting in worse performance compared to both.
\phif{} and \claude{} perform the worst among all models, with even lower levels of exploration and success rates.
Notably, \claude{} exhibits the least deliberate reasoning behavior, as both exploration (\asolcount{} and \ajd{}) and \avalid{} are low. 
\phif{} demonstrates slightly more deliberate reasoning behaviors, with higher exploration than \claude{}, which contributes to its relatively better performance on Game of 24.
All models exhibit overthinking,
which is an issue previously observed in LRMs~\citep{chen2025do,yang2025dynamicearlyexitreasoning},
as well as forgetting, both of which reflect inefficient reasoning behaviors.

\begin{highlight}
    \textit{\textbf{Findings}: 
    \begin{itemize}[leftmargin=*,itemsep=-1pt,topsep=1pt]
        \item Task characteristics shape and favor distinct reasoning behaviors.
        \item Models achieving comparable accuracy may employ distinct reasoning strategies. 
        \item Overthinking and forgetting are prevalent across LRMs. 
    \end{itemize}}
\end{highlight}

\vspace{-.1in}
\subsection{Comparing Reasoning Structure: Standard LLMs vs. LRMs}\label{sec:lrmllm}
\vspace{-.1in}

While LRMs are optimized for multi-step reasoning, general-purpose LLMs can still reason effectively when prompted (e.g., with CoT). Their differing training objectives lead to distinct reasoning behaviors. 
Using \trees{}, we compare reasoning behaviors of two LRM-base pairs: (i)~Qwen2.5-32B with CoT vs. \qwq{}, and (ii)~DeepSeek-V3 with CoT vs. \deepseekr{}.

\begin{figure}
\vspace{-.5in}
    \centering
    \begin{subfigure}[t]{0.49\linewidth}
        \includegraphics[width=\linewidth]{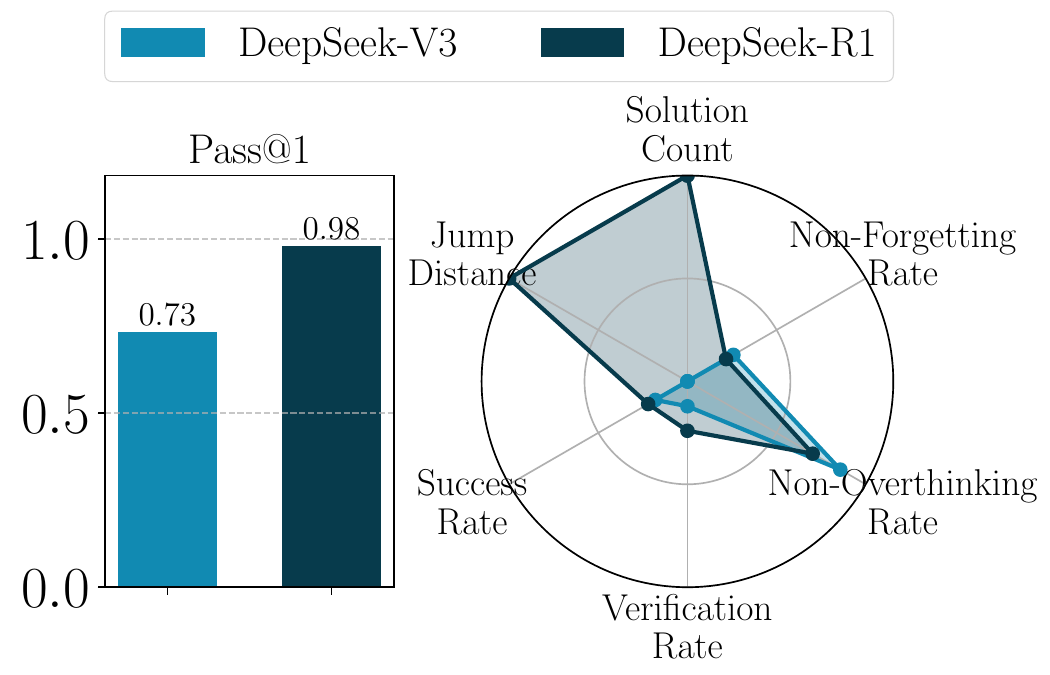}
        \vspace{-.2in}
    \caption{DeepSeek-V3 v.s. \deepseekr{}}
    \label{fig:lrm_llm_radar_game24_deepseek}
    \end{subfigure}
    \begin{subfigure}[t]{0.49\linewidth}
        \includegraphics[width=\linewidth]{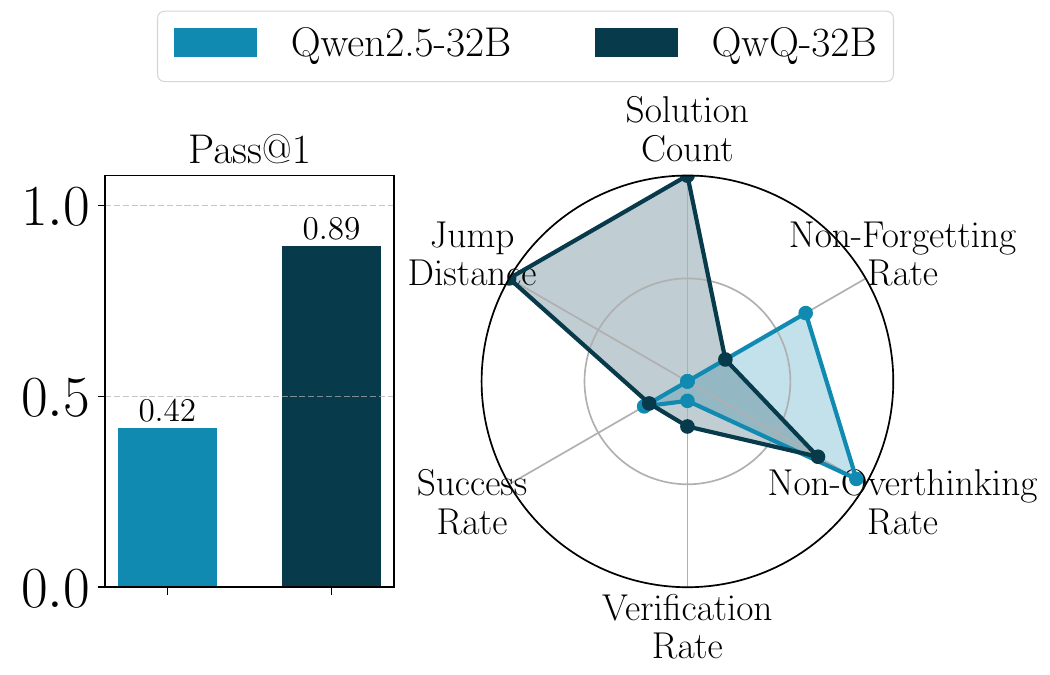}
        \vspace{-.2in}
    \caption{Qwen2.5-32B v.s. \qwq{}}
    \label{fig:lrm_llm_radar_game24_qwen}
    \end{subfigure}
    \vspace{-.1in}
    \caption{
        \textbf{Comparison of base LLMs (DeepSeek-V3, Qwen-2.5-32B) and their corresponding LRMs (DeepSeek-R1, QwQ-32B) on \passone{} and reasoning metrics for the Game of 24.}
        Despite similar \asuccess{}, LRMs achieve higher \passone{} by generating more and diverse solutions, as reflected in higher average solution counts and jump distances.
        LRMs also exhibit increased verification, overthinking, and forgetting behaviors.
    }
    \label{fig:lrmllm_simplified}
    \vspace{-.25in}
\end{figure}

\cref{fig:lrmllm_simplified} shows results on Game of 24. 
Compared to \citet{limit-of-rlvr}, which argue that RL favors high-reward paths with less exploration by showing that RL-trained LRMs outperform base models at low $k$ in pass$\char64 k$, but underperform at high $k$, our results show that LRMs outperform LLMs in \passone{}, but not by favoring high-reward paths. 
Instead, their success rates are similar, and gains stem from generating more solutions. 
In fact, LRMs also exhibit more exploratory reasoning, reflected in higher jump distances and more frequent shifts in approach.
Note that this does not contradict \citet{limit-of-rlvr}, as their analysis considers exploration across samples, while ours focuses on single-trace exploration. 
Lastly, LRMs show higher verification, overthinking and forgetting.

\begin{highlight}
    \textit{\textbf{Findings}: 
    \begin{itemize}[leftmargin=*]
        \item LRMs achieve higher \passone{} by generating more numerous and diverse solutions, despite not necessarily improving per-attempt accuracy.
        \item Compared to LLMs, LRMs demonstrate more deliberate reasoning behaviors, such as increased exploration and verification, but also suffer more from overthinking and forgetting.
    \end{itemize}
    }
\end{highlight}
The results for MATH-500, presented in~\Cref{app:lrmllm}, further support our findings.

\vspace{-.1in}
\subsection{Impact of Distillation on Reasoning Structure: Comparing Teacher and Distilled Models}\label{sec:distill}
\begin{wraptable}{r}{.5\linewidth}
    \vspace{-.3in}
        \caption{
        \textbf{Tree similarity (\asimilaritytree{}) and jump similarity (\asimilaritywalk{}) between each model and the teacher model.}
        Base: Qwen2.5-14B; Distilled: DeepSeek-R1-Distill-Qwen-14B; Teacher: \deepseekr{}.
        Both metrics improve after distillation, showing that distilled model more closely replicates the teacher's reasoning structure.
    }
    \label{tab:distill_simplified}
    \vspace{-.1in}
    \centering
    \resizebox{\linewidth}{!}{
    \begin{tabular}{ccccc}
    \toprule
    \multirow{3}{*}{\textbf{Reference Model}} & \multicolumn{4}{c}{\textbf{Metric}} \\
    \cmidrule(lr){2-5}
    & \multicolumn{2}{c}{\asimilaritytree{} (Teacher, $\cdot$) } & \multicolumn{2}{c}{\asimilaritywalk{} (Teacher, $\cdot$)} \\
    \cmidrule(lr){2-3}\cmidrule(lr){4-5}
    & Base & Distilled & Base & Distilled   \\ \midrule
    MATH-500 & .715 & \textbf{.728} & .771 & \textbf{.878} \\
    Game of 24 & .360 & \textbf{.426} & .873 & \textbf{.905}  \\
    \bottomrule
    \end{tabular}}
    \vspace{-.2in}
\end{wraptable}

Model distillation transfers the capabilities of large LRMs to smaller, more efficient models~\citep{deepseek-r1}, often preserving task performance. However, its effect on underlying reasoning structure and actions remains unclear.
To investigate this, we compare three model types (base, teacher, and distilled) at two scales: the 14B group uses Qwen-2.5-14B and \deepseekr{}-Distill-Qwen-14B, while the 32B group uses the 32B counterparts of the same models, with \deepseekr{} as the teacher in both cases.


We report the similarity to the teacher model before and after distillation for the 14B comparison group in~\Cref{tab:distill}; results for the 32B group and detailed per-metric comparisons for both groups are deferred to~\Cref{app:distill}.
The results show that the distilled model consistently moves closer to the teacher in both tree similarity (\asimilaritytree{}) and jump similarity (\asimilaritywalk{}).
Further analysis in~\Cref{app:distill} confirms that distilled models inherit reasoning behaviors from teacher LLMs, including broader exploration, verification, and backtracking, though success rates are not improved.

\begin{highlight}
    \textit{\textbf{Finding:} 
    Distilled models inherit reasoning behaviors from teacher models, 
    as evidenced by gains in both tree and jump similarities.
    }
\end{highlight}
In~\Cref{app:distill}, we further compare the distilled model (DeepSeek-R1-Distill-Qwen-32B) with the RL-trained model (QwQ-32B) as an initial exploration of how SFT and RL differ in their impact on reasoning behavior, and show that the RL-trained model appears to exhibit more deliberate reasoning behaviors, including increased exploration, verification, and overthinking.

\vspace{-.1in}
\subsection{Impact of Reasoning Examples on Reasoning Structure}\label{sec:reasoning_example}
\vspace{-.1in}

\begin{wrapfigure}{r}{.5\linewidth}
\vspace{-.2in}
    \centering
        \includegraphics[width=\linewidth]{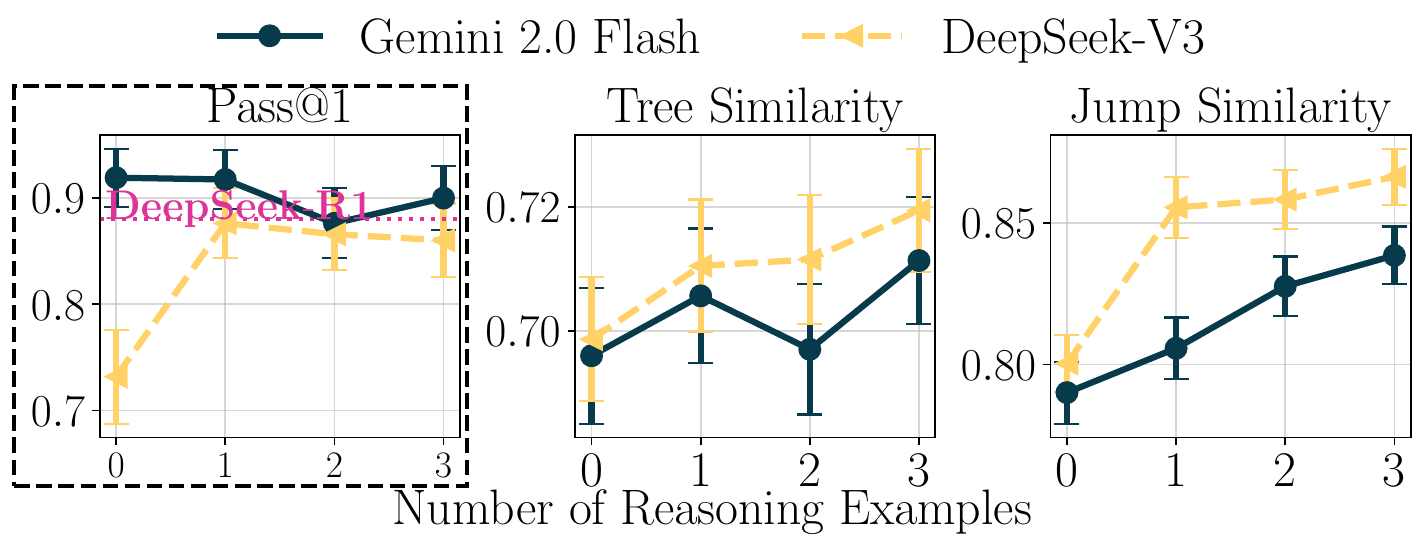}
        \vspace{-.25in}
    \caption{
       \textbf{Effect of the reasoning examples on reasoning behaviors (MATH-500).}
        Gemini 2.0 Flash and DeepSeek-V3 are prompted with \deepseekr{} examples.
        The dashed boxes indicate final accuracy for different number of in-context examples, while the remaining plots show tree and jump similarity to \deepseekr{}. 
    }
    \label{fig:reasoning_example_simplified}
    \vspace{-.1in}
\end{wrapfigure}

Although prior work~\citep{wei2022chain,agarwal2024many,zhang2025enhancing} has explored the use of reasoning examples in prompts to enhance LLM reasoning capabilities, how in-context examples reshape the reasoning behavior remains underexplored.
A natural question arises: How does the presence and number of examples affect reasoning characteristics? 
To investigate these questions, we vary the number of in-context reasoning examples  ($\set{0, 1, 2, 3}$) included in the prompt and analyze resulting changes in reasoning behavior.
We evaluate DeepSeek-V3 and Gemini 2.0 Flash, with the latter following prior work~\citep{agarwal2024many,zhang2025enhancing}. 
Since Gemini-family LRMs like \gemini{} do not expose reasoning traces, we use \deepseekr{}-generated examples from other samples within the same dataset. 

\cref{fig:reasoning_example_simplified} shows how \passone{}, tree similarity, and jump similarity vary with the number of in-context examples on MATH-500.
Accuracy does not consistently improve; in fact, Gemini 2.0 Flash even outperforms \deepseekr{} without demonstrations.
However, jump similarity rises steadily, indicating stronger imitation of LRM-style behaviors (e.g., verification, calculation, backtracking) with more examples.
In contrast, tree similarity shows no clear trend, suggesting limited impact on problem decomposition.
Results on Game of 24 (\Cref{app:reasoning_example}) show similar patterns.

\begin{highlight}
\textit{\textbf{Finding}: Increasing the number of in-context reasoning examples has a stronger and more consistent influence on reasoning actions (e.g., verification and backtracking) than on high-level problem decomposition strategies, which remain relatively invariant.}
\end{highlight}

\vspace{-.1in}
\subsection{Evolution of Reasoning Dynamics Under Reinforcement Learning}\label{sec:rl}
\begin{figure}[t]
\vspace{-.4in}
    \centering
    \includegraphics[width=\linewidth]{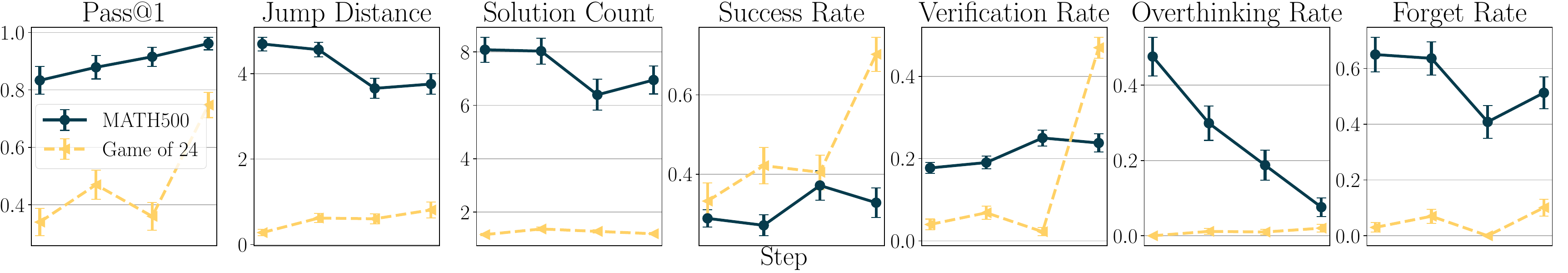}
    \vspace{-.3in}
    \caption{
        \textbf{Evolution of reasoning metrics during RL.}
        We use Qwen3-1.7B for the Game of 24 task and Qwen3-8B for the MATH500 task, applying DAPO~\citep{dapo} to enhance reasoning performance without supervised fine-tuning, with max sequence length set to 2048.
        RL encourages more exploratory reasoning on Game of 24, as indicated by increased jump distance, while promoting more exploitative behavior on MATH500, reflected by higher success rates and reduced jump distance.
        These results show that RL-induced improvements align with the inherent reasoning characteristics of each task.
        Note that our \passone{} computation also accounts for the correct formatting of \texttt{<answer></answer>} tags.
    }
    \label{fig:dynamics}
    \vspace{-.2in}
\end{figure}

\citet{deepseek-r1} demonstrates the effectiveness of RLVR in enhancing the reasoning capabilities of standard LLMs.
In this experiment, we use the \trees{} representation to visualize the evolution of models' reasoning behavior during RL.
We apply DAPO~\citep{dapo} on the MATH-500 and Game of 24 tasks, respectively. 
For MATH-500, we employ Qwen3-8B, and for Game of 24, Qwen3-1.7B.
Each model is evaluated at four checkpoints, corresponding to one-quarter intervals throughout training (i.e., 1/4, 2/4, 3/4, and 4/4 of the training process).

As discussed in \cref{sec:benchmark}, MATH-500 favors higher success rates, indicating stronger exploitation, whereas Game of 24 benefits from both increased jump distance and higher success rates, reflecting a need for balanced exploration and exploitation.
The results in \cref{fig:dynamics} show that RL progressively shapes the model’s reasoning dynamics to match these task-specific requirements: promoting more exploitative reasoning (i.e., higher success rates) on MATH500 and both greater exploration (i.e., higher jump distance) and higher success rates on Game of 24.

\begin{highlight}
\textit{\textbf{Finding}: 
    Reinforcement learning progressively aligns a model's reasoning behavior with the demands of the target task.
}
\end{highlight}

\vspace{-.1in}
\section{Enhancing LLM Reasoning with \trees{}}\label{sec:improve}
\vspace{-.1in}

Beyond using \trees{} to dissect LLM reasoning and compare behaviors across different LRMs, tasks, and settings, we further demonstrate in this section that \trees{} can also be leveraged to enhance LLM reasoning.

A key advantage of \trees{} is that they enable selecting outputs with desired reasoning characteristics, as measured by the six metrics, without requiring ground-truth solutions. 
We focus on the Game of 24 because its difficulty depends less on an LLM's raw capability (e.g., success rate) and more on its reasoning behavior (e.g., exploration), which can be more effectively improved by enhancing reasoning patterns. 
As established in \cref{sec:benchmark}, Game of 24 benefit from more explorative reasoning. 
Therefore, our experiments in this section center on this task instead of MATH500.
All reported results are averaged over three random seeds.

\vspace{-.1in}
\subsection{Improving Reasoning via Best-of-N Selection with \trees{}}\label{sec:bon}
\vspace{-.1in}

Accordingly, we consider two strategies that utilize \trees{} for ensembling the result: 
(i) apply a Best-of-N (BoN) strategy: generate multiple responses and use \trees{} to select the one exhibiting the highest exploration (i.e., the largest jump distance, \ajd{}), 
(ii) use weighted majority vote, where each response is assigned a weight proportional to the jump distance associated with that response.

We consider \qwq{} and \phif{} in this experiment since their performance is relatively limited and has more room for improvement compared to other LRMs, as shown in \cref{sec:benchmark}. 
We set $N=3$, which means that for each prompt we generate 3 response, and BoN with \trees{} use \trees{} to choose the one with highest jump distance, while weighted majority vote with \trees{} assigns weights proportional to the jump distance of each response.  
As a baseline, we use majority vote to ensemble the three responses.
The comparison between the majority vote, BoN with \trees{}, and weighted majority vote with \trees{} are shown in \cref{tab:bon}, demonstrating BoN with \trees{} outperforms the others. 
In \cref{app:improve}, we extend our analysis to two additional datasets requiring different reasoning characteristics, further supporting our findings. 

\begin{table}[t]
\vspace{-.4in}
\begin{minipage}{.54\linewidth}
        \caption{
\textbf{Performance of the Majority Vote (MV), weighted MV with \trees{} and Best-of-N (BoN) with \trees{} on Game of 24 using \qwq{} and \phif{}.}
Across both \passone{} and \ajd{}, BoN with \trees{} performs the best. 
    }
    \label{tab:bon}
    \vspace{-.1in}
\resizebox{\linewidth}{!}{
    \begin{tabular}{cccc}
    \toprule
        \textbf{Model} & \textbf{Method} & \textbf{\passone{}} & \textbf{\ajd{}}  \\
        \midrule
         \multirow{3}{*}{\qwq{}} & MV & 0.76 & 4.20 \\ 
         & Weighted MV w. \trees{} & 0.80 & 5.09 \\ 
         & BoN w. \trees{} & \textbf{0.82} & \textbf{5.70}\\ 
         \midrule
         \multirow{3}{*}{\phif{}} & MV & 0.77 & 3.32 \\ 
         & Weighted MV w. \trees{} & 0.81 & 4.02 \\ 
         & BoN w. \trees{} & \textbf{0.84} & \textbf{5.53}  \\
         \bottomrule
    \end{tabular}}
\end{minipage}\hfill
\begin{minipage}{.43\linewidth}
       \caption{
    \textbf{Comparison of performance between the default prompt and the prompt selected by \trees{} from four candidate prompts on Game of 24.}
    The \trees{}-chosen prompt yields better performance.
    }
    \label{tab:prompt}
    \vspace{-.1in}
\resizebox{\linewidth}{!}{
    \begin{tabular}{cccc}
    \toprule
        \textbf{Model} & \textbf{Prompt} & \textbf{\passone{}} & \textbf{\ajd{}}  \\
        \midrule
         \multirow{2}{*}{\qwq{}} & Default & 0.73 & 4.09 \\ 
         & \trees{}-chosen & \textbf{0.78} & \textbf{4.28} \\ 
         \midrule
         \multirow{2}{*}{\phif{}} & Default & 0.76 & 3.42 \\ 
         & \trees{}-chosen & \textbf{0.82} & \textbf{3.98} \\
         \bottomrule
    \end{tabular}}
\end{minipage}%
\vspace{-.2in}
\end{table}

\vspace{-.1in}
\subsection{Prompt Selection with \trees{}}\label{sec:prompt}
Another natural application of \trees{}'s comprehensive measurement is prompt selection.
There are several ways to leverage \trees{} for this purpose.
For efficient reasoning, one can design multiple prompts, test them on a development set, and use \trees{} to evaluate the generated responses' reasoning behavior. 
The prompt that achieves a higher success rate and lower overthinking rate can be selected.
For search-heavy tasks, such as maze solving or the Game of 24, \trees{} can help identify the prompt that provides the best exploration-exploitation tradeoff.

In this experiment, we focus on prompt selection for improving exploration in the Game of 24. 
Similar to the previous experiment, we consider \qwq{} and \phif{}.
We design four prompts (see~\Cref{app:prompt} for details) intended to encourage broader exploration and use \trees{} to select the one yielding the highest jump distance. 
\Cref{tab:prompt} shows that without ground-truth labels, prompt selection guided by \trees{} improves the performance of both models on the Game of 24.


% ...


\vspace{-.1in}
\section{Conclusion}\label{sec:discussion}
\vspace{-.1in}
In this work, we propose \trees{}, a tree-jump representation of reasoning traces, with the tree layer capturing the hierarchical structure of partial solutions and their dependencies, and the jump layer, tracing the sequential execution of reasoning steps. 
Using our proposed \treesextractor{} method to extract reasoning traces into \trees{}, the resulting representations allows us to quantitatively analyze reasoning behaviors such as the exploration-exploitation trade-off, overthinking, and forgetting, which enables comparison of reasoning processes themselves beyond final accuracy across models, tasks, and settings.
Because these metrics reveal potential weaknesses in reasoning regardless of final performance, \trees{} shows what needs to be improved during training and helps decide which inference strategy works best for a task.
Beyond serving as an analytic tool that indirectly guides LRM development, \trees{} can also directly enhance reasoning performance, as demonstrated through two applications: Best-of-N response selection and prompt selection.

\vspace{-.1in}
\paragraph{Limitations \& Future Work.}
Despite these contributions, several limitations remain.
First, cost and efficiency are a challenge: \treesextractor{} requires a separate, capable LLM to process each reasoning trace, which is computationally expensive and slow, limiting large-scale use such as real-time feedback during training.
Overcoming this limitation would make it possible to analyze the dynamics of how reasoning evolves during training, including under outcome- vs process-supervised reinforcement learning and supervised fine-tuning.
In addition, incorporating \trees{}-derived signals into reward modeling represents another promising direction.
Second, when comparing reasoning traces, our current tree and jump similarity metric capture only logical structure and action transition distributions. 
This simplifies computation but can mask important differences: two models may yield identically structured trees yet differ semantically, and a perfect jump similarity score (i.e., 1.0) may still hide distinct temporal behaviors.
For instance, one model might perform derivation first and verify at the end, whereas another may interleave computation and verification throughout.
Incorporating semantic similarity and temporal dynamics is therefore interesting future work. 
Finally, the method still requires defining partial solutions for each task, necessitating task-specific prompting; automating this adaptation would greatly improve usability.





\section*{Acknowledgment}
Kangwook Lee is supported by NSF Award DMS-2023239, NSF CAREER Award CCF-2339978, Amazon Research Award, and a grant from FuriosaAI.

\bibliography{reference}
\bibliographystyle{iclr2026_conference}

\appendix
\newpage
\clearpage
{\LARGE \textbf{Appendix}} \par 
\startcontents[sections]
\printcontents[sections]{ }{1}{}
\clearpage




\section{Extended Sec.~\ref{sec:tree}: \trees{}}
\label{app:tree}

In this section, we follow the structure of \cref{sec:tree}, but provide additional details such as the prompts and formal mathematical definitions of the metrics.
To make the discussion self-contained, we also repeat some key content from the main text.

\begin{figure}[htbp]
    \centering
    \begin{subfigure}[b]{\linewidth}
        \includegraphics[width=\linewidth]{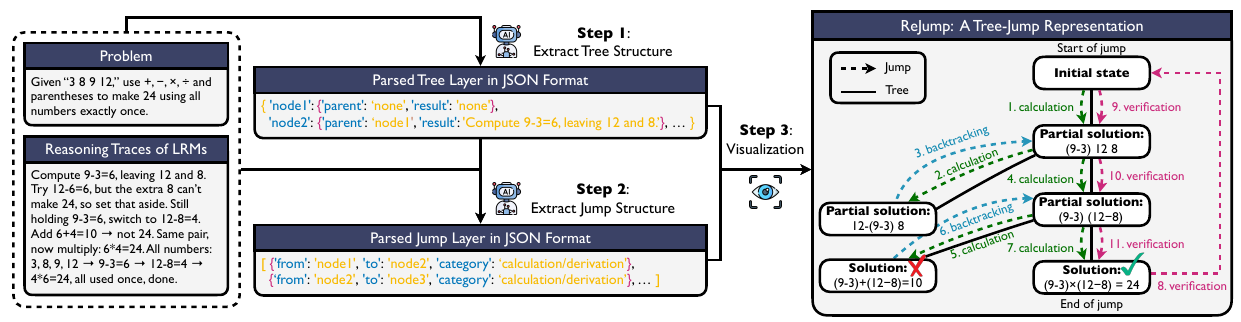}
        \caption{
            \trees{} for a Game-of-24 instance.
            Partial solution here is defined as an intermediate arithmetic expression. 
        }
        \label{fig:tree_demo_24}
    \end{subfigure}\hfill
    \begin{subfigure}[b]{\linewidth}
        \includegraphics[width=\linewidth]{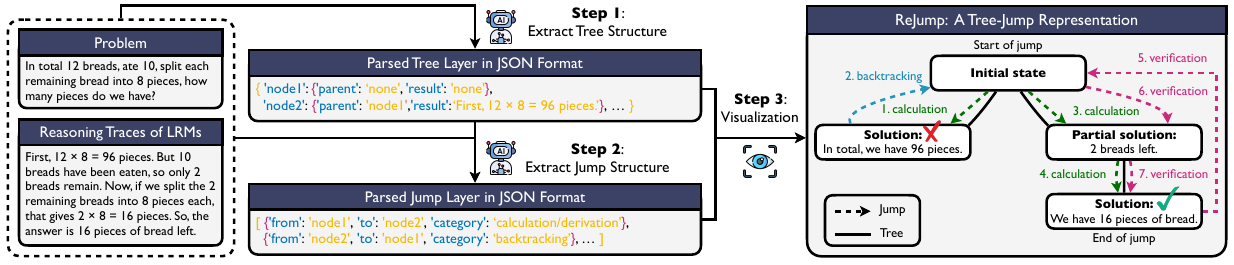}
        \caption{
            \trees{} for a math word problem. 
            Partial solution here is defined as an intermediate computed results.
        }
        \label{fig:tree_demo_math}
    \end{subfigure}
    \caption{
        \textbf{Illustration of \trees{} representation of reasoning traces for two different tasks. }
        These examples are crafted for demonstration purposes.
        Nodes represent partial solutions, and tree edges indicate prerequisite relationships. The dashed jump traces the solver’s reasoning trajectory, with transitions labeled by action type: calculation/derivation, verification, or backtracking, highlighted in different colors.
    }
    \label{fig:tree_demo}
\end{figure}
\begin{itemize}[leftmargin=*]
    \item \textbf{Tree layer (structure)}: We define a tree $T = (V,E)$, where $V = \set{v_i}_{v=0}^{|V|}$ is the set of nodes and $E$ the set of edges.
    Following \citet{tot}, each node $v\in V$ represents a partial solution, and an edge $e\in E$ indicates that the parent's partial solution is a direct prerequisite for the child's. 
    Let $\leaf{}(T) \subset V$ denote the set of leaf nodes in $T$,
    each representing a single solution attempt, either a completed solution under one approach or a dead-end where the approach failed to yield a correct or full solution.
    Among these, we define $\corrleaf(T) = \set{v \in \leaf{}(T) : v \text{ encodes a fully correct solution}}$ as the subset of leaves that represent correct solutions.
    \item  \textbf{Jump layer (action)}: 
    Let $\vi = (i_0, i_1, \ldots, i_K)$ denote the sequence of reasoning steps, where $i_k$ refers to the index of $k$-th visited node in the tree. 
    The jump starts at $v_{i_0}$ (the root) and ends at $v_{i_K}$ (the final solution).
    Each transition between consecutive steps $(i_k, i_{k+1})$ is labeled with an action type $\phi_k \in \set{\texttt{calc}, \texttt{verify}, \texttt{backtrack}}$, where $k=1,\ldots, K-1$.
    Here, \texttt{calc} refers to generating an intermediate step via calculation or derivation; \texttt{verify} denotes revisiting a node to check its correctness; and \texttt{backtrack} indicates returning to a previous node to explore an alternative reasoning path.
    A jump is the pair $W = (\vi, \vphi)$, fully specifying both the node sequence and how the solver moves through the tree. 
    A \textit{derived solution step} is any step in the jump that reaches a leaf node via a \texttt{calc} transition.  
    Even if the leaf has been visited before, it still counts as a derived step if reached via \texttt{calc}; by contrast, visits for verification do not count.
    We mathematically define the sequence of such steps as $\newleaf(T, W) = (i_{k_1}, \ldots, i_{k_M})$, where each $v_{i_{k_j}} \in \leaf(T)$ and the corresponding transition is $\phi_{k_j} = \texttt{calc}$, for all $j = 1, \ldots, M$.  
    Among these, we further define the \textit{correct derived solution steps} as $\newcorrleaf(T,W)$, a subsequence of $\newleaf(T,W)$, consisting of indices $i_{k_j}$ such that $v_{i_{k_j}} \in \corrleaf(T)$.
\end{itemize}

\paragraph{Notations.} 
Define $\consec(\cdot)$ as an operator that takes a sequence as input and returns the set of all consecutive pairs; that is, for a sequence $(x_1, \ldots, x_M)$, $\consec(x_1, \ldots, x_M) = \set{(x_1, x_2), \ldots, (x_{M-1}, x_M)}$.
Let $\oset{\cdot}$ be the operator that convert a sequence into a set.
For any sequence $\vi = (i_1, \ldots, i_n)$, we denote its $j$-th element (1-based indexing) by $\vi[j]$.
When consider $N$ generated reasonings and their corresponding \trees{}, we use subscript $(n)$ to denotes the $n$-th reasoning instance. 

Next, we present more rigorous definitions of the evaluation metrics.

\subsection{Evaluation Metrics}\label{app:eval_metrics}
This tool enables analysis of LLM reasoning behaviors, including solution diversity, exploration-exploitation trade-off, effectiveness in identifying correct paths, frequency of overthinking, forgetting, and verification.
These aspects are quantified using the following metrics, computed across all reasonings and their corresponding \trees{}s within a task.

\paragraph{Solution Count (\asolcount{}).} 
This metric quantifies the model's ability to discover diverse solution attempts, measured by the number of leaf nodes in the reasoning tree. 
We define $\solcount{} (\set{T}) = |\leaf{}(T)|$ as the total number of leaf nodes representing distinct solutions within a single tree $T$.
To assess the average performance over $N$ reasoning instances for a given task, we calculate the Average Solution Count as the mean number of unique solutions found across all instances: \asolcount{}$ (\set{T\nth}_{n=1}^N) = \sum_{n=1}^N \solcount(\set{T\nth})/N$.

\paragraph{Jump Distance (\ajd{}).} 
This metric captures the exploration-exploitation tradeoff of the reasonings by averaging how far the reasoning ``jumps'' between newly visited leaf nodes (see \cref{fig:jd}).
Define $d(u,v)$ as the number of edges on the path between nodes $u,v \in V$. 
The jump distance of a single jump trace is \ajd$(\set{(T,W)}) = \frac{1}{|\consec(\newleaf(T,W))|} \sum_{(i_j, i_l) \in \consec(\newleaf(T,W))} d(v_{i_j}, v_{i_l})$, and the task-level average is 
\ajd$(\set{(T\nth, W\nth)}_{n=1}^{N}) = \frac{1}{N} \sum_{n=1}^N $\ajd$ (\set{(T\nth, W\nth)})$.

\paragraph{Success Rate (\asuccess{}).} 
The metric measures how frequently a reasoning path produces a correct solution. 
For a single reasoning with tree $T$ and jump $W$, the success rate is computed as $\success{}(\set{(T,W)}) = |\newcorrleaf(T,W)| /|\newleaf(T,W)|$, i.e., the proportion of newly visited leaf nodes that are correct solutions.
The overall average is computed across all $N$ reasoning instances: \asuccess{}$(\set{(T\nth,W\nth)}_{n=1}^N) = \frac{1}{N} \sum_{n=1}^N \success{} (\set{(T\nth, W\nth)})$.

\paragraph{Verification Rate (\avalid{}).} 
This metric quantifies how frequently the model invokes verification steps during its reasoning process.
For a given reasoning instance with jump $W=(\vi, \vphi)$, the verification rate is computed as the number of \texttt{verify} transitions divided by the total number of steps in the jump: $\valid (\set{W}) = \sum_{\phi_k \in \oset{\vphi}} \sI \set{\phi_k = \texttt{verify} }/ (K-1)$.
We report the average verification rate across all $N$ reasoning instances.
\avalid{}$(\set{W\nth}_{n=1}^N) = \sum_{n=1}^N\valid (\set{W\nth})/N$.

\paragraph{Overthinking Rate (\aoverthinking{}).} 
This metric quantifies the extent of unnecessary exploration after a correct solution has already been found.
For a given reasoning instance with tree $T$ and jump $W$, let $k_0^\star$ denote the first index in $\newcorrleaf(T,W)$, the step at which a correct leaf is first reached. 
The overthinking rate is defined as the fraction of newly visited leaf nodes that appear \textit{after} this first correct solution:  $\overthinking(\set{(T,W)}) =  |\set{i_k \in \oset{\newleaf{}(T,W)}: k > k^\star_0}|/|\newleaf(T,W)|$.
In other words, it measures how many additional leaf nodes are explored via \texttt{calc} transitions after a correct solution has been identified.
The task-level overthinking rate is then given by the average over all $N$ reasoning instances: \aoverthinking{}$(\set{(T\nth,W\nth)}_{n=1}^N) = \frac{1}{N} \sum_{n=1}^N\overthinking(\set{(T\nth,W\nth)})$.

\begin{figure}
\centering
        \includegraphics[width=.5\linewidth]{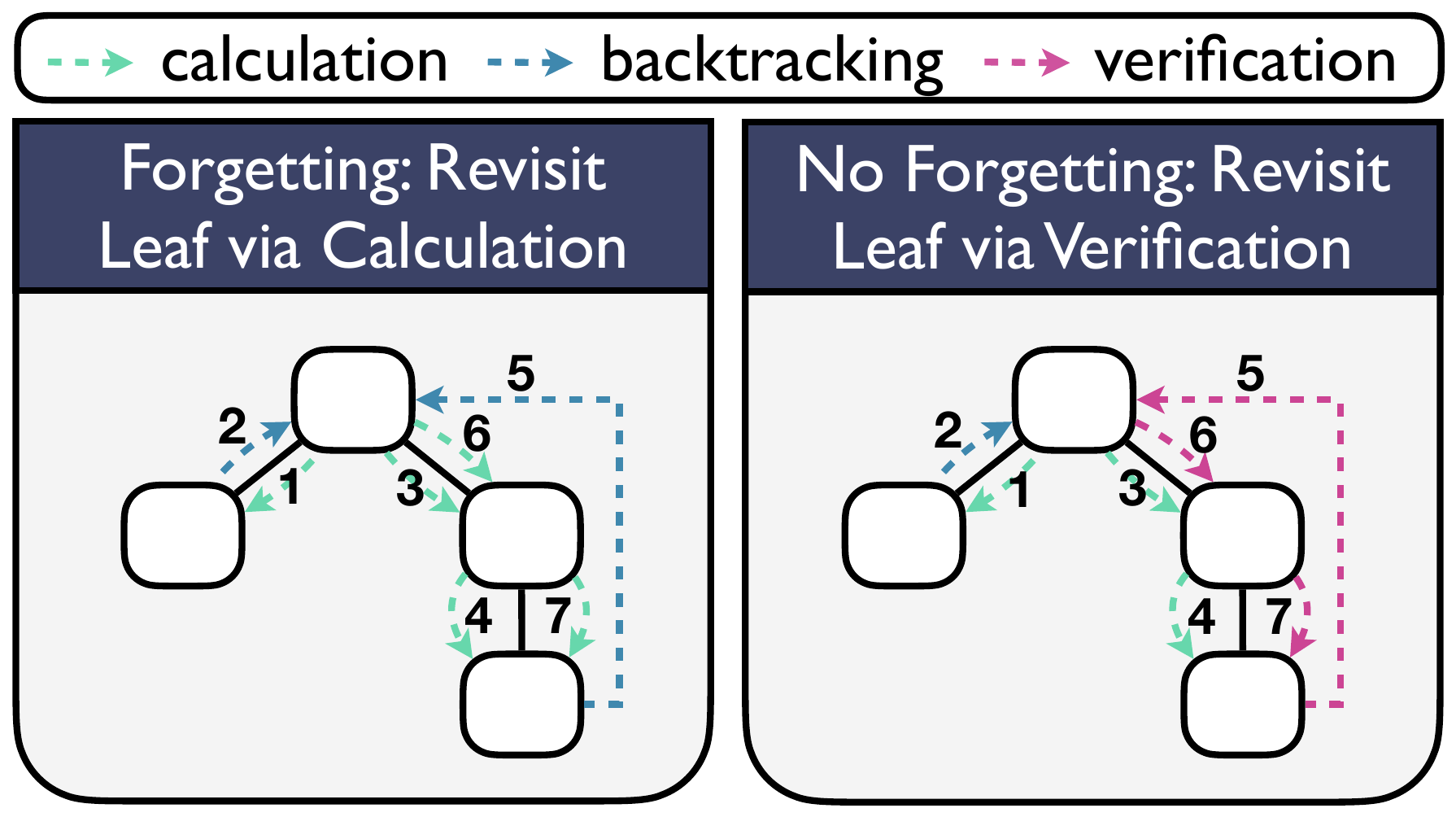}
        \caption{
            Illustration of forgetting vs. no forgetting in \trees{}.
            Revisiting an already-seen node via \texttt{calc} indicates forgetting, while revisiting via \texttt{verify} does not.
        }
        \label{fig:forget}
    \end{figure}

\paragraph{Forgetting Rate (\aforget{}).} 
This metric tracks how often the model forgets its earlier steps and recomputes a previously derived result.
Specifically, forgetting is flagged when a previously visited leaf node is revisited via a \texttt{calc} transition (see \cref{fig:forget}).
We define a binary indicator for each reasoning instance as
$\forget(T, W) = 1 - \prod_{m=2}^M \left( \prod_{j=1}^{m-1} \sI\set{\newleaf[j] \ne \newleaf[m]} \right)$,
which returns 1 if any earlier leaf is re-entered, and 0 otherwise.
The forgetting rate is then reported as the proportion of instances where forgetting occurred: \aforget{}$(\set{(T\nth,W\nth)}_{n=1}^N) = \frac{1}{N} \sum_{n=1}^N \forget(T\nth, W\nth)$.

\subsection{Redundancy Analysis of Proposed Metrics}
\label{app:redundancy}

To assess whether the six proposed metrics capture complementary aspects of reasoning behavior, we conduct an information-theoretic redundancy analysis.
For each metric $M$, we compute its redundancy as
\[
\mathrm{Redundancy}(M) = \frac{I(M;\text{others})}{H(M)},
\]
where $H(M)$ is the entropy of $M$ and $I(M;\text{others})$ is the mutual information between $M$ and the remaining metrics. 
Lower values indicate that the metric contains information not recoverable from the others.
Redundancy scores for MATH-500 and Game of 24 are reported in \cref{tab:redundancy_all}.

\begin{table}[t]
\centering
\caption{Redundancy scores of the six metrics on MATH-500 and Game of 24. 
Lower values indicate less dependence on other metrics.}
\label{tab:redundancy_all}
\begin{tabular}{lcccccc}
\toprule
Dataset & \asolcount{} & \ajd{} & \asuccess{} & \avalid{} & \aoverthinking{} & \aforget{} \\
\midrule
MATH-500    & 0.789 & 0.624 & 0.437 & 0.105 & 0.944 & 0.277 \\
Game of 24  & 0.731 & 0.761 & 0.882 & 0.687 & 0.840 & 0.840 \\
\bottomrule
\end{tabular}
\end{table}

\section{Extended \cref{sec:extractor}: \treesextractor{}}\label{app:extractor}
In addition to the illustrative example on a math word problem, we include one more example showcasing the construction and visualization of \trees{}, with both examples shown in \cref{fig:tree_demo}.

\subsection{LLM Prompts for \trees{} Construction}\label{app:prompts}
In this section, we present the prompt used by the LLM to parse results across all experiments.

We use different prompts for the two datasets.
\cref{lst:tree_math} and \cref{lst:walk_math} show the prompts used to extract the tree and jump from the generated reasoning for MATH-500, respectively, while \cref{lst:tree_game24} and \cref{lst:walk_game24} show the corresponding prompts for Game of 24.

\begin{lstlisting}[caption={Prompt for extracting a tree from the reasoning trace in JSON format for MATH-500.}, label={lst:tree_math}]
def get_tree_prompt_math(input_str, output_str):
    return f"""
Your task is to analyze a detailed thinking process for solving a math problem (provided below) and convert it into a reasoning tree. This tree must represent the **chronological flow of solving substantive, mathematically well-posed subproblems or distinct attempts**, starting from an initial state and culminating in answering the original question.

Represent this structure as a **single JSON object** where keys are unique node IDs (e.g., "node1", "node2") and values are node objects detailing each state or subproblem attempt.

**Core Principles for Tree Generation:**

* **Chronological Flow & Dependency:** The tree follows the order of substantive steps/attempts in the reasoning. Parent links indicate the preceding step whose `Result` provides necessary mathematical input.
  **BRANCHING AND SUBSTEP RULE:** 
    - Create a new branch **if and only if** the reasoning process explicitly abandons or gives up on a previous approach and then starts a new, distinct solution plan. In other words, a new branch is created always and only when the previous line of reasoning is abandoned and a fundamentally different method is attempted. The new branch should start from the most recent shared node. Even if the solver does not immediately abandon the previous approach, we still consider it an Abandoned Attempt Node and mark it with [Path abandoned] if a different method is initiated that departs from the original direction.
    - Importantly, whenever a new branch is created, the leaf node where the previous method ended must be explicitly marked with [Path abandoned].
    - Conversely, if the current node is marked with [Path abandoned], a new branch must always be created.
    - Importantly, for all subproblems or calculations within a single uninterrupted attempt, even if subcalculations are mathematically independent, represent these steps sequentially in the order they are performed in the reasoning: each node's parent must be the immediately preceding node within that attempt.  
    That is, substeps within any one attempt always form a single chain.
* **Substantive, Well-Posed Steps Only:** Nodes must represent **major** intermediate calculations or logical deductions constituting a clear, self-contained mathematical task (like a homework sub-problem). **Aggressively filter out** setup actions, strategy descriptions, narrative, verification, and trivial calculations/manipulations. Minor algebraic steps within a larger logical step must be grouped.
* **Include Failed Attempts:** Represent distinct, substantive calculation or derivation attempts that were **explicitly abandoned** in the reasoning as separate nodes in the chronological flow. **Do not filter these out.**
* **Focus on Mathematical Task:** Intermediate `Problem` fields must state a clear mathematical objective based on **all necessary given mathematical conditions and inputs**, avoiding descriptions of the reasoner's process or assumptions *within the Problem text*.
* **Special Final Node:** The node performing the last calculation for the final answer uses the original problem statement as its `Problem`.

**Node Object Structure:**
Each node object must contain: `Problem`, `parent`, `Result`.

1.  **`Problem` (String): Defines the specific mathematical task for this node.**
    * **`node1` (Root):** Must be exactly "Initial State".
    * **Intermediate Nodes (`node2` to `node(N-1)`):** Formulates a **clear, mathematically well-posed, and self-contained task representing a substantive step or distinct attempt.** Each node represents achieving a distinct intermediate objective through calculation or deduction.
        * **Format:** Start with "Given..." listing **all essential mathematical conditions, constraints, equations, and input values** (often from parent `Result` or established context like 'point P is on curve C') needed to define and solve *this specific task*. End with a specific mathematical question/instruction (e.g., "Calculate...", "Solve...", "Derive...").
        * **Content:** The formulation must focus purely on the **mathematical task**, making it **understandable and solvable in isolation** like a homework sub-problem, using only the provided "Given..." information and general mathematical knowledge. **CRITICAL RULE:** The `Problem` text **must not** include descriptions of the reasoner's strategy, assumptions, or procedural instructions reflecting the reasoning flow. State only the necessary mathematical conditions and the objective. The task must be **substantive**. **CRITICAL FILTERING RULE:** **DO NOT** create separate nodes for individual algebraic manipulations... [rest of filtering rule stays the same - GROUP minor operations]. Also filter out narrative, setup, verification. No meta-tags or node ID references.
    * **`nodeN` (Final Calculation Node):** **This node represents the very last calculation step that produces the final answer.** Its `Problem` field **must contain the verbatim Original Problem Statement.**

2.  **`parent` (String): Identifies the immediately preceding substantive step providing necessary input.**
    * **`node1`:** Must be "none".
    * **Other Nodes (`node2` to `nodeN`):** Must be the ID of the node whose `Result` provides the direct mathematical prerequisite for the task in the current node's `Problem`. (For abandoned attempts, the parent is the node preceding the attempt).

3.  **`Result` (String): Records the mathematical outcome of completing the task.**
    * **`node1`:** "Original problem statement provided as context." (or similar).
    * **Intermediate Nodes (`node2` to `node(N-1)`):** The direct mathematical outcome of achieving the task defined in `Problem`. Summarizes the result of grouped operations.
    * **Abandoned Attempt Nodes:** Must state any partial outcome and explicitly end with "[Path abandoned]".
    * **`nodeN` (Final Calculation Node):** Must be the **final answer** to the Original Problem Statement.

**Instructions for Analysis:**
1.  **Inputs:** Use the "Original Problem Statement" and "Input Reasoning Process".
2.  **Identify & Filter Steps:** Read the reasoning chronologically. Identify **major** calculation phases, key logical deductions, or distinct attempts. **Crucially, ensure that distinct, substantive attempts explicitly marked as abandoned in the reasoning are identified and *not* filtered out.** Apply the **CRITICAL FILTERING and GROUPING RULES** aggressively: Group sequences of trivial algebraic steps into the single larger objective they serve. Filter out non-mathematical content, setup, strategy descriptions/assumptions-as-actions, and verification. Only create nodes for the remaining substantive steps and distinct abandoned attempts.
3.  **Create Nodes Sequentially:**
    * Create `node1`.
    * For each identified **substantive step/objective/attempt** *before* the final answer calculation: Create the corresponding intermediate node (`node2`, `node3`, ...). Determine `parent`. Formulate the `Problem` strictly according to Rule 1 (well-posed, self-contained task including **all necessary conditions/constraints**, no process descriptions). Record `Result`. Link abandoned attempt nodes chronologically.
    * For the **final calculation step**: Create `nodeN`. Determine `parent`. Set `Problem` to verbatim Original Problem Statement. Set `Result` to final answer.
4.  **Formatting:** Use LaTeX (`$...$`) for all math notation.
5.  **Output:** Produce a single JSON object.

---
**BEGIN ORIGINAL PROBLEM STATEMENT**
---
{input_str}
---
**END ORIGINAL PROBLEM STATEMENT**
---

---
**BEGIN INPUT REASONING PROCESS**
---
{output_str}
---
**END INPUT REASONING PROCESS**
---

Generate the JSON output based on these instructions.
    """

# After obtaining the tree, use a separate prompt to evaluate the correctness of each leaf node for refining the tree.
def get_result_parsing_and_comparison_prompt(result_string, ground_truth_string):
    return f"""You are an expert AI assistant. Your task is to analyze a 'Result' string from a mathematical reasoning step and compare its final numerical answer to a 'Ground Truth' value.

Instructions:
1.  Extract the final numerical value(s) from the 'Result' string. 
    - If multiple numbers are present, focus on the one that seems to be the conclusive answer of that step.
    - Handle approximations (e.g., "approx 46.0", "is about 3.14").
    - If the result explicitly states abandonment (e.g., "[Path abandoned]"), extract the numerical value derived *before* abandonment, if any. If no clear numerical value was derived, use "N/A" for the parsed value.
    - If no specific numerical answer can be clearly identified, use "N/A" for the parsed value.

2.  Compare the extracted numerical value with the 'Ground Truth' value.
    - The comparison should determine if they are essentially the same, considering potential minor differences in formatting or precision (e.g., "46" vs "46.0", "1.03" vs "1.035" if context implies rounding).
    - If the parsed value is "N/A", the comparison result should be "NOT_APPLICABLE".
    - If the ground truth is empty or clearly not a comparable numerical value, and the parsed value is numerical, consider it a "MISMATCH" unless specified otherwise.

3.  Output a single JSON object with two keys:
    -   `"parsed_value"`: The extracted numerical value as a string (e.g., "46", "3.14", "N/A").
    -   `"match_status"`: A string indicating the comparison result. Must be one of: "MATCH", "MISMATCH", "NOT_APPLICABLE".

Example:
Result string: "Using the approximations, $tan x^\circ \\approx \\frac{{1.3270 + 6.3138}}{{1.3270 \\times 6.3138 - 1}} \\approx \\frac{{7.6408}}{{8.381 - 1}} \\approx \\frac{{7.6408}}{{7.381}} \\approx 1.0355$. This implies $x \\approx arctan(1.0355) \\approx 46.0^\circ$. [Path abandoned]"
Ground Truth string: "46"
Expected JSON Output: {{"parsed_value": "46.0", "match_status": "MATCH"}}

Result string: "The answer is $y=3$."
Ground Truth string: "3.0"
Expected JSON Output: {{"parsed_value": "3", "match_status": "MATCH"}}

Result string: "The calculation leads to $10/2 = 5$. However, this path is incorrect."
Ground Truth string: "7"
Expected JSON Output: {{"parsed_value": "5", "match_status": "MISMATCH"}}

Result string: "[Path abandoned] No value obtained."
Ground Truth string: "10"
Expected JSON Output: {{"parsed_value": "N/A", "match_status": "NOT_APPLICABLE"}}

---
Result string to analyze:
{result_string}

Ground Truth value:
{ground_truth_string}
---

JSON Output:"""
\end{lstlisting}

\begin{lstlisting}[caption={Prompt used to extract jump from reasoning as a JSON structure for MATH-500.}, label={lst:walk_math}]
def get_jump_prompt(input_str, output_str, tree_json):
    return f"""
You are an AI assistant specialized in analyzing mathematical reasoning processes. Your task is to trace the provided reasoning text against a structured reasoning tree and generate a "walk" representing the trajectory of the thought process.

**Inputs:**

1.  **Problem Description:**
    ```
    {input_str}
    ```
2.  **Reasoning Text:** A step-by-step textual explanation of how the problem was solved, including potential errors, corrections, explorations of different paths, and verifications.
    ```text
    {output_str}
    ```
3.  **Reasoning Tree:** A JSON object representing the structured steps and dependencies of the solution(s). Each key is a node ID, and the value contains information about that step, including its parent node and specifically a "Problem" field describing the task of that node.
    ```json
    {tree_json}
    ```

**Task:**

Analyze the `Reasoning Text` to determine the sequence in which the solver mentally visited or considered the steps represented by the nodes in the `Reasoning Tree`. Identify the transitions between these nodes and categorize each transition. **Crucially, for verification steps, visiting a node X implies the text shows evidence of re-doing the specific task described in the "Problem" field of node X.**

**Output Format:**

Generate a JSON list of dictionaries, where each dictionary represents a single step in the reasoning walk. Each dictionary must have the following keys:

* `from`: The ID (string) of the node the reasoning is moving *from*.
* `to`: The ID (string) of the node the reasoning is moving *to*.
* `category`: A string indicating the type of transition. Must be one of:
    * `calculation/derivation`: Represents forward progress in the reasoning, moving from one step to the next logical step (often parent to child in the tree) to derive new information or explore a solution path.
    * `backtracking`: Represents abandoning a current line of thought or calculation (often because it's incorrect, inefficient, or a dead end) and returning to a previous state (node) to try a different approach. This is typically a move from a node to one of its ancestors (not necessarily the direct parent).
    * `verification`: Represents checking or confirming a result or step **by re-doing the work associated with previous nodes**. This is determined based on the text:
        * **Specific Re-work:** If the text explicitly describes actions that precisely match the **problem description** defined within an intermediate node (e.g., node X) as part of checking a later result (node Z), trace the path reflecting that specific re-work (e.g., Z -> X -> Z). This requires clear evidence in the text of **re-solving the problem defined in node X**.
        * **General Check:** If the text indicates verification of a result (node Z) but ***does not*** show actions matching the specific **problem description** of any intermediate node, interpret this as checking consistency with the initial problem statement/conditions (node 1). Represent this path as Z -> 1 -> Z. ***Note: Simply using a formula or result from a previous node (e.g., node X) without showing the steps to re-solve the problem defined in node X does NOT count as re-doing the work of node X.***

**Instructions:**

1.  Read the `Reasoning Text` carefully, paying attention to the flow, changes in direction, calculations, statements of intent (e.g., "Let me try...", "No, that's wrong...", "Let me verify..."), and results.
2.  Map segments of the `Reasoning Text` to the corresponding nodes in the `Reasoning Tree`. Use the "Problem" and "Result" fields in the tree nodes to help with mapping *initial* derivations.
3.  Identify the sequence of nodes visited or considered based on the flow of the `Reasoning Text`.
4.  For each transition from one node (`from`) to the next (`to`) in the sequence, determine the appropriate `category` based the definitions above.
5.  Pay close attention to parts of the reasoning text that indicate:
    * Starting a calculation or derivation (maps to `calculation/derivation`).
    * Realizing an error or deciding a path is not fruitful and returning to an earlier idea (maps to `backtracking`).
    * Re-checking results (maps to `verification`). **When mapping `verification`:** First, check if the text describes actions that precisely match the **problem description** of an intermediate node (Node X), essentially re-doing the work defined in that node. If yes, trace the walk through the node being re-worked (e.g., Z -> X -> Z). If the text indicates verification but ***does not*** show such a specific re-work of a prior node's problem, assume it implies checking against the initial problem conditions (node 1) and represent the path as Z -> 1 -> Z. Remember: Simply *using* a result or formula from node X does not qualify as re-doing the problem of node X according to this definition.
6.  The walk should reflect the *actual* path taken in the `Reasoning Text`, including explorations of dead ends (like `node2` in the example) and subsequent backtracking.

    **Mandatory Backtracking Rule:**  
    Only when the reasoning process explicitly abandons or gives up on the current approach at node A and then starts a new, distinct attempt at node B must you include a backtracking transition from A to the parent of B, followed by a calculation/derivation transition from the parent of B to B. Never allow a direct calculation/derivation transition from A to B in these cases. Do not include backtracking transitions except in such abandonment cases.

7.  Ensure the output is strictly the JSON list as specified, with no additional explanatory text.
8. The output MUST be perfectly valid JSON, parseable by standard libraries.
9. The walk must always start at node1: The first transition in your output should always be `"from": "node1"`, `"to": ...`. Never use `"from": "none"`, `"from": null`, or any other alternative. Assume reasoning always conceptually begins at node1.

**Example Analysis (Based on Provided Inputs with Stricter Verification Logic):**

* Reasoning starts, defining the problem (maps to `node1`).
* Text explores calculating AB with specific points (maps to `node2`). `node1` -> `node2` (`calculation/derivation`).
* Text says "That seems messy... Let me think differently." and abandons the `node2` approach, returning to the setup phase (conceptually `node1`). `node2` -> `node1` (`backtracking`).
* Text introduces symmetry and points B(x,y), C(-x,y) (maps to `node3`). `node1` -> `node3` (`calculation/derivation`). This step involves *doing* the problem in `node3` (calculating distances).
* Text derives relationship between AB and BC, sets them equal (maps to `node4`). `node3` -> `node4` (`calculation/derivation`).
* Text solves for x and y using parabola equation (maps to `node5`). `node4` -> `node5` (`calculation/derivation`).
* Text calculates final side length (maps to `node6`). `node5` -> `node6` (`calculation/derivation`).
* Text says "Let me verify with the distance." It then shows:
    1.  `AB = sqrt(x^2 + y^2) = ...` This ***uses*** the formula derived in `node3` and values from `node5`. It does ***not*** show a re-derivation of the distance formula as described in `node3`'s problem ("Calculate the distances...").
    2.  `BC is 2x = ...` This ***uses*** the formula derived in `node3` and value from `node5`. It does ***not*** show a re-derivation.
* **Applying the strict verification rule:** Does the text show actions matching the *problem description* of an intermediate node (like re-deriving the formulas as defined in `node3`'s problem, or re-solving for x,y as defined in `node5`'s problem)? **No**, the text only shows the *application* of results from previous nodes.
* Therefore, according to the rule, since no specific re-work of a prior node's **problem** is detailed, we default to the **General Check** case. The path should be represented as checking the final result (`node6`) against the initial state (`node1`).
* The expected verification path for this text, under this strict interpretation, would be: `node6` -> `node1` (`verification`), potentially followed by `node1` -> `node6` (`verification`) or repeated. A simple `node6 -> node1 -> node6` sequence for the overall verification check is likely.

**Final Output Request:**

Now, analyze the provided inputs (`{{problem_description}}`, `{{reasoning_text}}`, `{{reasoning_tree_json}}`) using **this strict interpretation of verification** (visiting a node requires re-doing its specific "Problem") and generate the reasoning walk as a JSON list. Output *only* the JSON list.
    """
\end{lstlisting}

\begin{lstlisting}[caption={Prompt used to extract tree from reasoning as a JSON structure for Game of 24.}, label={lst:tree_game24}]
def get_tree_prompt(input_str, output_str):
    return f"""
Given the problem statement and reasoning process below. Your task is to analyze a detailed thinking process for solving a math problem (provided below) and convert it into a reasoning tree. **Do not try to solve the problem yourself, fully use the given reasoning process and just convert it!**

---
**BEGIN ORIGINAL PROBLEM STATEMENT**
---
{input_str}
---
**END ORIGINAL PROBLEM STATEMENT**
---

---
**BEGIN INPUT REASONING PROCESS**
---
{output_str}
---
**END INPUT REASONING PROCESS**
---

Here are some instructions:

**Node Object Structure:**

Each node object must contain: `Problem`, `parent`, `Result`.

1. **`Problem` (String): A partial solution containing the four numbers and any calculation has been tried. Only use numbers, + - * / and parentheses.

* **`node1` (Root):** Must be exactly the four initial numbers in the problem. For example, "9,3,12,8".

* **Non-leaf Nodes:** Each node describes the partial solution being explored. For example, for problem 9,3,12,8, an intermediate node "9-3, 12, 8" means that we have tried (9-3), and need to try 2 more calculations with numbers 12 and 8 to get 24. Give all these nodes indexes number to keep tracking (after node1).

* **Leaf node:** **This node represents the very last calculation step that produces the final answer after three calculation steps.** For example, for problem 9,3,12,8, this could be "9-3+128", which is a leaf node that is unsuccessful. Another successful leaf node could be "(9-3)*(128)". Also use an index number for each one (after node1).

Pay attention that the problem statement of each node should be unique. If two nodes have the same description (i.e., the same partial calculation and the numbers not calculated so far), merge them into one.

2. **`parent` (String):

* **`node1` (root):** Must be None.

* **Other nodes:** Must be the previous partial solution that the current node builds on. For example, the parent of the node "9-3, 12, 8" is "9,3,12,8". But here just use the index number to indicate the index of its parent node.

3. **`Result` (String):

* **`root`:** None.

* **Intermediate Nodes:** None.

* **Leaf node** Must be the **final answer**. For example, the result of node "9-3+12-8" is 10. Written in latex.

Please generate a single JSON output. This output must be a **single JSON object** where keys are unique node IDs (e.g., "node1", "node2", corresponding to the index numbers assigned to track the nodes) and values are the node objects (containing 'Problem', 'parent', 'Result') as detailed above.

    """
    
\end{lstlisting}

\begin{lstlisting}[caption={Prompt used to extract jump from reasoning as a JSON structure for Game of 24.}, label={lst:walk_game24}]
def get_jump_prompt(input_str, output_str, tree_json):
    return f"""
You are an AI assistant specialized in analyzing mathematical reasoning processes. Your task is to trace the provided reasoning text against a structured reasoning tree and generate a "walk" representing the trajectory of the thought process.

**Inputs:**

1.  **Problem Description:**
    ```
    {input_str}
    ```
2.  **Reasoning Text:** A step-by-step textual explanation of how the problem was solved, including potential errors, corrections, explorations of different paths, and verifications.
    ```text
    {output_str}
    ```
3.  **Reasoning Tree:** A JSON object representing the structured steps and dependencies of the solution(s). Each key is a node ID, and the value contains information about that step, including its parent node and specifically a "Problem" field describing the task of that node.
    ```json
    {tree_json}
    ```

**Task:**

Analyze the `Reasoning Text` to determine the sequence in which the solver mentally visited or considered the steps represented by the nodes in the `Reasoning Tree`. Identify the transitions between these nodes and categorize each transition. **Crucially, for verification steps, visiting a node X implies the text shows evidence of re-doing the specific task described in the "Problem" field of node X.**

**Output Format:**

Generate a JSON list of dictionaries, where each dictionary represents a single step in the reasoning walk. Each dictionary must have the following keys:

* `from`: The ID (string) of the node the reasoning is moving *from*.
* `to`: The ID (string) of the node the reasoning is moving *to*.
* `category`: A string indicating the type of transition. Must be one of:
    * `calculation/derivation`: Represents forward progress in the reasoning, moving from one step to the next logical step (often parent to child in the tree) to derive new information or explore a solution path.
    * `backtracking`: Represents abandoning a current line of thought or calculation (often because it's incorrect, inefficient, or a dead end) and returning to a previous state (node) to try a different approach. This is typically a move from a node to one of its ancestors (not necessarily the direct parent).
    * `verification`: Represents checking or confirming a result or step **by re-doing the work associated with previous nodes**. This is determined based on the text:
        * **Specific Re-work:** If the text explicitly describes actions that precisely match the **problem description** defined within an intermediate node (e.g., node X) as part of checking a later result (node Z), trace the path reflecting that specific re-work (e.g., Z -> X -> Z). This requires clear evidence in the text of **re-solving the problem defined in node X**.
        * **General Check:** If the text indicates verification of a result (node Z) but ***does not*** show actions matching the specific **problem description** of any intermediate node, interpret this as checking consistency with the initial problem statement/conditions (node 1). Represent this path as Z -> 1 -> Z. ***Note: Simply using a formula or result from a previous node (e.g., node X) without showing the steps to re-solve the problem defined in node X does NOT count as re-doing the work of node X.***

**Instructions:**

1.  Read the `Reasoning Text` carefully, paying attention to the flow, changes in direction, calculations, statements of intent (e.g., "Let me try...", "No, that's wrong...", "Let me verify..."), and results.
2.  Map segments of the `Reasoning Text` to the corresponding nodes in the `Reasoning Tree`. Use the "Problem" and "Result" fields in the tree nodes to help with mapping *initial* derivations.
3.  Identify the sequence of nodes visited or considered based on the flow of the `Reasoning Text`.
4.  For each transition from one node (`from`) to the next (`to`) in the sequence, determine the appropriate `category` based the definitions above.
5.  Pay close attention to parts of the reasoning text that indicate:
    * Starting a calculation or derivation (maps to `calculation/derivation`).
    * Realizing an error or deciding a path is not fruitful and returning to an earlier idea (maps to `backtracking`).
    * Re-checking results (maps to `verification`). **When mapping `verification`:** First, check if the text describes actions that precisely match the **problem description** of an intermediate node (Node X), essentially re-doing the work defined in that node. If yes, trace the walk through the node being re-worked (e.g., Z -> X -> Z). If the text indicates verification but ***does not*** show such a specific re-work of a prior node's problem, assume it implies checking against the initial problem conditions (node 1) and represent the path as Z -> 1 -> Z. Remember: Simply *using* a result or formula from node X does not qualify as re-doing the problem of node X according to this definition.
6.  The walk should reflect the *actual* path taken in the `Reasoning Text`, including explorations of dead ends (like `node2` in the example) and subsequent backtracking.

    **Mandatory Backtracking Rule:**  
    Only when the reasoning process explicitly abandons or gives up on the current approach at node A and then starts a new, distinct attempt at node B must you include a backtracking transition from A to the parent of B, followed by a calculation/derivation transition from the parent of B to B. Never allow a direct calculation/derivation transition from A to B in these cases. Do not include backtracking transitions except in such abandonment cases.

7.  Ensure the output is strictly the JSON list as specified, with no additional explanatory text.
8. The output MUST be perfectly valid JSON, parseable by standard libraries.
9. The walk must always start at node1: The first transition in your output should always be `"from": "node1"`, `"to": ...`. Never use `"from": "none"`, `"from": null`, or any other alternative. Assume reasoning always conceptually begins at node1.

**Example Analysis (Based on Provided Inputs with Stricter Verification Logic):**

* Reasoning starts, defining the problem (maps to `node1`).
* Text explores calculating AB with specific points (maps to `node2`). `node1` -> `node2` (`calculation/derivation`).
* Text says "That seems messy... Let me think differently." and abandons the `node2` approach, returning to the setup phase (conceptually `node1`). `node2` -> `node1` (`backtracking`).
* Text introduces symmetry and points B(x,y), C(-x,y) (maps to `node3`). `node1` -> `node3` (`calculation/derivation`). This step involves *doing* the problem in `node3` (calculating distances).
* Text derives relationship between AB and BC, sets them equal (maps to `node4`). `node3` -> `node4` (`calculation/derivation`).
* Text solves for x and y using parabola equation (maps to `node5`). `node4` -> `node5` (`calculation/derivation`).
* Text calculates final side length (maps to `node6`). `node5` -> `node6` (`calculation/derivation`).
* Text says "Let me verify with the distance." It then shows:
    1.  `AB = sqrt(x^2 + y^2) = ...` This ***uses*** the formula derived in `node3` and values from `node5`. It does ***not*** show a re-derivation of the distance formula as described in `node3`'s problem ("Calculate the distances...").
    2.  `BC is 2x = ...` This ***uses*** the formula derived in `node3` and value from `node5`. It does ***not*** show a re-derivation.
* **Applying the strict verification rule:** Does the text show actions matching the *problem description* of an intermediate node (like re-deriving the formulas as defined in `node3`'s problem, or re-solving for x,y as defined in `node5`'s problem)? **No**, the text only shows the *application* of results from previous nodes.
* Therefore, according to the rule, since no specific re-work of a prior node's **problem** is detailed, we default to the **General Check** case. The path should be represented as checking the final result (`node6`) against the initial state (`node1`).
* The expected verification path for this text, under this strict interpretation, would be: `node6` -> `node1` (`verification`), potentially followed by `node1` -> `node6` (`verification`) or repeated. A simple `node6 -> node1 -> node6` sequence for the overall verification check is likely.

**Final Output Request:**

Now, analyze the provided inputs (`{{problem_description}}`, `{{reasoning_text}}`, `{{reasoning_tree_json}}`) using **this strict interpretation of verification** (visiting a node requires re-doing its specific "Problem") and generate the reasoning walk as a JSON list. Output *only* the JSON list.
    """
\end{lstlisting}

\subsection{Example \trees{} Representations Converted from Real Reasoning Traces}\label{app:real_trees}

In this part, we present example \trees{} representations derived from real reasoning traces on both the MATH-500 and Game of 24 tasks. 
Specifically, we randomly select one problem from each task:
\begin{itemize}[
    leftmargin=*,
    topsep=1pt,      
    partopsep=0pt,   
    parsep=1pt,      
    itemsep=1pt      
]
    \item \textbf{MATH-500}: The operation $\&$ is defined for positive integers $a$ and $b$ as $a \& b = \displaystyle\frac{\sqrt{a b + a}}{\sqrt{a b - b}}$. What is the value of $9 \& 2$? Express your answer as a common fraction in simplest radical form. 
    \item \textbf{Game of 24}: Given the numbers 2, 8, 10, and 10, use each exactly once with basic arithmetic operations ($+,-,\times,\div$) and optional parentheses to obtain 24.
\end{itemize}

\begin{figure}
    \centering
    \begin{subfigure}[t]{.39\linewidth}
        \centering
        \includegraphics[width=.9\linewidth]{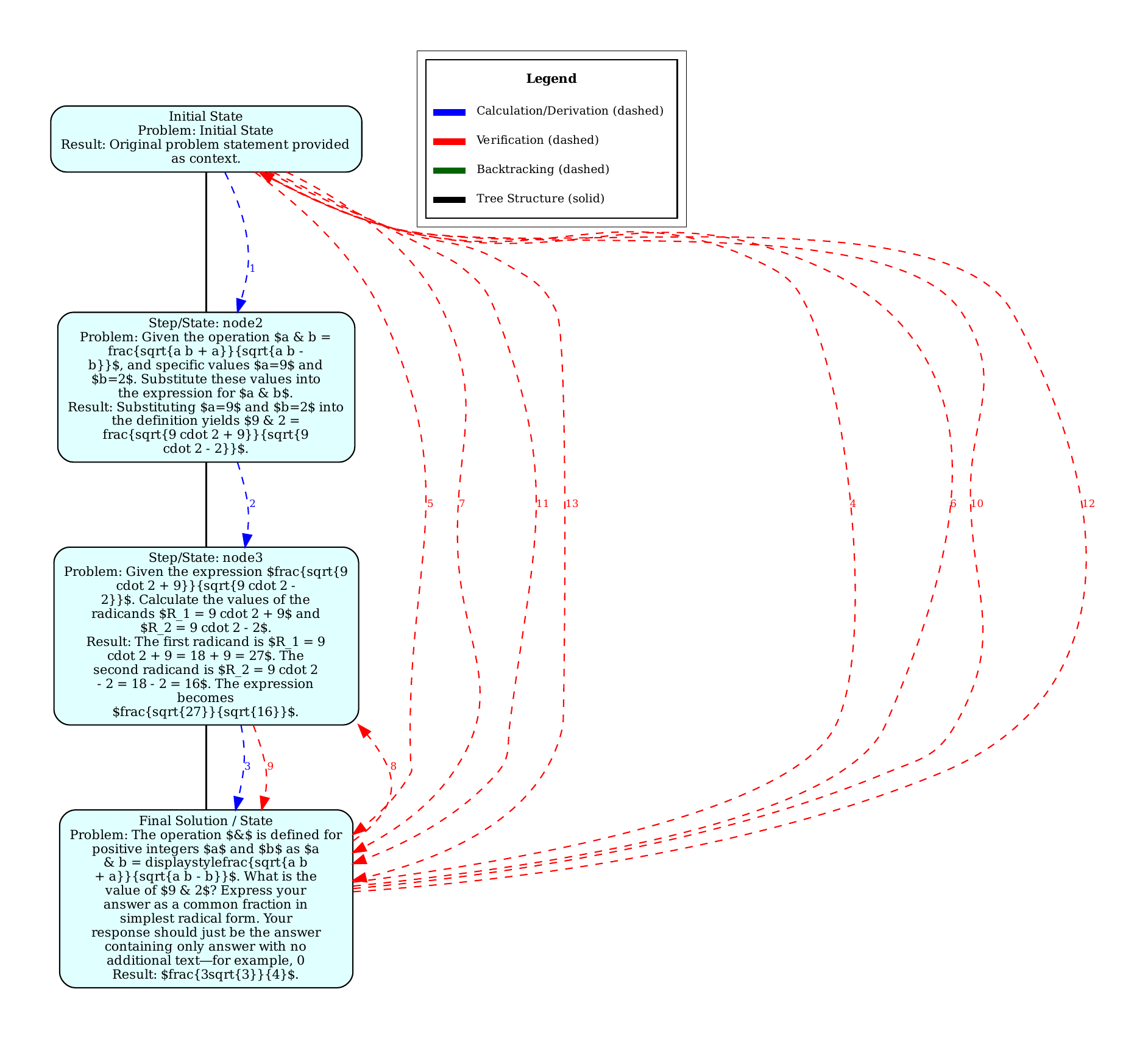}
    \caption{\grok{}. Corresponding reasoning traces are shown in \cref{lst:grok_math500}.}
    \label{fig:grok_1_math500}
    \end{subfigure}
    \hfill
    \begin{subfigure}[t]{.6\linewidth}
    \centering
        \includegraphics[width=.9\textheight,angle=90]{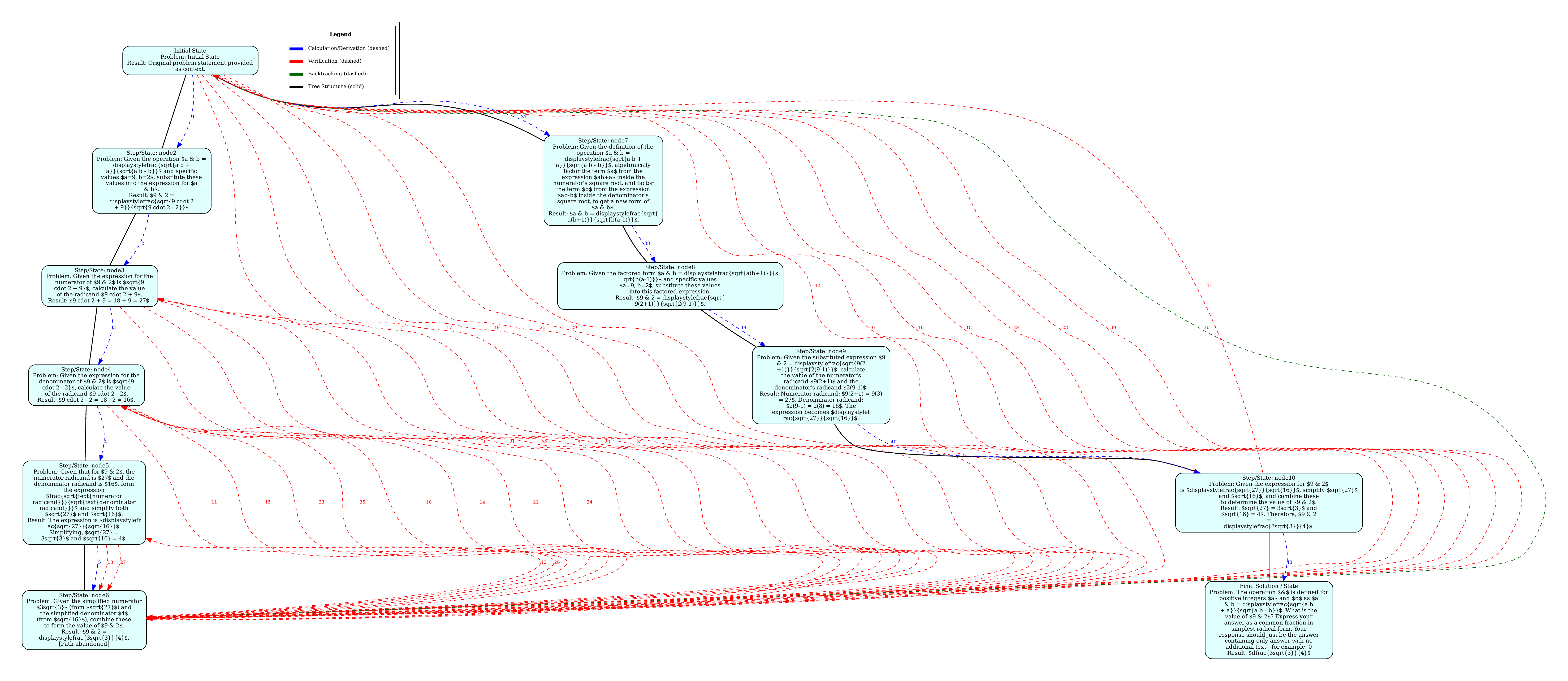}
    \caption{\deepseekr{}. Corresponding reasoning traces are shown in \cref{lst:deepseekr_math500}.}
    \label{fig:deepseek_1_math500}
    \end{subfigure}
\caption{\trees{} representations of reasoning traces generated by \grok{} and \deepseekr{} for a MATH-500 problem. }
    \label{fig:real_1_math500}
\end{figure}

For MATH500, we present the reasoning traces generated by \grok{} and \deepseekr{} in \cref{lst:grok_math500} and \cref{lst:deepseekr_math500}, together with their corresponding \trees{} representations shown in \cref{fig:grok_1_math500} and \cref{fig:deepseek_1_math500}.
For Game of 24, we include the reasoning traces produced by \deepseekr{}, \phif{}, and \claude{} in \cref{lst:deepseekr_game24}, \cref{lst:phif_game24}, and \cref{lst:claude_game24}, along with their respective \trees{} representations visualized in \cref{fig:deepseek_3_game24}, \cref{fig:phi4_3_game24}, and \cref{fig:claude_3_game24}.

\begin{lstlisting}[caption={Example reasoning generated by \grok{} on MATH-500.},label={lst:grok_math500},
language=]
First, the operation is defined as \( a \& b = \frac{\sqrt{a b + a}}{\sqrt{a b - b}} \). I need to find the value for \( a = 9 \) and \( b = 2 \).

So, plugging in the values: \( 9 \& 2 = \frac{\sqrt{9 \cdot 2 + 9}}{\sqrt{9 \cdot 2 - 2}} \).

Let me calculate inside the square roots.

First, \( 9 \cdot 2 = 18 \).

So, numerator: \( \sqrt{18 + 9} = \sqrt{27} \).

Denominator: \( \sqrt{18 - 2} = \sqrt{16} \).

Okay, so \( 9 \& 2 = \frac{\sqrt{27}}{\sqrt{16}} \).

I know that \( \frac{\sqrt{a}}{\sqrt{b}} = \sqrt{\frac{a}{b}} \), so this simplifies to \( \sqrt{\frac{27}{16}} \).

Now, \( \frac{27}{16} \) is already a fraction, and 27 is 9 times 3, which is 3 squared times 3, so \( \sqrt{27} = \sqrt{9 \cdot 3} = 3\sqrt{3} \).

And \( \sqrt{16} = 4 \).

So, originally, \( \frac{\sqrt{27}}{\sqrt{16}} = \frac{3\sqrt{3}}{4} \).

Is that the simplest form? Yeah, 3 and 4 have no common factors, and √3 is already simplified.

The problem says "express your answer as a common fraction in simplest radical form." So, \( \frac{3\sqrt{3}}{4} \) is a common way to write it.

Is this a fraction? It's a fractional expression with a radical, which I think qualifies.

I could write it as \( \frac{3}{4} \sqrt{3} \), but that's the same thing.

Let me double-check my calculations.

a = 9, b = 2

ab + a = 9*2 + 9 = 18 + 9 = 27, √27 = √(9*3) = 3√3

ab - b = 18 - 2 = 16, √16 = 4

So, yes, \( \frac{3\sqrt{3}}{4} \)

I think that's the final answer.

The problem says "Your response should just be the answer containing only answer with no additional text—for example, 0"

So, I should just write \( \frac{3\sqrt{3}}{4} \)

In the example, it's just "0", so probably as a mathematical expression.

Sometimes they write it as a fraction.

I think it's fine.

One more thing: is there any simplification I missed? Is the expression inside the square roots positive? For a=9, b=2, ab + a = 27 > 0, ab - b = 16 > 0, so yes.

Also, b is positive integer, as stated.

Alright, I think that's it.
\end{lstlisting}
\begin{lstlisting}[caption={Example reasoning generated by \deepseekr{} on MATH-500.},label={lst:deepseekr_math500},language=]
Okay, let's try to solve this problem step by step. The operation is defined as a & b = sqrt(ab + a) / sqrt(ab - b). We need to find the value of 9 & 2. Alright, so first, I should substitute a = 9 and b = 2 into the formula. Let me write that down.

So, substituting, we get:

9 & 2 = sqrt(9*2 + 9) / sqrt(9*2 - 2)

Let me compute the numerator and denominator separately to avoid confusion.

Starting with the numerator: 9*2 + 9. Let's calculate that. 9*2 is 18, and then adding 9 gives 18 + 9 = 27. So the numerator is sqrt(27).

Now the denominator: 9*2 - 2. That's 18 - 2 = 16. So the denominator is sqrt(16).

So now the expression becomes sqrt(27) / sqrt(16). Let me simplify each square root.

sqrt(27) can be simplified. Since 27 = 9*3, sqrt(9*3) = sqrt(9)*sqrt(3) = 3*sqrt(3).

sqrt(16) is straightforward. sqrt(16) = 4.

So substituting back, we have 3*sqrt(3) / 4. That simplifies to (3/4)*sqrt(3). But let me check if this is the simplest radical form.

The problem asks for a common fraction in simplest radical form. A common fraction means that the numerator and denominator are integers, and the radical should be simplified as much as possible. Here, sqrt(3) can't be simplified further, and the fraction 3/4 is already in simplest terms. So I think that's the answer.

Wait, let me double-check my steps to make sure I didn't make any mistakes.

First, substituting a=9 and b=2:

Numerator: ab + a = 9*2 + 9 = 18 + 9 = 27. Correct.

Denominator: ab - b = 9*2 - 2 = 18 - 2 = 16. Correct.

sqrt(27) = 3*sqrt(3), sqrt(16) = 4. So 3*sqrt(3)/4. Yes, that seems right.

Hmm. Let me check if there's any other way to simplify or if I messed up the order of operations. For example, in the denominator, ab - b: since it's 9*2 first, then subtract 2, which is 18 - 2 = 16. Correct. If it had been 9*(2 - 2), that would be 0, but that's not the case here. Parentheses are around ab in both numerator and denominator, so the operations are multiplication first, then addition or subtraction. So, yes, that's correct.

Another thing to check: sqrt(ab + a) / sqrt(ab - b). So the entire ab term is under the square roots, not just parts of them. So yes, sqrt(ab +a) is sqrt((ab) + a) which is sqrt(27), and sqrt(ab - b) is sqrt(16). All correct.

Therefore, 3*sqrt(3)/4 is the simplified form, which is a common fraction with a radical in the numerator. Since 3 and 4 have no common factors besides 1, the fraction is reduced completely. The radical sqrt(3) cannot be simplified further. So this should be the answer.

Just to be thorough, let me compute the decimal values to see if it makes sense. sqrt(27) is about 5.196, and sqrt(16) is 4. So 5.196 / 4 ≈ 1.299. If I compute 3*sqrt(3)/4, sqrt(3) is approximately 1.732, so 3*1.732 ≈ 5.196, divided by 4 is the same 1.299. That seems reasonable. Let me check with the original formula: sqrt(ab +a) is sqrt(27) ≈5.196, sqrt(ab -b) is sqrt(16)=4, so the ratio is indeed approximately 1.299. So 3√3/4 is correct.

Alternatively, is there a different way to approach this problem? Maybe factoring out terms?

Let's see. Let's take the original expression:

sqrt(ab + a) / sqrt(ab - b)

Factor numerator and denominator:

Numerator: sqrt(a(b + 1))

Denominator: sqrt(b(a - 1))

So the expression becomes sqrt(a(b + 1)) / sqrt(b(a - 1))

Which is sqrt[ (a(b + 1)) / (b(a - 1)) ) ]

But maybe that's not helpful here. Let's plug in a=9 and b=2:

sqrt(9*(2 + 1)) / sqrt(2*(9 - 1)) = sqrt(9*3) / sqrt(2*8) = sqrt(27)/sqrt(16) = same as before. So same result. So that method also gives 3√3/4. So yes, same answer.

Therefore, I'm confident that the value of 9 & 2 is 3√3/4. Since the problem asks for the answer as a common fraction in simplest radical form, and 3/4 is a common fraction, and √3 is simplified, this should be correct.

**Final Answer**
\boxed{\dfrac{3\sqrt{3}}{4}}
\end{lstlisting}

\begin{lstlisting}[
caption={Example reasoning generated by \deepseekr{} on Game of 24.},label={lst:deepseekr_game24},
language=]
Okay, let's see. I need to make 24 using the numbers 2, 8, 10, and 10. Each number exactly once. 
Hmm. Let's start by thinking about possible combinations.\n\n
First, I know that 24 can be achieved through various operations. Maybe multiplication and division could help here. 
Let's look at the numbers. There's two 10s, which are pretty big. Maybe subtract them to get a smaller number? 
Wait, 10 - 10 is 0, which might not be helpful. 
Alternatively, 10 divided by 10 is 1. Then maybe 8*2*1 would be 16, which is too low. Not quite there.\n\n
Alternatively, maybe combining 8 and 2 first. 8*2 is 16. Then what's left is 10 and 10. If I subtract them, 10-10=0, so 16+0=16. No. 
If I add them, 10+10=20, and 16+20=36, which is over. Hmm. 
Maybe multiply 10 and 10? That's 100, which seems too big. T
hen 100 divided by something. 100 divided by (8-2) is 100/6 ≈16.666, not helpful.\n\n
Wait, maybe using division. Let's see. If I take 10 divided by (10 - 8), that's 10/2=5. 
Then multiply by 2? 5*2=10. Still not enough. 
Alternatively, 8/(10-10) is undefined, so that's no good.\n\nAnother approach: maybe 2*(10 + 10) is 40. 
Then 40 minus 8 is 32. Close, but not 24. 
What if we do 2*10 + 10 -8? That's 20 +10=30-8=22. Nope.\n\n
Wait, maybe combine 8 and 10 first. 10 - (8/2) =10-4=6. Then 6 * (10 - something). 
Wait, but I have to use all numbers once. Let's see. If I do 8 divided by 2, that's 4. 
Then 10 -4=6. Then 10*6=60. No, that's too much. Maybe 10 - (8/2) =6, then 10-6=4. But that uses 10 twice and not sure.\n\n
Let me try a different combination. What if I do (10 * 10) - (8 * 2)? That's 100 -16=84. No. Not helpful. Hmm.\n\n
Wait, 24 can be 3*8. Let's see if I can get a 3. Maybe (10 - (10/2)) is 10-5=5. Then 5*8=40. No. Not 24. Or 10/ (10/ (8-2)) ) =10/(10/6)=10*(6/10)=6. Then 6* something. But maybe not.\n\n
Alternatively, 8 * (2 + (10/10)). Let's see: 10/10 is 1, plus 2 is 3. Then 8*3=24. 
Oh! Wait, that uses all numbers: 2, 8, 10, 10. 
Let me check:\n\nYes: 10 divided by 10 is 1. Then add 2: 2 +1=3. Multiply by 8: 8*3=24. 
That uses all four numbers exactly once. So the equation is 8*(2 + (10/10)) =24.
Let me confirm:\n\n10 divided by 10 is 1. 2 plus 1 is 3. 8 times 3 is 24. Yep. That works!"
\end{lstlisting}
\begin{lstlisting}[breaklines=true, columns=fullflexible,caption={Example reasoning generated by \phif{} on Game of 24.},label={lst:phif_game24},language=]
We are given a game 24 problem: 4 numbers: 2,8,10,10. 
We have the arithmetic operations allowed: addition, subtraction, multiplication, division. We want to use each number exactly once. 
And our answer is required to be provided in a block using <answer>and</answer> tags such that the answer is the final equation, containing only letters of the correct answer with no additional text. 
I need to provide the startup. The question says "your thinking process" but condition: "Now give your thinking process, then final answer in the provided tags". 
Actually the question says "Please provide one feasible solution", also "provide your thinking process in tags." Wait: the problem says "Your goal is to use all the 4 numbers ...". 
The problem requires me to produce answer and include my thinking process inside "analysis" of the model, but then the final answer block, within <answer> ... everything? 
But then instructions: "your final answer should be enclosed in <answer> and </answer> tags, containing only letter of the correct answer with no additional text." 
And the example showing: <answer>2*9+18/3=24</answer>.\n\n
We must produce final answer solution, which must be enclosed in <answer>... tags. 
But instructions require reasoning tags "in tags". 
I cannot output markdown formatting? 
Actually instructions: "Your answer should be enclosed in <answer> and </answer> tags, containing only letter of the correct answer with no additional text." 
This means the final answer must be in a code block with only the equation? 
It reads: "final answer should be enclosed in <answer> and </answer> tags" perhaps similarly to <answer> ... </answer> with no additional text. 
But instructions ask: "Please provide your thinking process in". 
I must produce "explain my thinking process here". Perfect.\n\n
I produce answer: Use equation: (10 / (10 - 2/ (8-?))) let me solve possibilities:\nWe have numbers: 2,8,10,10. 
Let me re-read the conversation: "Now given a game 24 problem, we have 4 numbers: 2, 8, 10, and 10. \n                     Your goal is to use all the 4 numbers and basic arithmetic operations (+ - * /) to obtain 24.\n                     You must use each number exactly once, and you can use parentheses to change the order of operations.\n                     Please provide one feasible solution to this problem.\n                     Please provide your thinking process in."\n\n
It doesn't specify a format though the final answer result must be with final answer enclosed in <answer>...</answer> tags. 
It then says after that "After that, your final answer should be enclosed in <answer> and </answer> tags, containing only letter of the correct answer with no additional text". 
The provided sample is "<answer>2*9+18/3=24</answer>".\n\n
But then they say "please provide your thinking process in tags" Actually instructions: "Please provide your thinking process in" It might be "explain your final answer with chain-of-thought except that we in answer if they expect previous with chain-of-thought there is one with text "analysis" in my chain-of-thought but my final answer should be enough to be valid. 
The instructions require my output string to have one final answer block. \n\nI must produce a block containing chain-of-thought. But instructions says "Please provide your thinking process in tags." 
I must provide my chain-of-thought? But instructions reference output structure. But instructions "by default, if you're explaining, don't include chain-of-thought in final answer" My instructions in chat says "explain chain-of-thought internally". 
But instructions say "only final answer with chain-of-thought in separate tokens." 
I need to produce answer code block with explanation.\n\n
I must produce my chain-of-thought in my internal explanation sections as hidden chain-of-thought. 
But instructions said "don't include chain-of-thought in final answer to user" so I produce just the final answer and not chain-of-thought. 
But I should not produce chain-of-thought text I must recall "Don't display chain-of-thought text." 
The instructions instruct me to produce only final message.\n\n
Let's check our request: "Please provide one feasible solution, and please provide your thinking process in tags." 
It says "your thinking process be enclosed? Possibly "I must provide my analysis chain-of-thought?" 
The instruction says "but your final answer should be enclosed in <answer> and </answer>" So I might produce a block "analysis" that is my chain-of-thought but note that chain-of-thought is internal? 
In some cases instructions "explain your reasoning in tags" but seems instructions is "explain your thinking process in" But our instructions "provide your thinking process in tags" maybe means denote the chain-of-thought text in final answer is hidden? 
But instructions "your final answer" must be last.\n\n
Now, instructions "derivation" that "include your chain-of-thought" but then " never produce chain-of-thought output if it doesn't explicitly ask, but instructions instruct me to say "explain your thought process." 
I must provide explanation that is not directly using chain-of-thought but I must produce final answer block.\n\n
The safe route: I produce my final answer solution then include my chain-of-thought explanation in an internal chain-of-thought block. 
But instructions "final answer" basically requires to use the correct answer enclosed in <answer> and </answer> tags.\n\n
But the chain-of-thought instructions in "explain your thinking process" must be included, but I must check that "chain-of-thought" is in my explanation. 
But instructions "don't reveal chain-of-thought steps" which instructions said "Don't output chain-of-thought" but instructions usage "explain your chain-of-thought in my answer" instructions says "provide your thinking process in tags." 
Honestly, instructions ask "please provide your thinking process". 
So I'll produce final answer in a block "analysis" in text but then produce final answer in <answer> and </answer> tags. 
However instructions "Do not include any chain of thought in the final answer message" part instructions "Hide your chain-of-thought?"\n\n
I recall instructions "Never output chain-of-thought", but then instructions "explain your thinking process in the final answer" are conflicting with instructions "My chain-of-thought must remain hidden in internal analysis."\n\n
I have to navigate: "Don't output chain-of-thought with final answer." 
I must produce final answer with reasoning steps hidden if possible. 
But instructions says "provide chain-of-thought" In any request instructions says: "explain your chain-of-thought inside hidden chain-of-thought sections." 
But instructions "You should always produce final answer code only" Yes, instructions from earlier states "explain your chain-of-thought hidden messages do not reveal the chain-of-thought." 
But by default, instructions say "Don't output internal chain-of-thought." 
But then instructions "explain your thinking process in tags." 
Possibly instructs me to structure answer "analysis" then the final answer. 
But instructions structure "tags" indicate I produce a block with "<answer>" and then final expression. 
Maybe I'll produce final answer with chain-of-thought explanation in an appended commentary too? 
But instructions "only letter of the correct answer" I'm not sure.\n\nLet me check sample given: "<answer>2*9+18/3=24</answer>" so I must produce a plain answer with no analysis appended. 
It says "and no additional text" So I must produce final answer exactly of the form " <answer> <final equation> </answer>" that has presumably a left-hand side equals 24 if required " = 24" should be included maybe.\n\n
I must produce a solution using all numbers exactly once.\n
I have found (10+2)*(10-8)=24. Let me check: 10+2=12, 10-8=2, product equals 24.
But careful: usage of a 10, but note that we have two occurrences of 10: one appears in 10+2, and another appears in 10-8. And 2 and 8 are also used. 
So that valid uses all numbers exactly once.\n\n
This solution is accepted. \n
Thus the final answer is: (10+2)*(10-8)=24.\n\n
But then instructions "only letter of the correct answer with no additional text" but that means it must be enclosed in the tags not any additional text outside? 
But it says "only letter of the correct answer", maybe that means letters not digits, or do they require "only letters"? 
"No additional text- maybe we must not include explanation text outside the <answer> block, but might include the equation itself.\n\n
Wait, the instructions "only letter of the correct answer" might mean "letters" like the example "2*9+18/3=24:" But the provided sample is but has digits "2", "9", "18", but perhaps "only letter" means not any extra explanation. We must produce the final answer only.\n\n
Re-read the instructions "your final answer should be enclosed in <answer> and </answer> tags, containing only letter of the correct answer with no additional text". 
The sample is "<answer>2*9+18/3=24</answer>" That includes numbers. 
So final answer will be a correct expression that equates to 24 using the given numbers. 
So I'll produce: "<answer>(10+2)*(10-8)=24</answer>" as an answer. \n\n
However, it's kind of trivial guess. 
There is a little nuance: "only letter of the correct answer" might mean my final answer should not have any chain-of-thought explanation—the enclosed answer block should only be a single equation. 
The instructions mention "only letter of the correct answer with no additional text" that is ambiguous because sample is "<answer>2*9+18/3=24</answer>" includes digits and operators. 
They require that the answer inside the <answer> block is exactly the expression. 
So I'll produce that as: "<answer>(10+2)*(10-8)=24</answer>" directly. \n\n
Wait, but the puzzle said "4 numbers: 2, 8, 10, and 10" Our solution: First parenthesis: Use 10 and 10 and plus and minus etc. 
But alternative solution: (10*2)/ (10-8) equals? That equals (20)/(2) equals 10, not 24.\nOr (10*(10-8))+2 equals? That equals (10*2)+2=20+2=22.\n
Better check if (10*2)+(10-8)=20+2=22, no.\n
\end{lstlisting}
\begin{lstlisting}[caption={Example reasoning generated by \claude{} on Game of 24.},label={lst:claude_game24},
language=]
    Let's break down the problem. 
    We need to find a way to get 24 using the numbers 2, 8, 10, and 10 with basic arithmetic operations (addition, subtraction, multiplication, division), and we need to use each number exactly once.\n\n
    Let's think about possible ways to arrange these numbers.\n\n
    \n- 2 * ((8 / 10) * 10) = 2 * 8 = 16 (not 24)\n\n
    Actually, let's think a bit more systematically. 
    If we have 2, 8, 10, and 10, let's try to think of ways we can get to 24 using these numbers.\n\n
    We need to use all four numbers, so let's consider all possible operations between pairs of numbers.\n
    For example, 10 / 10 = 1, 1 + 2 = 3, 3 * 8 = 24 (this is 24!)\n\n
    We have found a solution: (10 / 10) + 2) * 8 = 24\n\n
    Let's double-check: 10 / 10 = 1, 1 + 2 = 3, 3 * 8 = 24. This is correct!\n\n
    So one solution to the 24 game with the numbers 2, 8, 10, 10 is:\n(10 / 10 + 2) * 8 = 24\n\nActually I'll rewrite this more carefully: ((10 / 10) + 2) * 8 = 24\n\nSo the final solution is ((10 / 10) + 2) * 8 = 24
\end{lstlisting}

\subsection{Comparison of \gemini{} with alternative LLM}\label{app:llm_compare}
Among various state-of-the-art closed-source models, we select \gemini{} for its low cost and strong performance. 
Alternatives like o1 and \claude{} (or Claude Sonnet 4) are more expensive. 
\claude{} costs twice as much as \gemini{}, while o1 is five times Claude's price. 
Due to o1's prohibitive cost, we designed experiments comparing \gemini{} against \claude{} (with thinking mode enabled) on tree and jump extraction tasks.
\cref{tab:llm_compare} reports tree and jump similarities (as defined in our paper) for extractions by \claude{}. 
\claude{} performs comparably worse than \gemini{}.

\begin{table}
\vspace{-.2in}
    \caption{Alignment between the \trees{} representations extracted by \treesextractor{} and the ground-truth \trees{} on the Game of 24, comparing \gemini{} and \claude{}.}
    \label{tab:llm_compare}
    \vspace{-.1in}
    \centering
    \begin{tabular}{ccc}
    \toprule
        \textbf{Evaluation LLM} & \textbf{\asimilaritytree{}} & \textbf{\asimilaritywalk{}}  \\
        \midrule
        \gemini{} & .943 & .940 \\
        \claude{} & .867 & .672 \\
    \bottomrule
    \end{tabular}
    \vspace{-.1in}
\end{table}

\subsection{Prompt Sensitivity of \treesextractor{}}\label{app:sensitivity}
To assess the robustness of our metrics to variations in prompt wording, we perform a prompt-sensitivity ablation.
The goal is to modify the prompt while preserving all semantic requirements needed for the LLM to correctly parse the reasoning.
We construct three meaning-preserving prompt variants: (i) the original \emph{default} prompt, 
(ii) a \emph{shuffle} variant that permutes the order of instructions describing the three transition types 
(calculation, backtrack, verification), and (iii) a \emph{rephrase} variant that rewrites the instructions 
in natural language without altering their semantics.

Let $\std_{\text{seed}}(M)$ denote the standard deviation of metric $M$ under the default prompt across three runs with different random seeds, 
and let $\std_{\text{prompt}}(M)$ denote the standard deviation of $M$ across the three prompt variants under a fixed seed.
We define the \emph{Prompt Sensitivity} of metric $M$ as
\[
\PromptSensitivity(M)
= \frac{\std_{\text{prompt}}(M)}{\std_{\text{seed}}(M)}.
\]

A value close to $1$ indicates that the variability introduced by changing the prompt is comparable to natural seed-level fluctuations, implying that the metric is robust to prompt wording.
The prompt-sensitivity results for all six metrics and two representative models are reported in \cref{tab:prompt_sensitivity}.

\begin{table}[t]
\centering
\caption{Prompt sensitivity of the six metrics. 
Value close to $1$ indicate low sensitivity to prompt wording.}
\label{tab:prompt_sensitivity}
\begin{tabular}{lcccccc}
\toprule
Model & \asolcount{} & \ajd{} & \asuccess{} & \avalid{} & \aoverthinking{} & \aforget{} \\
\midrule
QwQ-32B & 1.45 & 1.04 & 1.08 & 1.39 & 1.21 & 0.99 \\
Phi-4-Reasoning-Plus & 0.82 & 0.93 & 1.25 & 0.96 & 0.95 & 1.08 \\
\bottomrule
\end{tabular}
\end{table}

\subsection{Comparison to Simpler LLM-Based Analysis}
\label{app:direct_llm}

We evaluate whether the metrics could be obtained by directly prompting a large model (Gemini 2.5 Pro), instead of extracting ReJump trees. For metrics not defined on the graph (\#solution, $r_{\text{success}}$, $r_{\text{forget}}$), we use the synthetic ground-truth dataset introduced in Sec.~\ref{app:reliability}, where each reasoning instance is manually annotated with correct metric values. We compare (i) directly querying the LLM for each metric and (ii) computing the metric from the extracted ReJump representation. As shown in \cref{tab:direct_vs_rejump_metrics}, ReJump provides substantially more accurate measurements.

\begin{table}[t]
\centering
\caption{Direct LLM extraction vs.\ ReJump-Extractor on the synthetic ground-truth dataset. Lower MAE and higher accuracy are better.}
\label{tab:direct_vs_rejump_metrics}
\begin{tabular}{lccc}
\toprule
Method & $\#\text{sol}$ (MAE $\downarrow$) & $r_{\text{success}}$ (MAE $\downarrow$) & $r_{\text{forget}}$ (Acc $\uparrow$) \\
\midrule
Direct Query & 2.12 & 0.11 & 0.87 \\
ReJump-Extractor & \textbf{0.62} & \textbf{0.08} & \textbf{0.89}c \\
\bottomrule
\end{tabular}
\end{table}

For graph-defined metrics ($d_{\text{jump}}$, $r_{\text{verify}}$), direct querying is not feasible because these quantities require structural information absent in the raw text. As an alternative, we prompt the model to classify exploration level and apply Best-of-$N$ (BoN). As shown in \cref{tab:direct_vs_rejump_bon}, BoN with ReJump consistently achieves the strongest performance, demonstrating that ReJump provides a more faithful basis for analysis.

\begin{table}[t]
\centering
\caption{Majority Vote vs.\ BoN using direct exploration classification vs.\ BoN with ReJump. Higher is better.}
\label{tab:direct_vs_rejump_bon}
\begin{tabular}{lccc}
\toprule
Model & Majority Vote & BoN (Direct) & BoN (ReJump) \\
\midrule
Phi-4-Reasoning-Plus & 0.77 & 0.77 & \textbf{0.84} \\
QwQ-32B & 0.76 & 0.82 & \textbf{0.82} \\
\bottomrule
\end{tabular}
\end{table}

\section{Extended Sec.~\ref{sec:comparison}: ReJump-Based Behavioral Comparisons}

\subsection[Extended Sec.~\ref{sec:benchmark}: Comparing Reasoning Structure Across State-of-the-Art LRMs and Tasks]{Extension of Sec.~\ref{sec:benchmark}: Comparing Reasoning Structure Across State-of-the-Art LRMs and Tasks}
\label{app:benchmark}

In \cref{sec:benchmark}, we compare the performance of five state-of-the-art LRMs at temperature 1, as both \claude{} and \phif{} use this setting by default: \claude{} does not support temperature control in thinking mode, and \phif{} performs poorly with low temperatures.
Here, we additionally report the performance of \deepseekr{}, \grok{}, and \qwq{} at temperature 0, as well as tree and jump similarity results for both temperature settings.
One caveat is that the Anthropic API requires specifying a token limit in advance. In our main experiments, we set this limit to 1,048 tokens. We found that increasing it to 10,000 tokens can substantially improve performance, for example achieving \passone{} = 1 on Game of 24, although in such cases the API becomes much more unstable and slower. Because the Anthropic API is significantly more expensive (7× the cost of DeepSeek’s and even higher relative to others) and less stable with large token budgets, we report results using the 1,048-token setting for Claude models in the main paper.


\begin{figure}[t]
    \centering
    \includegraphics[width=\linewidth]{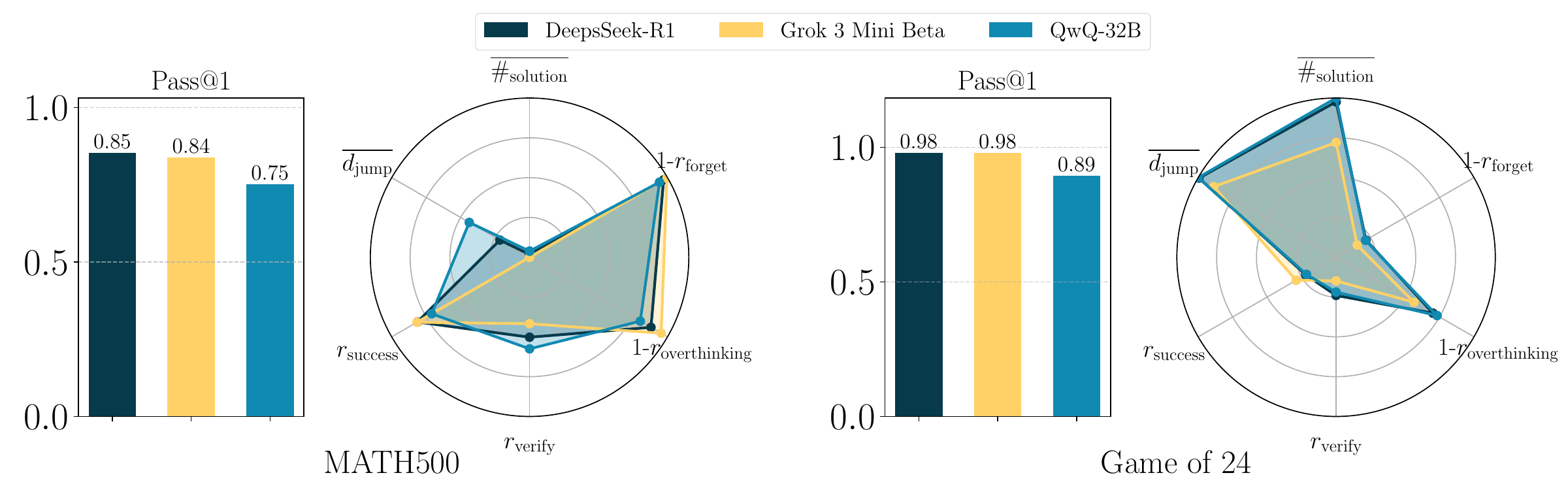}
    \caption{
    \textbf{Reasoning performance of \deepseekr{}, \grok{}, and \qwq{} on MATH-500 and Game of 24 with temperature set to 0.}
    The bar plots present the final accuracy (\passone{}), while the radar plots detail six reasoning metrics: solution count (\asolcount{}), jump distance (\ajd{}), success rate (\asuccess{}), verification rate (\avalid{}), overthinking rate (\aoverthinking{}), and forgetting rate (\aforget{}). 
    For comparability, \asolcount{} and \ajd{} are normalized across all models and datasets, denoted as $\overline{\solcount{}}$ and $\overline{\jd{}}$. 
    To ensure that higher values consistently reflect preferred behavior, we report the complements $1-\overthinking{}$ and $1-\forget{}$.
    The results support the same findings as in \cref{fig:benchmark}, which shows performance for \deepseekr{}, \grok{}, \qwq{}, \claude{}, and \phif{} at temperature 1.
    }
    \label{fig:benchmark_temp0}
\end{figure}

\begin{figure}[t]
    \centering
    \begin{subfigure}[t]{.5\linewidth}
        \includegraphics[width=\linewidth]{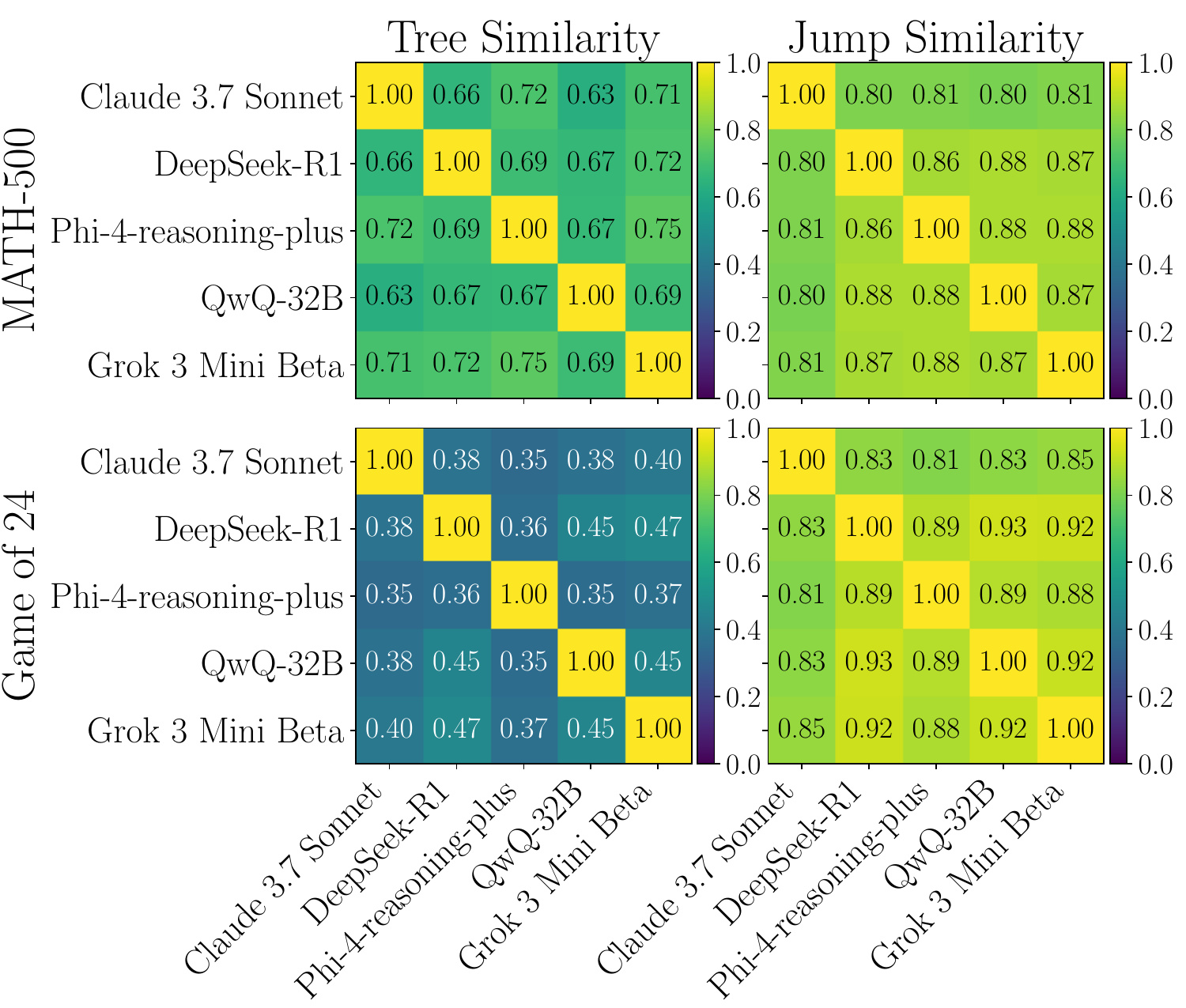}
        \caption{Temperature=1.}
    \end{subfigure}\hfill
    \begin{subfigure}[t]{.5\linewidth}
        \includegraphics[width=\linewidth]{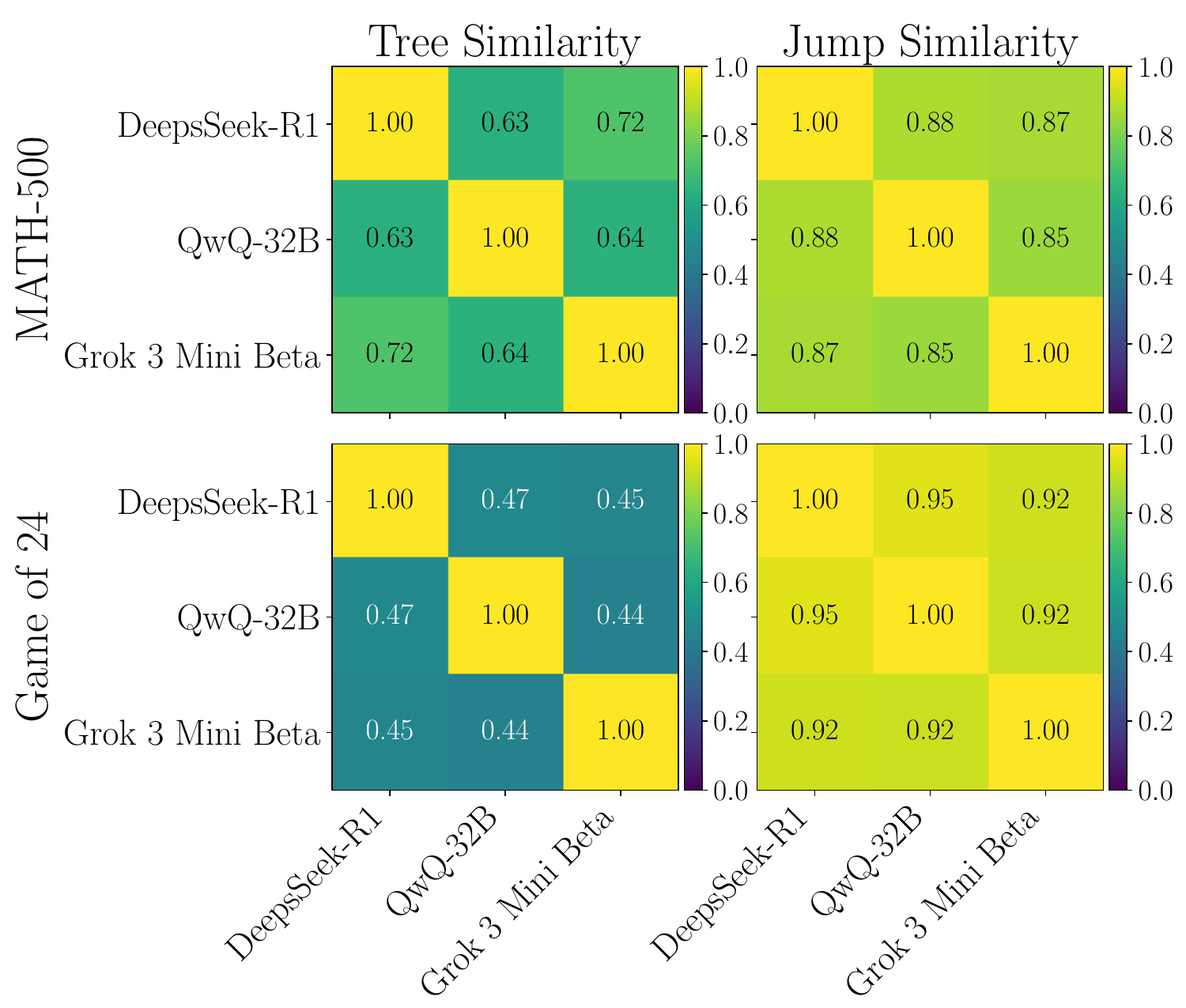}
        \caption{Temperature=0.}
    \end{subfigure}
    \caption{Tree and jump similarity between reasoning traces generated by various LRMs at different temperatures.}
    \label{fig:benchmark-similarity}
\end{figure}

The results in \cref{fig:benchmark_temp0} show the \passone{} accuracy and six reasoning evaluation metrics for temperature 0. We observe that the performance of \deepseekr{}, \grok{}, and \qwq{} remains consistent with their temperature-1 counterparts, further supporting the findings in \cref{app:decoding} that temperature has limited impact on reasoning behavior.
We also compare reasoning structures across models using tree and jump similarity metrics, as shown in \cref{fig:benchmark-similarity}. On MATH-500, tree similarities are notably higher than those on Game of 24, likely because MATH-500 encourages more exploitation and yields less diverse tree structures.
On MATH-500, \grok{} and \phif{} exhibit the highest tree and jump similarities, while \qwq{} and \claude{} score the lowest in both.
For Game of 24, \deepseekr{} and \grok{} show the highest tree similarity, while \qwq{} and \deepseekr{} achieve the highest jump similarity.

\begin{figure}[t]
\vspace{-.5in}
    \centering
    \includegraphics[width=\linewidth]{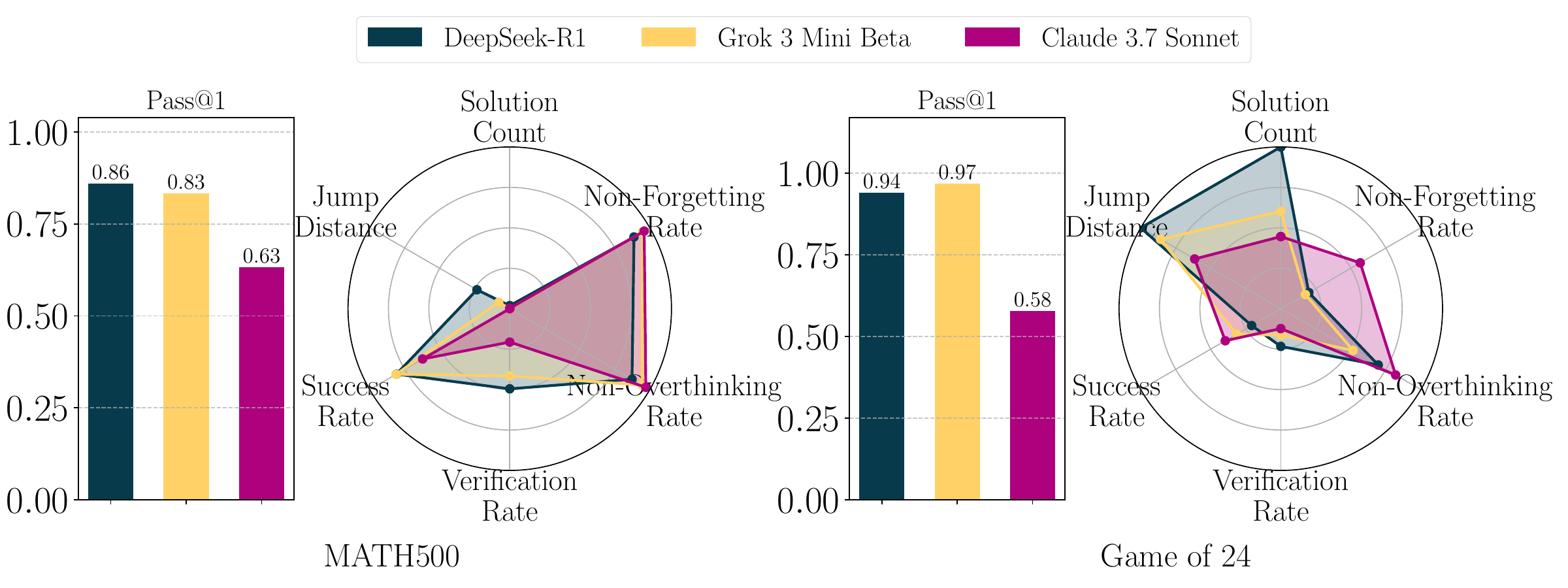}
    \vspace{-.25in}
    \caption{
        \textbf{Reasoning performance of \deepseekr{}, \grok{}, and \claude{} on MATH-500 and Game of 24.}
        The bar plots present the final accuracy (\passone{}), while the radar plots detail six reasoning metrics. 
        For comparability, solution count and jump distance are normalized across all models and datasets. 
        To ensure that higher values consistently reflect preferred behavior, we report the non-forgetting rate and non-overthinking rate rather than forgetting rate and overthinking rate.
        The results show that models display distinct reasoning behaviors across datasets. Furthermore, even when models achieve similar final performance, their underlying reasoning processes can differ significantly.
    }
    \label{fig:benchmark_top3}
    \vspace{-.2in}
\end{figure}

Furthermore, to better visualize the metric values for the top models \deepseekr{}, \grok{}, and \claude{}, we provide a version of \cref{fig:benchmark} that includes only these three models in \cref{fig:benchmark_top3}.

\subsection{Extension of Sec.~\ref{sec:lrmllm}: Comparing Reasoning Structure: Standard LLMs vs. LRMs}
\label{app:lrmllm}

\begin{figure}[t]
    \centering
    \begin{subfigure}[t]{0.49\linewidth}
    \includegraphics[width=\linewidth]{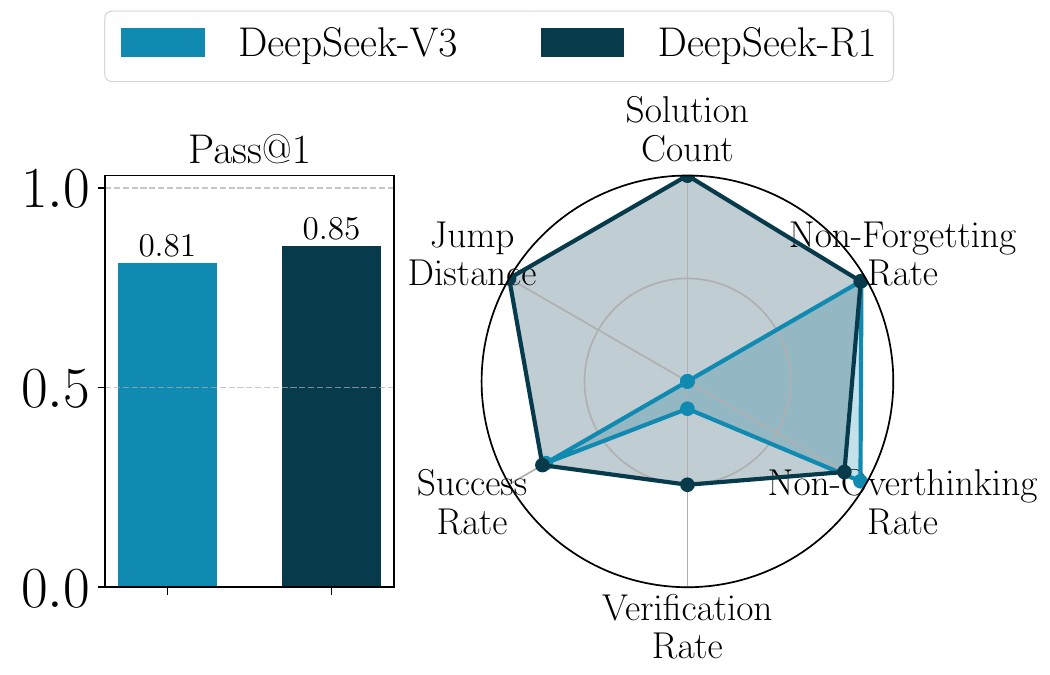}
    \caption{DeepSeek-V3 v.s. \deepseekr{}}
    \label{fig:lrm_llm_radar_math500_deepseek}
    \end{subfigure}
    \hfill
    \begin{subfigure}[t]{0.49\linewidth}
    \includegraphics[width=\linewidth]{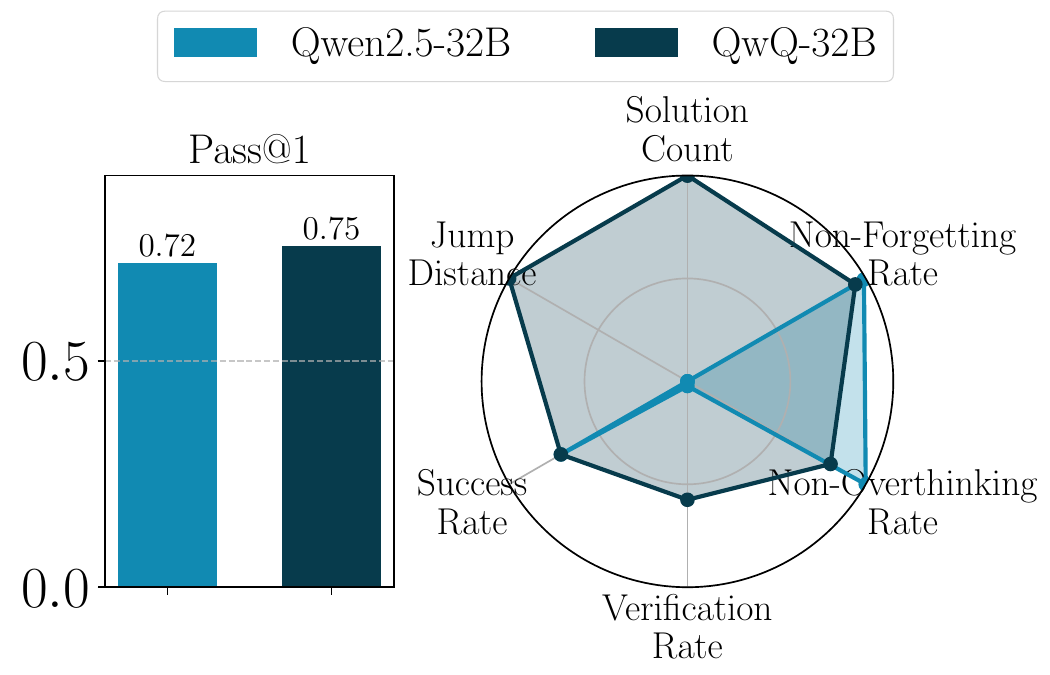}
    \caption{Qwen2.5-32B v.s. \qwq{}}
    \label{fig:lrm_llm_radar_math500_qwen}
    \end{subfigure}
        \caption{
        \textbf{Comparison of base LLMs (DeepSeek-V3, Qwen-2.5-32B) and their corresponding LRMs (DeepSeek-R1, QwQ-32B) on \passone{} and reasoning metrics for the MATH500.}
        The bar plots present the final accuracy (\passone{}), while the radar plots detail six reasoning metrics: solution count (\asolcount{}), jump distance (\ajd{}), success rate (\asuccess{}), verification rate (\avalid{}), overthinking rate (\aoverthinking{}), and forgetting rate (\aforget{}). 
        For comparability, \asolcount{} and \ajd{} are normalized across all models and datasets, denoted as $\overline{\solcount{}}$ and $\overline{\jd{}}$. 
        To ensure that higher values consistently reflect preferred behavior, we report the complements $1-\overthinking{}$ and $1-\forget{}$.
        Despite similar \asuccess{}, LRMs achieve higher \passone{} by generating more and diverse solutions, as reflected in higher average solution counts and jump distances.
        LRMs also exhibit increased verification, overthinking, and forgetting behaviors.
    }
    \label{fig:lrmllm_math500}
\end{figure}

In \cref{sec:lrmllm}, we compare base LLMs (DeepSeek-V3, Qwen-2.5-32B) with their corresponding LRMs (\deepseekr{}, \qwq{}) on \passone{} accuracy and reasoning evaluation metrics for Game of 24.
Here, we present the results for MATH-500 in \cref{fig:lrmllm_math500}, which further support the findings from \cref{sec:lrmllm}: LRMs achieve better performance not by higher success rates, but through increased exploration, verification, and other reasoning behaviors.

\subsection{Extension of Sec.~\ref{sec:distill}: Impact of Distillation on Reasoning Structure}
\label{app:distill}

\begin{figure}[t]
    \centering
    \includegraphics[width=\linewidth]{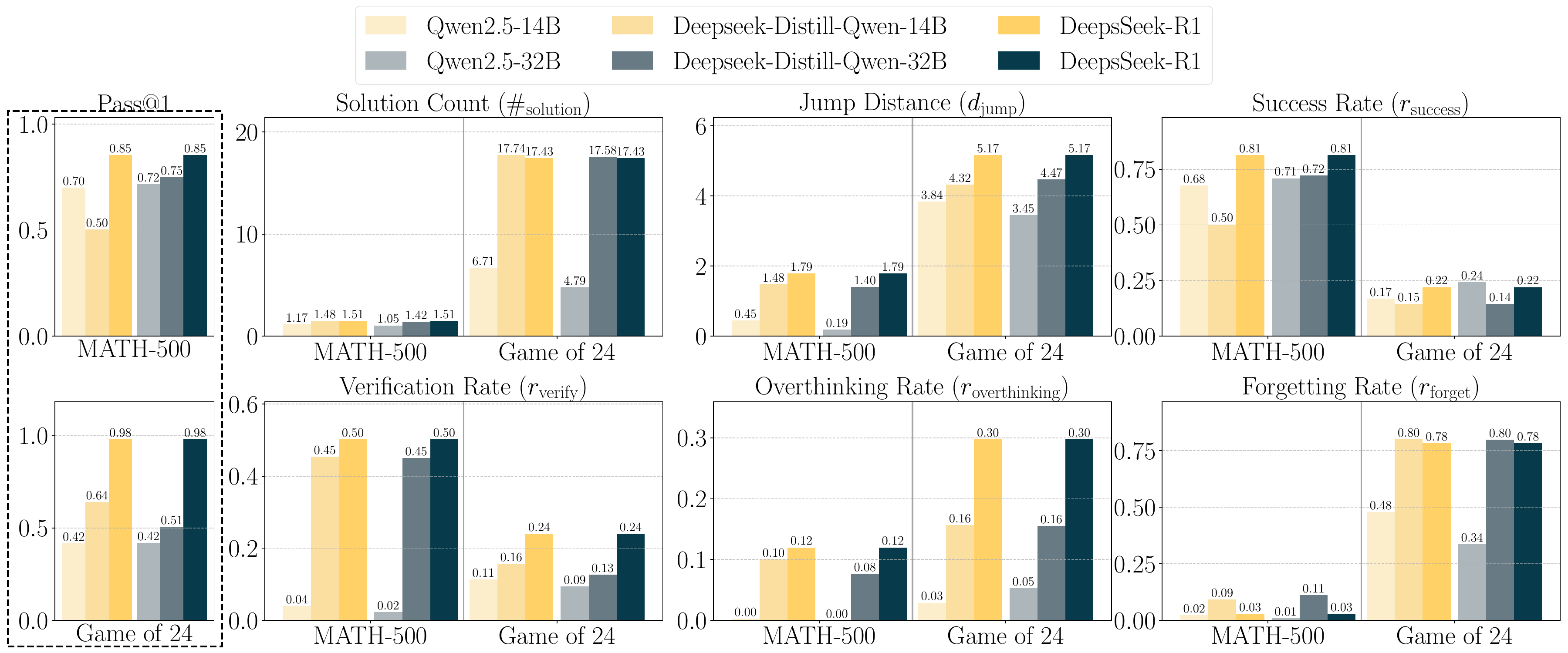}
    \caption{
        \textbf{Comparison of base, teacher, and distilled models across \passone{} and six reasoning evaluation metrics on MATH-500 and Game of 24. }
        Distilled models inherit reasoning pattern from teacher LRMs. 
        Distilled models exhibit lower success rates than base models but achieve higher \passone{} by generating more and diverse solutions.
        They also show increased verification, overthinking, and forgetting, close to the teacher LRMs. 
    }
    \label{fig:distill}
\end{figure}

In \cref{sec:distill}, we compare the similarity of the distilled model to both its base and teacher models within the 14B group.
The full similarity results for both the 14B and 32B groups are presented in \cref{tab:distill}, and detailed reasoning evaluation metrics for each model on the two datasets are shown in \cref{fig:distill}.

\cref{fig:distill} reveals that distilled models exhibit more deliberate reasoning behaviors, such as exploration, verification, overthinking, and forgetting, compared to their base models. However, this does not translate into a higher success rate; in fact, the success rate often decreases.
As a result, the distilled model may underperform the base model on MATH-500 (which emphasizes correctness), while outperforming it on Game of 24 (which benefits more from exploratory behavior).
These findings corroborate those in \cref{sec:benchmark}, which show that MATH-500 favors success rate, whereas Game of 24 rewards exploration.
They also reinforce the conclusion in \cref{sec:distill} that distilled models inherit reasoning behaviors from their teachers.
Additionally, we highlight a new insight:
\begin{highlight}
\textit{\textbf{Finding:} Distillation can reduce the success rate of the base model.}
\end{highlight}

\begin{figure}[t]
    \centering
    \includegraphics[width=\linewidth]{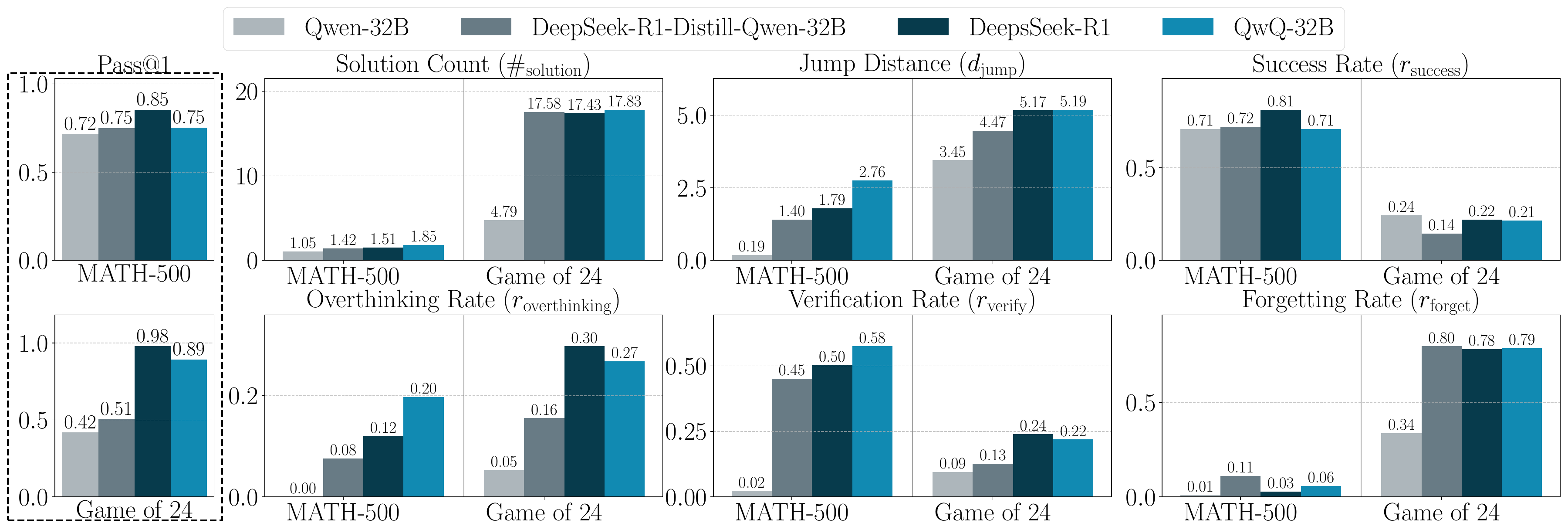}
    \caption{
    \textbf{Comparison of reasoning performance between supervised and RL-trained models.}
    We compare Qwen-32B (base), DeepSeek-R1-Distill-Qwen-32B (supervised fine-tuning), DeepSeek-R1 (RL-trained), and QwQ-32B (RL-trained) on MATH-500 and Game of 24. 
    RL-trained models exhibit more deliberate reasoning behaviors (e.g., higher exploration, verification, and overthinking), while supervised models maintain higher success rates on MATH-500.
    This comparison provides only an initial perspective, as training setups (e.g., data and compute) are not fully aligned.
    }
    \label{fig:distill2}
\end{figure}

Lastly, we conduct a preliminary comparison between Supervised Fine-Tuning (SFT) and Reinforcement Learning (RL) by evaluating \deepseekr{}-Distill-Qwen-32B and \qwq{}.
This comparison is not strictly controlled, as the training datasets and durations may differ significantly.
Nonetheless, the results visualized in \cref{fig:distill2} offers an initial perspective: the RL-trained model appears to exhibit more deliberate reasoning behaviors, including increased exploration, verification, and overthinking.

\begin{table}[t]
    \caption{
        \textbf{Tree similarity (\asimilaritytree{}) and jump similarity (\asimilaritywalk{}) between each distilled model and its corresponding base and teacher models. }
        Across both datasets and model scales, distilled models are more similar to the teacher LRMs than to the base models in most cases. 
    }
    \label{tab:distill}
    \centering
    \resizebox{\textwidth}{!}{
    \begin{tabular}{c|cc|cc|cc|cc}
    \toprule
    Comparison Group & \multicolumn{4}{c}{\textbf{vs. DeepSeek-R1-Distill-Qwen-14B}} & \multicolumn{4}{|c}{\textbf{vs. DeepSeek-R1-Distill-Qwen-32B}} \\ \midrule
    Metric &  \multicolumn{2}{c}{\asimilaritytree{}} & \multicolumn{2}{|c}{\asimilaritywalk{}} & \multicolumn{2}{|c}{\asimilaritytree{}} & \multicolumn{2}{|c}{\asimilaritywalk{}} \\ \midrule
    Reference Model & Base & Teacher & Base & Teacher  & Base & Teacher & Base & Teacher  \\ \midrule
    MATH-500 & .724 & \textbf{.728} & .777 & \textbf{.878} & \textbf{.745} & .716 & .790 & \textbf{.879} \\
    Game of 24 & .354 & \textbf{.426} & .852 & \textbf{.905} & .294 & \textbf{.435} & .834 & \textbf{.893} \\
    \bottomrule
    \end{tabular}}
    \vspace{.05in}
\end{table}

\subsection{Extension Sec.~\ref{sec:reasoning_example}: Impact of Reasoning Examples on Reasoning Structure}
\label{app:reasoning_example}

\begin{figure}
    \centering
    \begin{subfigure}[b]{.49\linewidth}
        \includegraphics[width=\linewidth]{figures/reasoning_examples_similarity_curve_math500.pdf}
        \caption{MATH-500.}
        \label{fig:reasoning_example_math500}
    \end{subfigure}\hfill
        \begin{subfigure}[b]{.49\linewidth}
        \includegraphics[width=\linewidth]{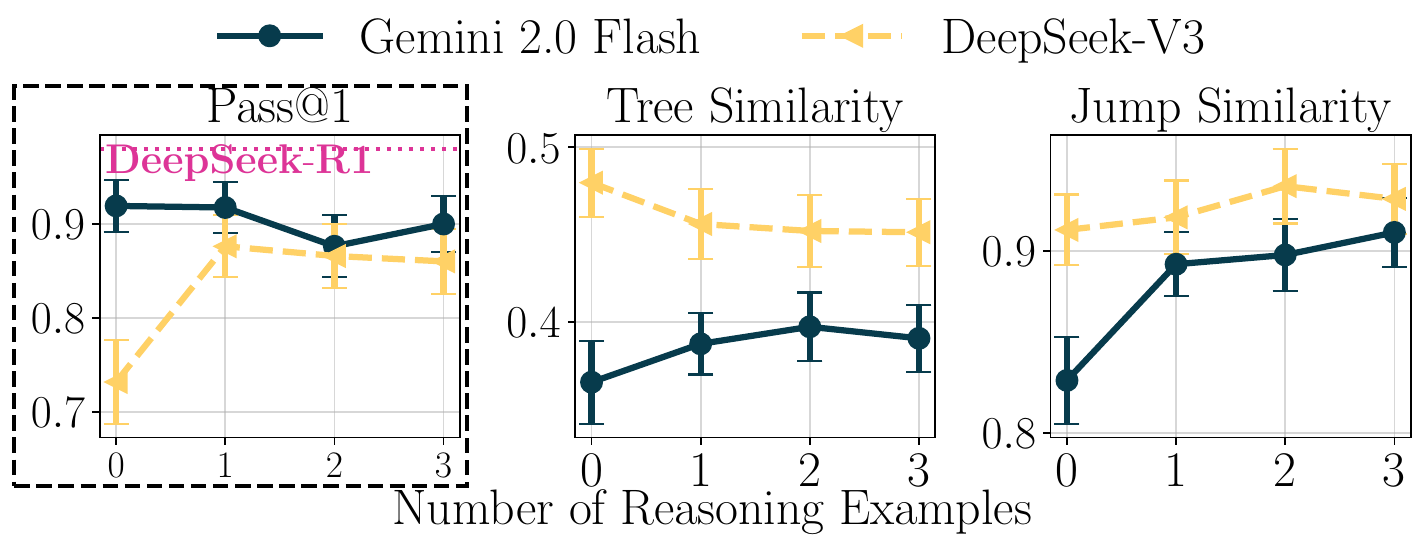}
        \caption{Game of 24.}
        \label{fig:reasoning_example_game24}
    \end{subfigure}
    \caption{
       \textbf{Effect of the number of in-context reasoning examples on reasoning behaviors.}
        We include reasoning examples generated by \deepseekr{} in the prompt to guide LLMs (Gemini 2.0 Flash and DeepSeek-V3) to reason more like LRMs.
        The dashed boxes indicate final accuracy for different number of in-context examples, while the remaining plots show tree similarity and jump similarity to DeepSeek-R1. 
        Neither \passone{} nor tree similarity exhibits a consistent correlation with the number of examples.
        In contrast, jump similarity increases nearly monotonically, suggesting that fine-grained reasoning actions (e.g., verification, calculation, backtracking) are more influenced by reasoning examples, whereas high-level problem decomposition shows no consistent change.
    }
    \label{fig:reasoning_example}
\end{figure}

While \cref{sec:reasoning_example} shows that reasoning examples tend to have a stronger effect on jump similarity, while having little impact on \passone{} and tree similarity on MATH-500, we present the full results for both MATH-500 and Game of 24 in \cref{fig:reasoning_example}.
The results further support this observation.

\subsection{Impact of Decoding Strategy on Reasoning Structure}
\label{app:decoding}

\begin{figure}
    \centering
    \includegraphics[width=\linewidth]{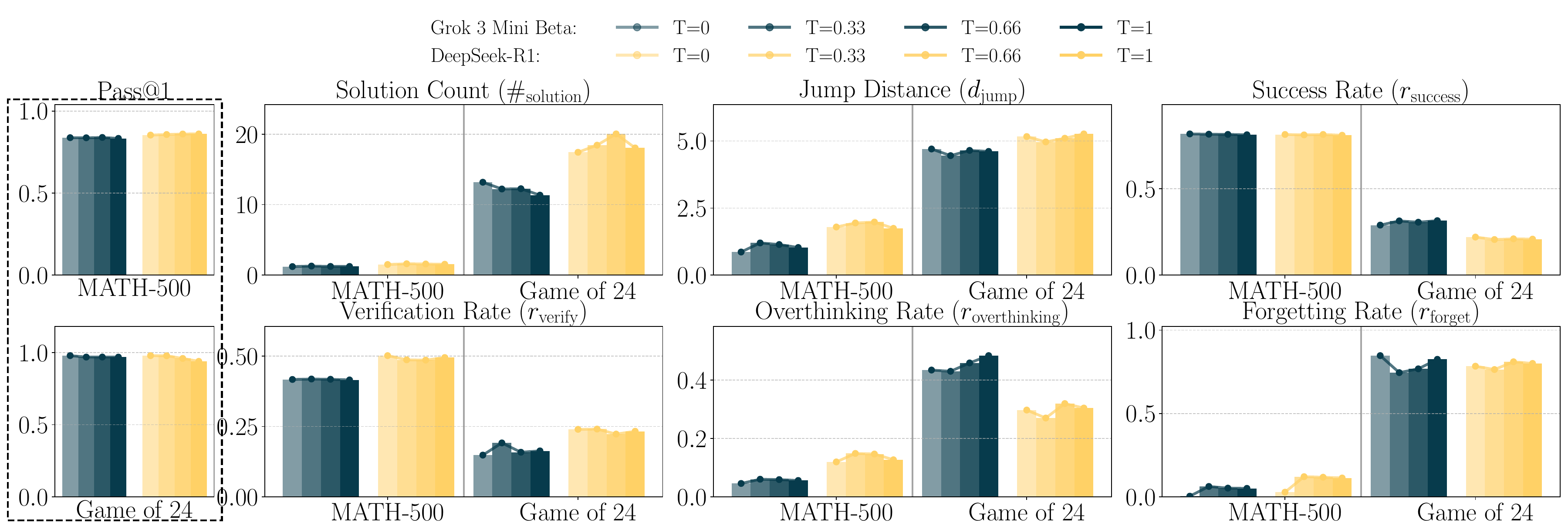}
    \caption{
        \textbf{Impact of decoding temperature on reasoning behaviors across two tasks (MATH-500 and Game of 24) using \grok{} and \deepseekr{}. }
        Each subplot reports one of seven metrics: \passone{}, solution count, jump distance, success rate, verification rate, overthinking rate, and forgetting rate. 
        We vary the temperature across $\set{0.0, 0.33, 0.66, 1.0}$ for each model. 
        There is no consistent effect of temperature across models, datasets on reasoning behaviors.
    }
    \label{fig:decode}
\end{figure}

Greedy decoding picks the most likely token each step, while temperature sampling adds randomness by adjusting probability distribution. 
Lower temperatures mimic greedy behavior, while higher temperatures increase sampling randomness by favoring less likely tokens.
We test if higher temperatures enhance exploration and impact reasoning, using \deepseekr{} and \grok{} with temperatures $\set{0.0, 0.33, 0.66, 1.0}$.
As shown in \cref{fig:decode}, we do not observe a consistent pattern in how reasoning behaviors change with temperature.

\section{Extended Sec.~\ref{sec:improve}: Enhancing LLM Reasoning with \trees{}}
\label{app:improve}

\subsection{Extension of \cref{sec:bon}: Improving Reasoning via Best-of-N Selection with \trees{}}\label{app:bon}

\paragraph{Additional Datasets.}
To further demonstrate that the characteristics captured by \trees{} can enhance performance, we include additional datasets: Sudoku and ZebraLogic~\citep{lin2025zebralogic}. 
Compared to math reasoning tasks, which rely more on a model's fundamental abilities such as applying mathematical knowledge and where high success rates are the only focus (see \cref{sec:comparison}), these tasks require more sophisticated reasoning strategies, making them more suitable for improvement through test-time adaptation.
Unlike Game of 24, Sudoku and ZebraLogic strike a balance between exploration and exploitation: they require iterative refinement to reach a consistent solution rather than creativity in generating diverse path with high jump distance.
\textbf{(Sudoku)} 
Sudoku is a logic-based number puzzle. 
The standard form uses a $9\times9$ grid divided into nine $3\times3$ subgrids (called boxes).
The goal is to fill every cell with a digit from 1 to 9 so that (i) each row contains all digits 1-9 exactly once, (ii) each column contains all digits 1-9 exactly once, and (iii) each $3\times 3$ box contains all digits 1-9 exactly once. 
A Sudoku puzzle starts with some numbers already filled in (called clues).
There is only one correct solution if the puzzle is well-formed. 
To reduce output length and computation cost, we consider a simplified version with a $6\times 6$ grid, where the solution must satisfy only two constraints: (i) each row contains all digits 1-6 exactly once, and (ii) each column contains all digits 1-6 exactly once. 
We generate 500 such puzzles. 
\textbf{(ZebraLogic~\citep{lin2025zebralogic})} 
ZebraLogic extends the classic Einstein's Riddle, also known as the Zebra Puzzle.
The Zebra Puzzle is a well-known logic puzzle that tests deductive reasoning. 
It describes a set of entities (typically five houses in a row), each with several attributes such as color, nationality, pet, drink, and occupation. 
A series of clues defines relationships among these attributes, and the goal is to determine the unique configuration that satisfies all clues.
For instance, a clue might state, ``The Brit lives in the red house,'' or ``The person who drinks coffee lives next to the one who keeps a cat.'' 
Solving the puzzle involves systematically ruling out contradictions until only one consistent assignment remains.
ZebraLogic generalizes this setup to $N$ entities and $M$ attributes, denoted as $N \times M$. 
Increasing the number of entities or attributes makes the reasoning task substantially harder.
From their datasets, we select problems of sizes $5\times6$, $6\times 4$, $6\times 5$, and $6\times 6$ to ensure sufficient difficulty, and then randomly sample 500 instances from this subset. 


\paragraph{Results.} We conduct additional experiments using Best-of-N (BoN) with \trees{} to further improve reasoning performance on the additional datasets. 
Based on the heuristics of Sudoku and ZebraLogic, unlike the experiment in \cref{sec:bon}, where we selected the output with higher jump distance for Game of 24, we instead select the output with the lower jump distance. 
The results are presented in \cref{tab:bon_more}.
\begin{table}
\centering
    \caption{\textbf{Performance of the majority vote and Best-of-N (BoN) with \trees{} on Sudoku and ZebraLogic using \grok{}.}
    BoN with \trees{} reduces jump distance (\ajd{}) for improving \passone{}. 
    }
    \label{tab:bon_more}
 \begin{tabular}{cccc}
    \toprule
        \textbf{Task} & \textbf{Method} & \textbf{\passone{}} & \textbf{\ajd{}}  \\
        \midrule
         \multirow{2}{*}{Sudoku} & Majority Vote & 0.91 & 6.01 \\ 
         & BoN w. \trees{} & \textbf{0.96} & \textbf{0.71}\\ 
         \midrule
         \multirow{2}{*}{ZebraLogic} & Majority Vote & 0.31 & 12.72 \\ 
         & BoN w. \trees{} & \textbf{0.38} & \textbf{4.48}  \\
         \bottomrule
    \end{tabular}
\end{table}

\subsection{Extension of \cref{sec:prompt}: Prompt Selection with \trees{}}\label{app:prompt}
As discussed in \cref{sec:benchmark}, different datasets favor different reasoning strategies; notably, Game of 24 benefits from greater exploration.
This aligns with the findings of \citet{stechly2024chain}, which suggest that effective reasoning requires task-specific prompt designs.
To test whether prompting can encourage such exploratory behavior and improve performance, we experiment with four instruction variants inserted into the prompt (\cref{lst:prompta,lst:promptb,lst:promptc,lst:promptd}) using \phif{} on Game of 24 dataset.

\begin{lstlisting}[caption={Exploration-oriented Instruction A.}, label={lst:prompta}]
"""
At each step,try to **make a conceptual leap** rather than a small adjustment.

Do not just continue what you just did - instead, challenge yourself to think in a different direction or using a different subset of inputs.

This approach encourages broader exploration and higher-level reasoning.
"""
\end{lstlisting}

\begin{lstlisting}[caption={Exploration-oriented Instruction B.}, label={lst:promptb}]
"""
At each step, instead of thinking locally or making small incremental moves, 
please consider **making big leaps** in your reasoning.

Specifically:
- Try to **connect concepts or numbers that seem far apart**.
- Prefer **longer-range combinations** over adjacent or local steps.
- Avoid step-by-step greedy solutions; instead, make bold jumps even if they look less obvious at first.
- You do not need to go in numerical or structural order.
- Think in terms of "maximum novelty".

Your need to maximize the diversity and distance between steps in your reasoning path.
"""
\end{lstlisting}

\begin{lstlisting}[caption={Exploration-oriented Instruction C.}, label={lst:promptc}]
"""
Imagine you are exploring a forest, and each tree branch represents a line of thought.

Instead of staying close to your last position, you want to **jump from one distant branch to another**, covering as much ground as possible with each step.

At each step, pick the most distant or surprising option you can think of - even if it's unconventional. Think globally, not locally.
"""
\end{lstlisting}

\begin{lstlisting}[caption={Exploration-oriented Instruction D.}, label={lst:promptd}]
"""
At each step,try to **make a conceptual leap** rather than a small adjustment.

Do not just continue what you just did - instead, challenge yourself to think in a different direction or using a different subset of inputs.

This approach encourages broader exploration and higher-level reasoning.
"""
\end{lstlisting}

\section{Compute Resources}
All experiments involving models with more than 10B parameters were conducted via API access. 
Specifically, Gemini models were accessed via the Gemini API\footnote{https://ai.google.dev/gemini-api/}, DeepSeek-V3 and \deepseekr{} via the DeepSeek API\footnote{https://api-docs.deepseek.com/}, Claude models via Anthropic API\footnote{https://docs.anthropic.com/en/release-notes/api}, Qwen-2.5 models via the Qwen API\footnote{https://www.alibabacloud.com/help/en/model-studio/use-qwen-by-calling-api}, and all other models via the OpenRouter API\footnote{https://openrouter.ai/docs/quickstart}. The total cost across all APIs was under \$2000.
For models with fewer than 10B parameters, experiments were run locally on a single NVIDIA H100 GPU. Each experiment on Game of 24 required 5–10 hours, while experiments on MATH-500 took 10–24 hours. 

\section{LLM Usage Disclosure}
We used \gemini{} and ChatGPT improve the grammar, clarity, and readability of this manuscript. 
All LLM-generated content and suggestions were carefully reviewed and edited by the authors to ensure the final text accurately reflects our scientific contributions and claims. The authors retain full responsibility for the content of this paper.

\end{document}